\documentclass[11pt]{article}

\usepackage[preprint]{acl}

\usepackage{times}
\usepackage{latexsym}

\usepackage[T1]{fontenc}

\usepackage[utf8]{inputenc}

\usepackage{microtype}

\usepackage{inconsolata}

\usepackage{graphicx}

\usepackage{hyperref}
\usepackage{url}

\usepackage{booktabs}  
\usepackage{amsmath} 
\usepackage{amssymb}

\usepackage{cleveref}
\usepackage{xspace}
\usepackage{enumitem}
\usepackage{arydshln}
\usepackage{wrapfig}
\usepackage{subfigure}

\usepackage{amsthm}

\usepackage{tcolorbox}
\tcbuselibrary{breakable,skins}
\usepackage{multirow}

\newcommand{\se}{\mathcal{R}}
\newcommand{\Ours}{\textsc{DeepControl}\xspace}

\usepackage{listings}
\makeatletter
\renewcommand{\verbatim@font}{\normalfont\rmfamily}
\makeatother

\title{Adaptive Information Control for Search-Augmented LLM Reasoning}

\author{Siheng Xiong, Oguzhan Gungordu, James C. Kerce, Faramarz Fekri\\
Georgia Institute of Technology\\
\texttt{\{sxiong45,ogungordu3\}@gatech.edu, 
clayton.kerce@gtri.gatech.edu,}\\
\texttt{faramarz.fekri@ece.gatech.edu}
}

\begin{document}
\maketitle

\begin{abstract}
Search-augmented reasoning agents interleave multi-step reasoning with external retrieval, but uncontrolled retrieval can introduce redundant evidence, saturate the context, and destabilize reinforcement learning (RL).
Existing outcome-based RL methods provide only sparse terminal rewards, offering limited guidance for intermediate information-acquisition decisions.
We propose \Ours, an adaptive information-control framework based on \emph{information utility}, a state-dependent estimate of the marginal value of retrieved evidence.
The framework regulates information acquisition along two axes: \emph{extent}, i.e., whether retrieval should continue, and \emph{resolution}, i.e., how much retrieved detail should be exposed.
It implements these controls through retrieval-continuation guidance, hierarchical granularity control, and an annealed control-forcing scheme.
This enables the policy to internalize effective acquisition behavior during training and operate without external control at test time.
Across seven benchmarks, \Ours consistently outperforms strong RL and retrieval baselines without explicit information control; compared with Search-R1, it improves average performance by +9.4 and +8.6 points on Qwen2.5-7B and Qwen2.5-3B, respectively.
Additional analyses show improved search effectiveness, training stability, and evidence utilization.
The code and model checkpoints are available at \url{https://github.com/xiongsiheng/DeepControl}.
\end{abstract}

\begin{figure}[t]
\centering\includegraphics[width=\linewidth]{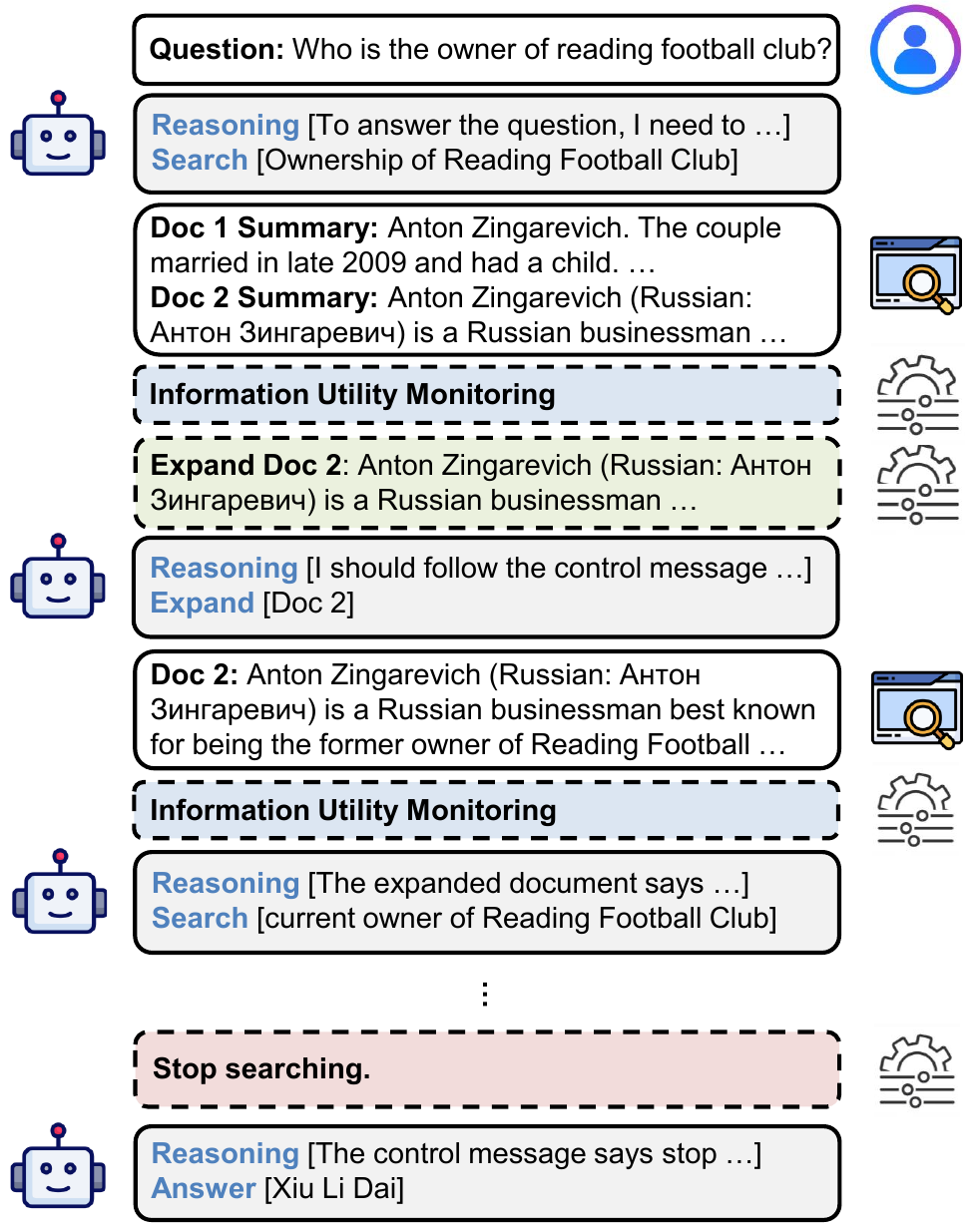}
\caption{Example rollout of \Ours. The agent iteratively performs reasoning and information acquisition via \texttt{search} and \texttt{expand}. After each retrieval step, a lightweight controller monitors the utility of the obtained information and issues control messages to guide subsequent agent actions.}
\label{fig:example}
\end{figure}

\section{Introduction}

Recent advances have enabled search-augmented reasoning agents that interleave multi-step reasoning with external information acquisition, allowing language models to solve complex, knowledge-intensive tasks beyond their parametric knowledge \citep{zheng2025deepresearcher, du2025deepresearch, huang2025deep, zhou2026lras}. 
As these agents operate over increasingly rich information environments, with retrievable content growing in amount, length, and structural complexity, their performance is no longer limited by search availability or reasoning capacity alone. 
Instead, a key bottleneck is \emph{uncontrolled information acquisition}. 
In practice, repeatedly retrieving more evidence can lead to context saturation, redundant or noisy information accumulation, and interference between reasoning and retrieved content, ultimately degrading decision quality rather than improving it \citep{yu2024rankrag, jin2025search, sun2026diagnosing,xie2026slimsearcher}. 
These failures suggest that more retrieval does not necessarily yield better reasoning.

To mitigate such issues, prior work \citep{jin2025search, zheng2025deepresearcher} has predominantly relied on outcome-based reinforcement learning (RL) \citep{guo2025deepseek}, using final answer correctness as the sole training signal to guide both reasoning and retrieval decisions.
However, outcome-only supervision provides limited guidance for \emph{intermediate} retrieval decisions, especially in long-horizon settings \citep{xiongenhancing}. 
As a result, agents may over-retrieve when evidence is weak or queries are poorly specified, accumulating unnecessarily long contexts instead of relying on internal knowledge; conversely, they may terminate retrieval prematurely even when additional evidence remains beneficial. 
More fundamentally, outcome-only signals are ill-suited to regulate \emph{when} to retrieve, \emph{how much} to retrieve, and \emph{at what granularity} to expose evidence.
These are not isolated failures of stopping, retrieval, or context construction, but different manifestations of a broader \emph{information-acquisition control} problem.

What is missing is explicit and adaptive control over information acquisition. 
We argue that information acquisition should be controlled along two complementary axes: \emph{extent}, namely whether to continue acquiring additional evidence, and \emph{resolution}, namely how much detail of the retrieved content should be exposed.
This view differs from prior approaches that mainly improve retrieval quality, stopping behavior, or evidence expansion in isolation, without treating them as a unified control problem under RL.

In this work, we introduce \Ours, an adaptive information control framework for search-augmented reasoning agents. 
Our method augments standard online RL with utility-driven training-time control signals that provide intermediate guidance for regulating both the extent and the resolution of information acquisition. 
An annealed control strategy gradually reduces external intervention during training, enabling the policy to internalize effective information acquisition behaviors while retaining the flexibility of learning from interaction.

In summary, our main contributions are threefold:
\begin{itemize}
    \item We propose \emph{information utility}, a state-dependent measure of the marginal value of retrieved evidence for search-augmented reasoning. 
    The utility combines novelty and effectiveness and serves as a practical signal for information acquisition control during training.

    \item Building on information utility, we introduce two complementary control mechanisms: \emph{retrieval continuation control}, which regulates acquisition extent by mitigating premature stopping and over-retrieval, and \emph{granularity control}, which regulates acquisition resolution by selectively expanding high-utility content within hierarchical information structures. 
    We further adopt an annealed control strategy that facilitates internalization.

    \item We conduct extensive experiments across multiple tasks and datasets, showing consistent improvements in reasoning accuracy, training stability, and evidence utilization across diverse search-augmented reasoning benchmarks.
\end{itemize}

\begin{figure*}[t]
    \centering
    \subfigure[Novelty]{\includegraphics[width=0.32\textwidth]{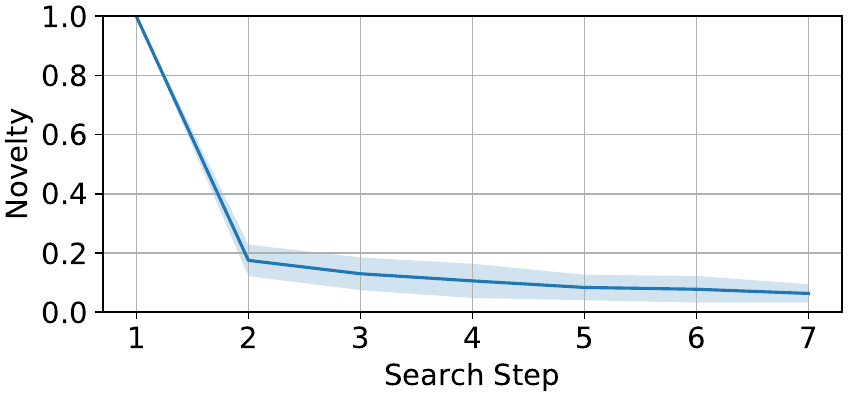}
    }
    \subfigure[Effectiveness]{\includegraphics[width=0.32\textwidth]{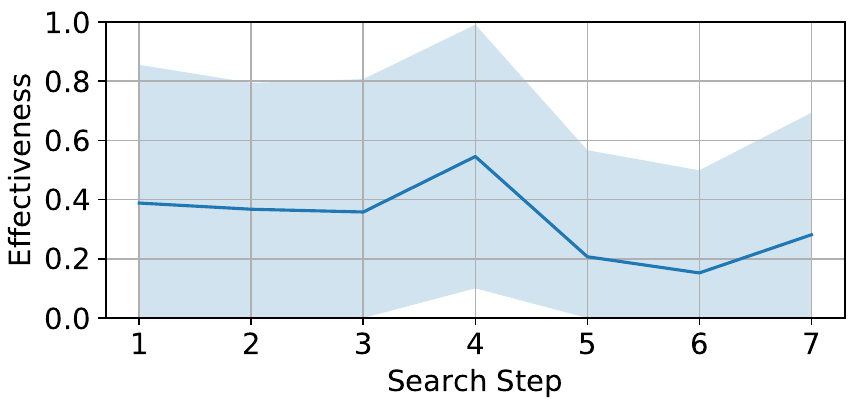}
    }
    \subfigure[Utility]{\includegraphics[width=0.32\textwidth]{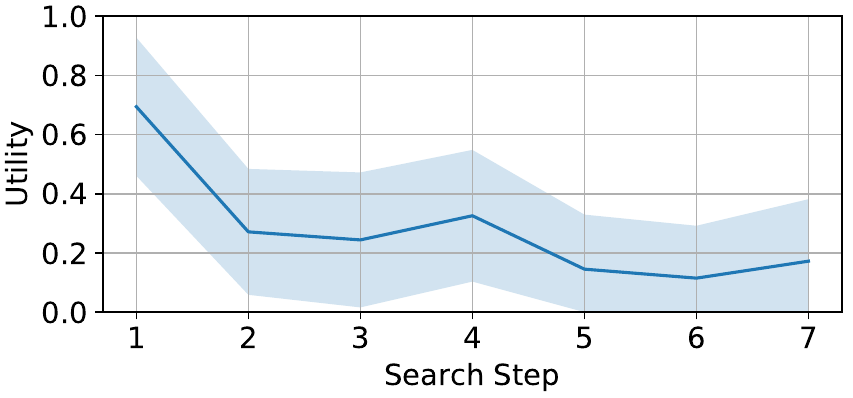}
    }
    \vspace{-7pt}
    \caption{Information novelty, effectiveness, and utility across search steps. Utility combines novelty and effectiveness to quantify the overall usefulness of newly retrieved information as retrieval progresses.}
    \label{fig:novelty_effectiveness_utility_curves}
\end{figure*}

\section{Preliminaries}
\label{sec:preliminaries}

\subsection{Problem Formulation}
\label{sec:setting}

We consider a search-augmented reasoning agent that interleaves multi-step reasoning with external retrieval. 
Given a task $u \sim \mathbb{P}(\mathcal{U})$, the agent with policy $\pi_\theta$ interacts with a search engine $\se$ and maintains a reasoning state $s_t$ containing the accumulated context, including retrieved evidence and intermediate reasoning. 
At each step $t$, the agent samples a structured action
\[
a_t = (h_t, \alpha_t, \xi_t) \sim \pi_\theta(\cdot \mid u, s_t),
\]
where $h_t$ denotes reasoning tokens, $\alpha_t$ denotes the action type (e.g., \texttt{retrieve}), and $\xi_t$ denotes the action parameters (e.g., search queries). 
A rollout trajectory is $\tau = (s_0, a_0, \ldots, a_{T-1}, s_T)$, which terminates when the agent outputs a final answer or reaches a step limit.

\subsection{Online RL with Search-Augmented Reasoning Agents}
\label{sec:online_rl}

Online RL alternates between a \emph{rollout phase}, where trajectories are generated by the current policy, and an \emph{update phase}, where the policy is optimized using collected rollouts. The objective is to maximize task success while regularizing deviation from a reference policy $\pi_{\text{ref}}$.

\paragraph{Proximal Policy Optimization.}
Proximal Policy Optimization (PPO)~\citep{schulman2017proximal} is a widely used actor--critic algorithm for LLM post-training~\citep{ouyang2022training}. PPO maximizes a clipped surrogate objective with advantages $A_t$ computed via Generalized Advantage Estimation (GAE) using a value function $V_{\zeta}$, with clipping parameter $\epsilon$ controlling update stability.

\paragraph{Group Relative Policy Optimization.}
Group Relative Policy Optimization (GRPO)~\citep{shao2024deepseekmath} is a group-based policy-gradient method widely used in recent LLM post-training. Instead of learning a value function, it estimates advantages \emph{relatively} within a group of sampled responses for the same prompt and optimizes the policy with KL regularization to a reference policy.

\paragraph{Adaptations for search-augmented reasoning.}
In search-augmented reasoning, retrieved content is produced by an external search engine rather than the policy, so policy-gradient updates apply only to tokens generated by the language model. Existing search-augmented RL approaches typically rely on an outcome-based reward
\(
\mathbb{I}\!\left[y_{\text{pred}} = y_{\text{gold}}\right],
\)
which evaluates final answer correctness using Exact Match (EM).

\paragraph{Limitations of outcome-based RL training.}
Search-augmented reasoning with outcome-based RL enables agents to learn tool usage, but introduces several issues (see the failure cases in \Cref{box:error_cases_motivation}):

\begin{enumerate}[label=\arabic*), left=5pt]
\item \textbf{Suboptimal search behavior.}
Agents often exhibit suboptimal search. 
For example, when evidence is unavailable or queries are poorly specified, agents may over-retrieve and accumulate unnecessary context instead of relying on internal knowledge.
Without explicit control signals, policy learning receives little guidance for intermediate retrieval decisions.

\item \textbf{Information overload.}
Many approaches \citep{lin2023ra, yu2024rankrag, jin2025search} append raw retrieved content to the context, quickly exceeding context limits, especially with long sources (e.g., webpages or papers). Mitigations such as small top-$k$ (e.g., $k=3$) risk missing key evidence, while longer contexts can exacerbate context saturation and make training and inference more expensive.

\item \textbf{Unstable training.}
Outcome-based RL provides sparse supervision, making optimization sensitive to errors along long reasoning trajectories \citep{xiong2025enhancing}. This issue is amplified when starting from weaker base models, where inaccurate exploration further destabilizes training.
\end{enumerate}

\begin{figure*}[t]
\centering
\includegraphics[width=\linewidth]{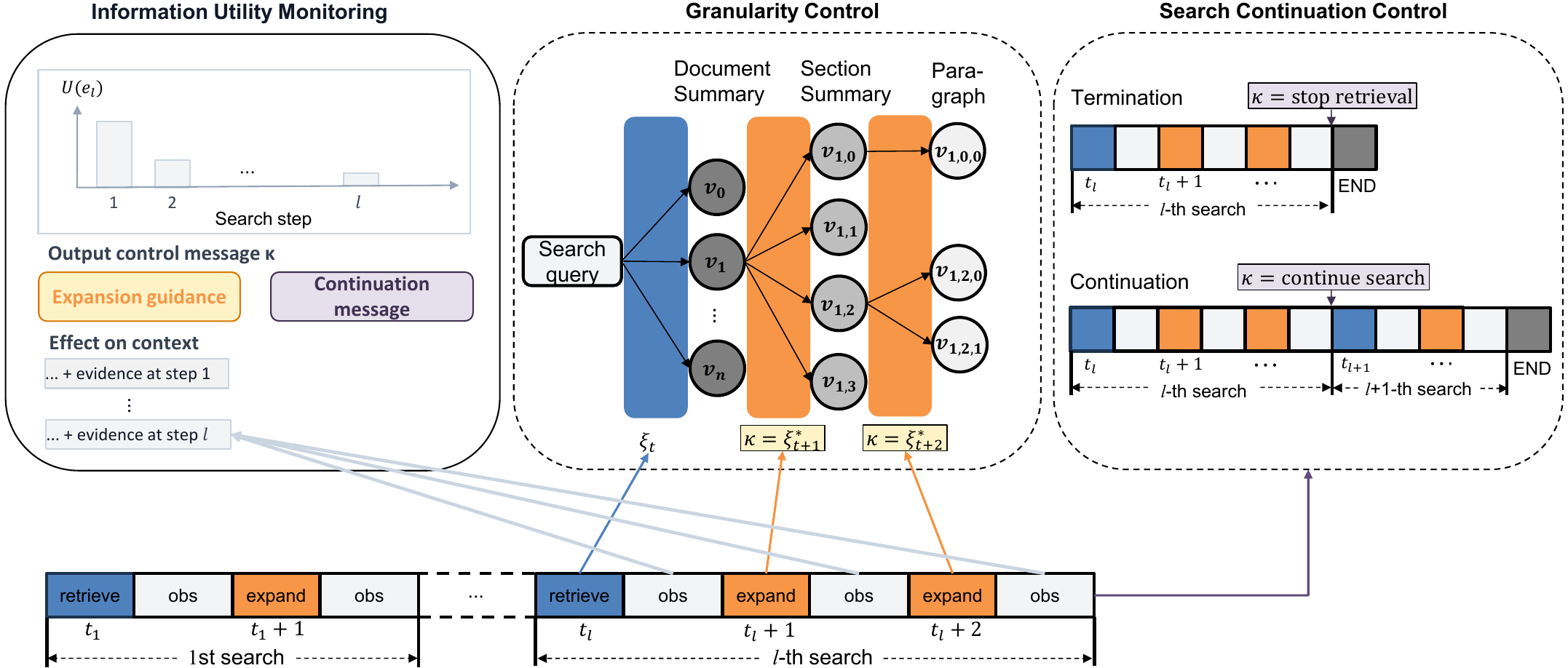}
\caption{Overview of adaptive information control. The controller monitors step-level utility $U(e_l)$ and regulates information acquisition along two axes: (i) \emph{resolution}, via expansion guidance $\xi^\ast$, and (ii) \emph{extent}, via continuation messages $\kappa$ that stop or continue search. All interventions are aligned with the timeline (bottom).}
\label{fig:information_control}
\end{figure*}

\section{Adaptive Information Control}
Since sparse outcome rewards provide little supervision for intermediate information-acquisition decisions, we introduce a shared utility signal for training-time control over two axes:
\emph{extent} (whether to continue retrieval) and
\emph{resolution} (how much retrieved detail to expose).

\subsection{Information Utility}
\label{sec:info_utility}

The value of external information acquisition is \emph{state-dependent} and must be evaluated relative to the agent’s current reasoning state. 
In our framework, information acquisition is organized into discrete \emph{search steps} (\Cref{fig:information_control}), each consisting of a retrieval action followed by optional expansion actions that refine the retrieved information (\Cref{sec:granularity_control}).

Let \( t = \{0,1,\ldots,T-1\} \) index primitive actions (e.g., \texttt{retrieve}, \texttt{expand}, \texttt{answer}) and \( l = \{0,1,\ldots,L-1\} \) index search steps. Let \( t_l \) denote the primitive step of the \(l\)-th retrieval. The \(l\)-th search step spans from \(t_l\) to \(t_{l+1}-1\), where \(t_{l+1}\) is the next retrieval (or termination). Let \( u \) denote the task and \( s_{t_l} \) the reasoning state before the \(l\)-th retrieval, with retrieval output \( e_l \). We define the information utility of the \(l\)-th search step as
\begin{align} 
U(e_l \mid & u, s_{t_l}) = \rho \cdot \mathrm{Novelty}(e_l \mid s_{t_l}) \nonumber + \\
&(1-\rho) \cdot \mathrm{Effectiveness}(e_l \mid u, s_{t_l}). 
\end{align}
where \( \rho \in [0,1] \) balances novelty and effectiveness (\Cref{fig:novelty_effectiveness_utility_curves}). 
For simplicity, we write \(U(e_l)\) for \(U(e_l \mid u, s_{t_l})\). In our framework, information utility serves as a dense, state-dependent \emph{training-time control signal} for retrieval decisions, in contrast to outcome-based rewards that provide only terminal supervision.
Detailed definitions and discussions are provided in \Cref{appendix:info_utility}.
Alternative utility formulations are discussed and evaluated in \Cref{para:alternative_utility_formulations}.

\subsection{Granularity Control via Hierarchical Selective Expansion}
\label{sec:granularity_control}

Granularity control regulates the \emph{resolution} of information exposure.
We expose coarse evidence first and selectively expand into finer-grained units only when beneficial.

At search step $l$, retrieval returns $k$ hierarchical sources
$e_l=\{\mathcal{G}_l^{(i)}\}_{i=1}^{k}$, each a rooted tree
$\mathcal{G}_l^{(i)}=(\mathcal{V}_l^{(i)},\mathcal{E}_l^{(i)})$, where nodes \( v \in \mathcal{V}_l^{(i)} \) are evidence units at different resolutions and edges \( (v, v') \in \mathcal{E}_l^{(i)} \) indicate that $v'$ is a refinement of $v$.
After retrieval, instead of injecting all leaf-level content, we initialize by appending only the retrieved root nodes to the current context; the resulting injected set is denoted by \(\mathcal{C}_{t_l} \subseteq \bigcup_{i=1}^{k}\mathcal{V}_l^{(i)}\).
The agent may then perform a variable number of \texttt{expand} actions to incrementally grow the observed set until the next retrieval (or termination). An expansion action at primitive step \(t'\) selects refinement edges \(\xi_{t'} \subseteq \bigcup^{k}_{i=1}\mathcal{E}_l^{(i)}\) and updates
\begin{equation}
\mathcal{C}_{t'}=\mathcal{C}_{t'-1}\cup\{\,v' \mid v\in \mathcal{C}_{t'-1},\ (v,v')\in \xi_{t'}\,\}.
\end{equation}
During training, we score retrieved leaves with utility \(U(\cdot)\), select the top-\(k_{\text{expand}}\) leaves, and trace their ancestors to derive target expansions \(\{\mathcal{C}^\ast_{t_l+1}, \ldots, \mathcal{C}^\ast_{t_{l+1}-1}\}\).
The model is then supervised to follow these targets, prioritizing high-utility evidence while limiting context growth.
The effect of hierarchical evidence construction is discussed and evaluated in \Cref{para:hierarchy_effect}.

\subsection{Search Continuation Control}
\label{sec:retrieval_control}

Search continuation control regulates the \emph{extent} of information acquisition.
We intervene only when the agent’s search decision appears clearly suboptimal under the utility signal.

\paragraph{Termination.}
If the information utility remains below a threshold \( \delta_{\text{stop}} \) for \( m_{\text{stop}} \) consecutive search steps, we define the stopping index
\begin{equation}
l^\star
=
\min_{l \in [m_{\text{stop}}-1, L-1]}
\max_{j \in [l-m_{\text{stop}}+1,l]} U(e_j) < \delta_{\text{stop}}.
\end{equation}
Upon reaching \( l^\star \), a control signal
\( \kappa \) = "\texttt{Stop searching}"
is injected, explicitly terminating further search steps.

\paragraph{Continuation.}
Conversely, the agent may terminate search even when additional evidence remains useful.
During training, we trigger a one-shot continuation when recent utilities remain high but the model is still insufficiently confident about the gold answer.
Concretely, let \(S_l\) denote the aggregated target score (defined in \Cref{appendix:info_utility}) computed under evidence \(\mathcal{C}_{t_{l+1}-1}\). If the agent attempts to terminate at search-step index \(l\) and
\begin{equation}
\label{eq:continue_trigger}
S_l \le \tau_{\text{score}}
\;\wedge\;
\min_{j \in [l-m_{\text{cont}}+1,l]} U(e_j)\ge \delta_{\text{cont}},
\end{equation}
we inject a one-shot control signal
\(\kappa\) = "\texttt{Continue the search for one additional step}".
Here \(\tau_{\text{score}}\) is a confidence threshold on the gold-answer score (see \Cref{para:gold_free_utility_estimation} for \textit{gold-answer-free} variants), and the information utility is above $\delta_{\text{cont}}$ for $m_{\text{cont}}$ consecutive search steps. 

\paragraph{Discussion.}
Under this design, search continuation is primarily governed by the agent’s learned policy, while information utility serves as a training-time monitoring signal that triggers corrective control when necessary. Detailed hyperparameter settings and ablations are provided in \Cref{appendix:retrieval_continuation_control}.

\subsection{Reinforcement Learning with Information Control}
\label{sec:rl_with_control}

External control signals can stabilize early training, but they must be internalized for reliable test-time behavior. We therefore adopt an annealed \emph{control-forcing} RL scheme with two rollout modes and a composite reward.

\begin{figure}[t]
\centering
\includegraphics[width=0.95\linewidth]{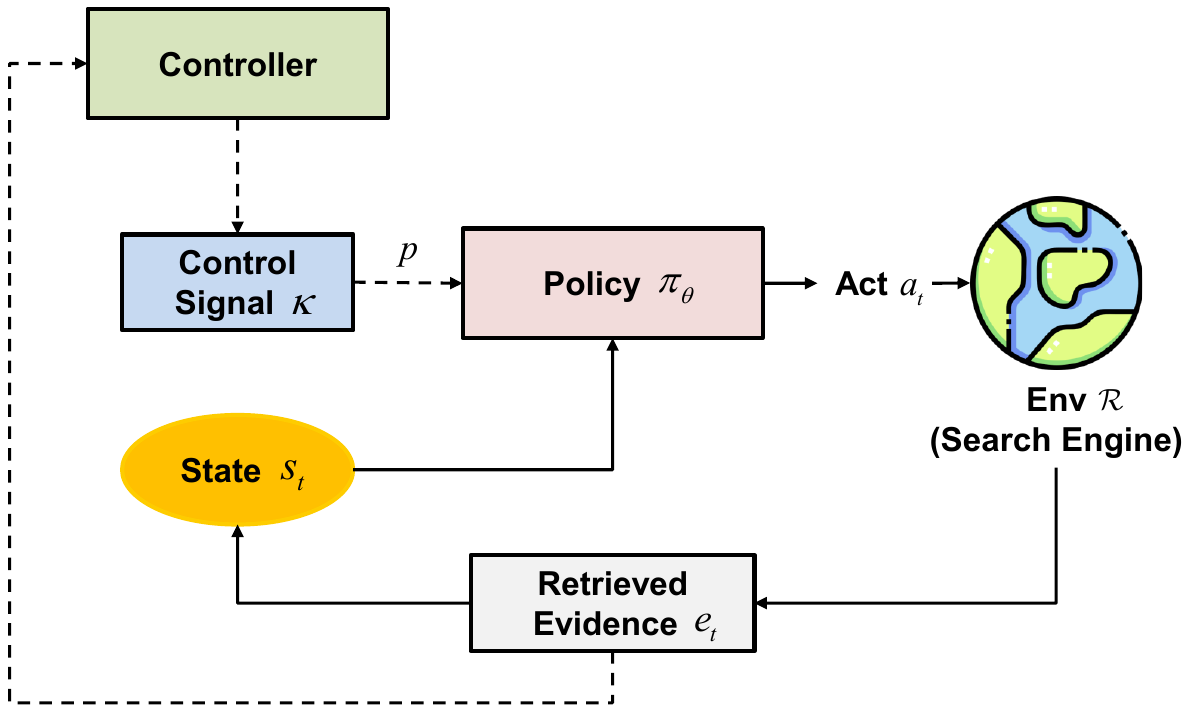}
\caption{Trajectories generated in rollout mode with and without information control.}
\label{fig:retrieval_with_reasoning_with_control}
\end{figure}

\paragraph{Rollout Modes.}
During rollouts, the agent samples between two modes (\Cref{fig:retrieval_with_reasoning_with_control}), selecting mode (1) with probability $p$ and mode (2) with $1-p$.

\noindent
\textbf{(1) With Information Control.}
During controlled rollouts, a training-time controller monitors retrieval utility and triggers a control signal $\kappa$ at time $t^\star$ when abnormal behavior is detected. The next action is sampled as
\(
a_{t^\star} \sim \pi_\theta(\cdot \mid u,s_{t^\star},\kappa).
\)

\noindent
\textbf{(2) Without Information Control.}
The policy acts autonomously:
\(
a_t \sim \pi_\theta(\cdot \mid u,s_t).
\)

\paragraph{Update Schedule.}
We adopt an annealed \emph{control-forcing curriculum} that gradually removes control signals so that the final policy performs reliably without external intervention. Concretely, we schedule $p$ across epochs and optimize under a progressively shifting mixture of the two rollout modes: early training uses frequent control, mid training reduces control, and the final stage removes control entirely.

\paragraph{Reward Design.}
We use a composite reward that preserves an outcome-driven objective while adding auxiliary signals for search behavior:
\begin{equation}
{r}_{\phi}(\tau,y_{\text{gold}})
= r_{\text{correct}}(\tau,y_{\text{gold}}) - r_{\text{penalty}}(\tau),
\end{equation}
where \(r_{\text{correct}}\) is an F1-based outcome reward (with a small format floor \(\lambda_{\text{format}}\) for valid outputs), and \(r_{\text{penalty}}\) penalizes tool-usage violations and control non-compliance (capped by \(\lambda^{\max}_{\text{penalty}}\)). 
Design details and the effect of reward shaping are discussed in \Cref{para:reward_design,para:control_vs_reward_shaping}.

\begin{table*}[t]
    \centering
    \renewcommand{\arraystretch}{0.9}
    \caption{Main results with best performance in bold. $^\dagger/^\star$ represents in-domain/out-domain datasets.}\label{tab:main}
    \resizebox{0.9\textwidth}{!}{
    \begin{tabular}{lcccccccc}
        \toprule
        \textbf{Methods} & \multicolumn{3}{c}{\textbf{General QA}} & \multicolumn{4}{c}{\textbf{Multi-Hop QA}} \\
        \cmidrule{2-9}
         & \textbf{NQ$^\dagger$} & \textbf{TriviaQA$^\star$} & \textbf{PopQA$^\star$} & \textbf{HotpotQA$^\dagger$} & \textbf{2wiki$^\star$} & \textbf{Musique$^\star$} & \textbf{Bamboogle$^\star$} & \textbf{Avg.} \\
        \midrule
        \multicolumn{8}{l}{\textbf{Qwen2.5-7b-Base/Instruct}} \\
        Direct Inference & 0.134 & 0.408 & 0.140 & 0.183 & 0.250 & 0.031 & 0.120 & 0.181 \\
        CoT & 0.048 & 0.185 & 0.054 & 0.092 & 0.111 & 0.022 & 0.232 & 0.106 \\
        IRCoT & 0.224 & 0.478 & 0.301 & 0.133 & 0.149 & 0.072 & 0.224 & 0.239 \\
        Search-o1 & 0.151 & 0.443 & 0.131 & 0.187 & 0.176 & 0.058 & 0.296 & 0.206 \\
        RAG & 0.349 & 0.585 & 0.392 & 0.299 & 0.235 & 0.058 & 0.208 & 0.304 \\
        SFT & 0.318 & 0.354 & 0.121 & 0.217 & 0.259 & 0.066 & 0.112 & 0.207  \\
        R1-base & 0.297 & 0.539 & 0.202 & 0.242 & 0.273 & 0.083 & 0.296 & 0.276  \\
        R1-instruct & 0.270 & 0.537 & 0.199 & 0.237 & 0.292 & 0.072 & 0.293 & 0.271  \\
        Rejection Sampling & 0.360 & 0.592 & 0.380 & 0.331 & 0.296 & 0.123 & 0.355 & 0.348 \\
        Search-R1-base & {0.480} & {0.638} & {0.457} & {0.433} & 0.382 & {0.196} & {0.432} & {0.431}  \\
        Search-R1-instruct & 0.393 & 0.610 & 0.397 & 0.370 & {0.414} & 0.146 & 0.368 & 0.385 \\
        \hdashline
        \specialrule{0em}{1pt}{1pt}
        \Ours & \textbf{0.558} & \textbf{0.682} & \textbf{0.521} & \textbf{0.471} & \textbf{0.439} & \textbf{0.221} & \textbf{0.458} & \textbf{0.479} \\
        \midrule
        \multicolumn{8}{l}{\textbf{Qwen2.5-3b-Base/Instruct}} \\
        Direct Inference & 0.106 & 0.288 & 0.108 & 0.149 & 0.244 & 0.020 & 0.024 & 0.134 \\
        CoT & 0.023 & 0.032 & 0.005 & 0.021 & 0.021 & 0.002 & 0.000 & 0.015 \\
        IRCoT & 0.111 & 0.312 & 0.200 & 0.164 & 0.171 & 0.067 & 0.240 & 0.181 \\
        Search-o1 & 0.238 & 0.472 & 0.262 & 0.221 & 0.218 & 0.054 & \textbf{0.320} & 0.255 \\
        RAG & 0.348 & 0.544 & 0.387 & 0.255 & 0.226 & 0.047 & 0.080 & 0.270  \\
        SFT & 0.249 & 0.292 & 0.104 & 0.186 & 0.248 & 0.044 & 0.112 & 0.176  \\
        R1-base & 0.226 & 0.455 & 0.173 & 0.201 & 0.268 & 0.055 & 0.224 & 0.229  \\
        R1-instruct & 0.210 & 0.449 & 0.171 & 0.208 & 0.275 & 0.060 & 0.192 & 0.224  \\
        Rejection Sampling & 0.294 & 0.488 & 0.332 & 0.240 & 0.233 & 0.059 & 0.210 & 0.265 \\
        
        Search-R1-base & {0.406} & {0.587} & {0.435} & 0.284 & 0.273 & 0.049 & 0.088 & 0.303  \\
        Search-R1-instruct & 0.341 & 0.545 & 0.378 & {0.324} & {0.319} & {0.103} & 0.264 & {0.325}  \\
        \hdashline
        \specialrule{0em}{1pt}{1pt}
        \Ours & \textbf{0.533} & \textbf{0.645} & \textbf{0.512} & \textbf{0.402} & \textbf{0.371} & \textbf{0.118} & 0.298 & \textbf{0.411} \\
        \bottomrule
    \end{tabular}}
\end{table*}

\begin{figure*}[t]
    \centering
    \subfigure[\Ours vs. PPO]{\includegraphics[width=0.24\textwidth]{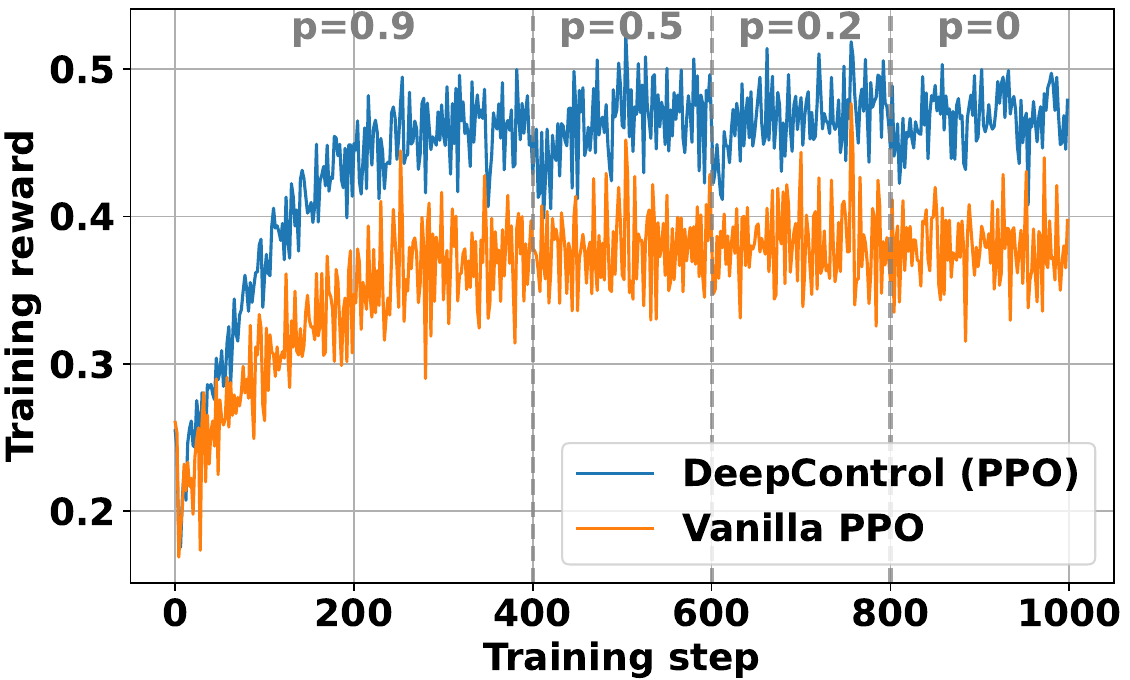}
    }
    \subfigure[PPO vs. GRPO]{\includegraphics[width=0.24\textwidth]{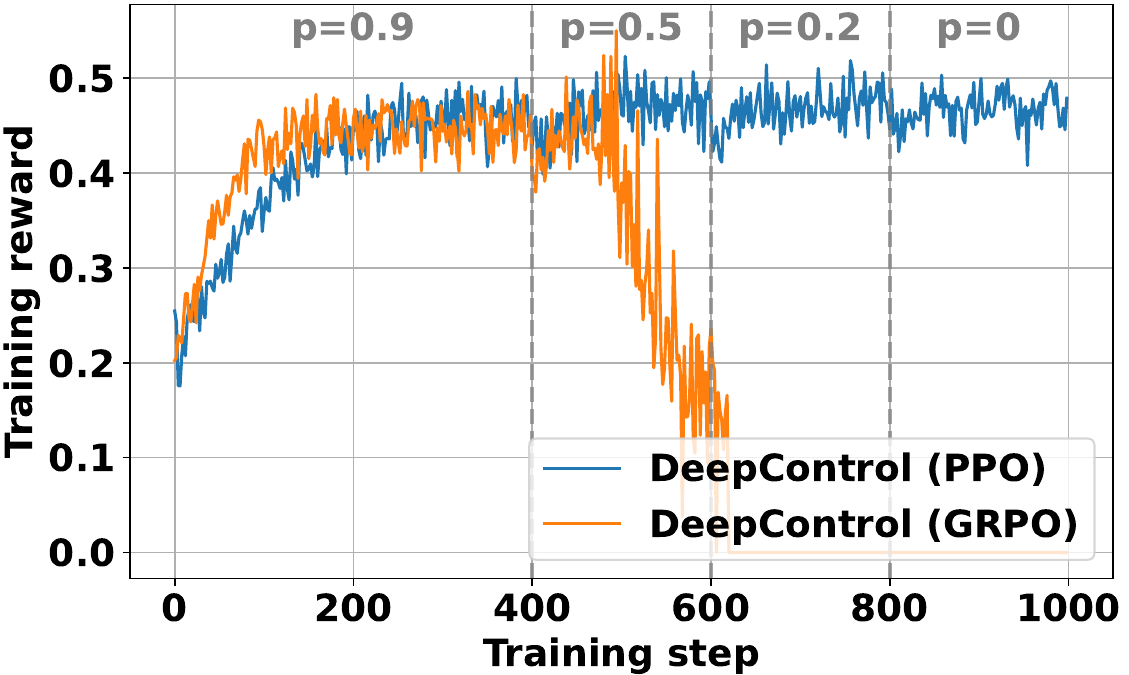}
    }
    \subfigure[Response length]{\includegraphics[width=0.235\textwidth]{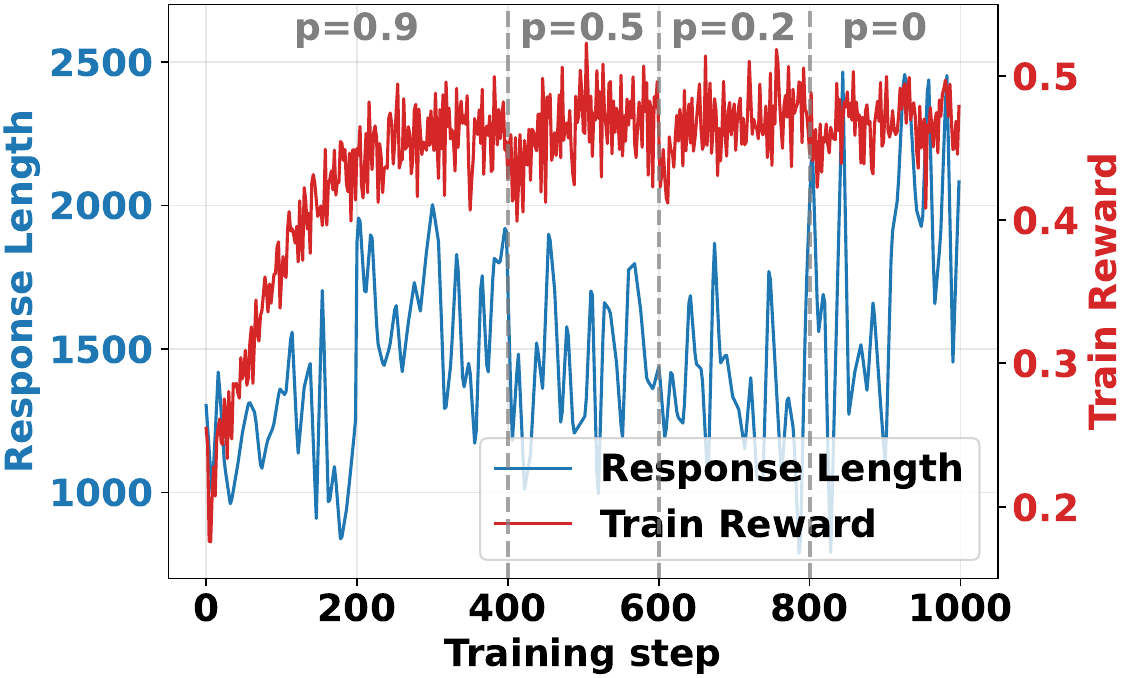}
    }
    \subfigure[Behavior vs. Control]{\includegraphics[width=0.24\textwidth]{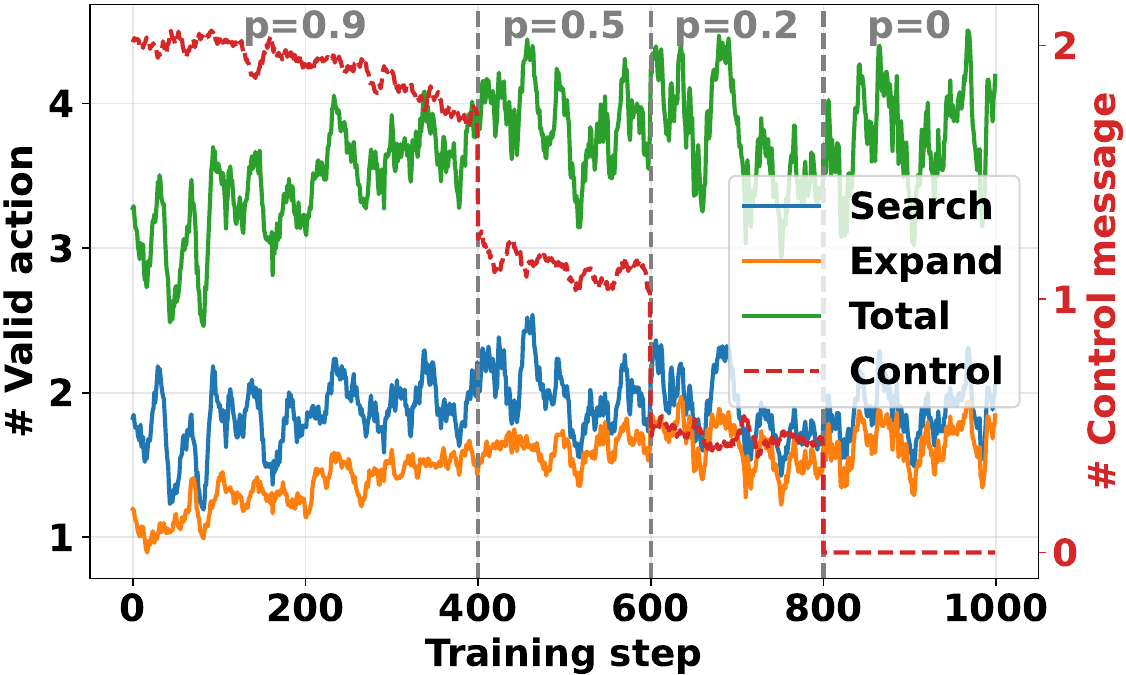}
    }
    \vspace{-5pt}
    \caption{Training dynamics with Qwen2.5-3B-Instruct. (a) \Ours achieves higher training reward than vanilla PPO under the same setup. (b) GRPO suffers from reward collapse, while PPO remains stable. (c) Response length increases early in training and later stabilizes. (d) Search and expand actions increase as the policy internalizes its behavior, while the control messages decrease under the annealing schedule.}
    \label{fig:joint-study}
\end{figure*}

\section{Experiments}

\subsection{Experimental Setup}\label{sec:exp-setup}

\paragraph{Datasets.}
We evaluate \Ours on seven benchmarks:
\emph{General QA} (NQ~\citep{kwiatkowski2019natural}, TriviaQA~\citep{joshi2017triviaqa}, PopQA~\citep{mallen2022not}) and
\emph{Multi-hop QA} (HotpotQA~\citep{yang2018hotpotqa}, 2WikiMultiHopQA~\citep{ho2020constructing}, Musique~\citep{trivedi2022musique}, Bamboogle~\citep{press2022measuring}).

\paragraph{Baselines.}
We compare \Ours against three groups of baselines:
\emph{(i) Inference without retrieval}: Direct inference and CoT~\citep{wei2022chain};
\emph{(ii) Inference with retrieval}: RAG~\citep{lewis2020retrieval}, IRCoT~\citep{trivedi2022interleaving}, and Search-o1~\citep{li2025search};
\emph{(iii) Fine-tuning-based methods}: SFT~\citep{chung2024scaling}, RL without search (R1)~\citep{guo2025deepseek}, rejection sampling with search~\citep{ahn2024large}, and Search-R1~\citep{jin2025search}.
For R1, rejection sampling, and Search-R1, we use the fine-tuned versions from~\citet{jin2025search}.
Across all methods, we use the same retriever, corpus, effective retrieval budget, training data, and pretrained models.

\paragraph{Implementation Details.}
We use Qwen2.5-3B-Instruct~\citep{yang2024qwen2} and Qwen2.5-7B-Instruct as the base models.
For retrieval, we use the 2018 Wikipedia dump~\citep{karpukhin2020dense} with E5~\citep{wang2022text} as the retriever.
Unlike prior retrieval methods~\citep{lin2023ra} that append raw passages, our method uses hierarchical selective expansion while controlling the effective evidence budget for fair comparison.
Specifically, each root contains the passage title and first sentence, and each leaf contains the full passage text. 
Further corpus construction details are described in \Cref{appendix:hierarchical_corpus}. 
Following~\citet{jin2025search}, we train on the merged NQ and HotpotQA training sets using RL, and evaluate on seven benchmarks.
We report Exact Match (EM), following~\citet{yu2024rankrag}.
Additional training details, hyperparameters, and ablations are provided in \Cref{appendix:model_training}.

\begin{table*}[t]
    \centering
    \renewcommand{\arraystretch}{0.85}
    \caption{Ablation study (LLM: Qwen2.5-3B-Instruct; RL algorithm: PPO). We evaluate the impact of different control signals and reward design.}
    \label{tab:ablation}
    \resizebox{\textwidth}{!}{
    \begin{tabular}{lcccccccc}
        \toprule
        \textbf{Method} & \textbf{NQ} & \textbf{TriviaQA} & \textbf{PopQA} & \textbf{HotpotQA} & \textbf{2wiki} & \textbf{Musique} & \textbf{Bamboogle} & \textbf{Avg.} \\
        \midrule
        {\Ours} & \textbf{0.533} & \textbf{0.645} & \textbf{0.512} & \textbf{0.402} & \textbf{0.371} & \textbf{0.118} & \textbf{0.298} & \textbf{0.411}  \\
        \hline
        \multicolumn{8}{l}{Ablation of control signals} \\
        \hline
        \quad w/o Granularity Control
        & 0.470 & 0.580 & 0.440 & 0.340 & 0.310 & 0.080 & 0.230 & 0.364 \\

        \quad w/o Search Continuation Control
        & 0.490 & 0.610 & 0.460 & 0.360 & 0.340 & 0.100 & 0.260 & 0.380 \\

        \quad w/o Control
        & 0.406 & 0.545 & 0.378 & 0.284 & 0.273 & 0.049 & 0.088 & 0.303 \\
        \hline
        \multicolumn{8}{l}{Ablation of reward design} \\
        \hline
        \quad w/o Penalty
& 0.485 & 0.612 & 0.470 & 0.343 & 0.316 & 0.082 & 0.228 & 0.362 \\

\quad w/o Format Floor
& 0.526 & \textbf{0.645} & 0.501 & 0.389 & 0.352 & 0.103 & 0.282 & 0.400 \\

        \bottomrule
    \end{tabular}}
\end{table*}

\begin{figure*}[t]
\centering
\vspace{-8pt}
    \begin{minipage}[t]{0.6\textwidth}
    \vspace{0pt}
        \centering
\subfigure[Utility Distribution]{\includegraphics[width=0.48\linewidth]{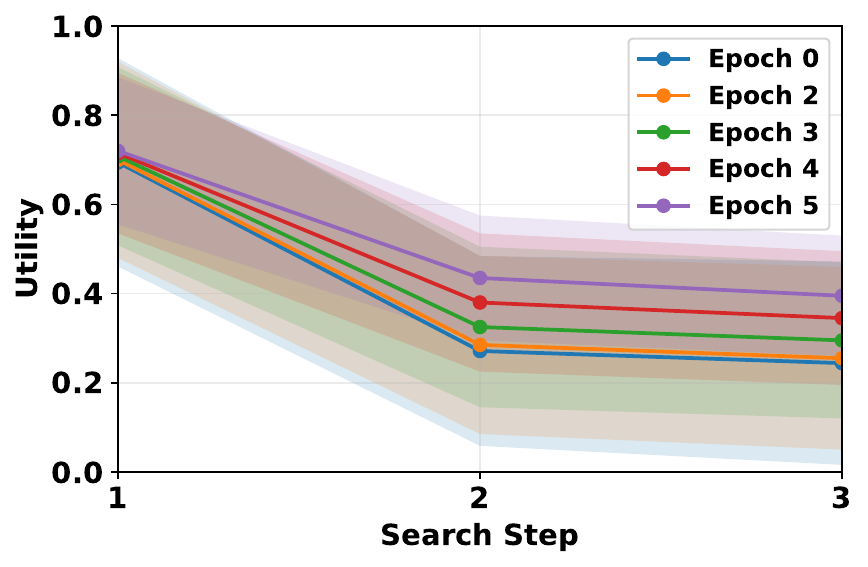}
    }
    \subfigure[Anneal Sensitivity]{\includegraphics[width=0.48\linewidth]{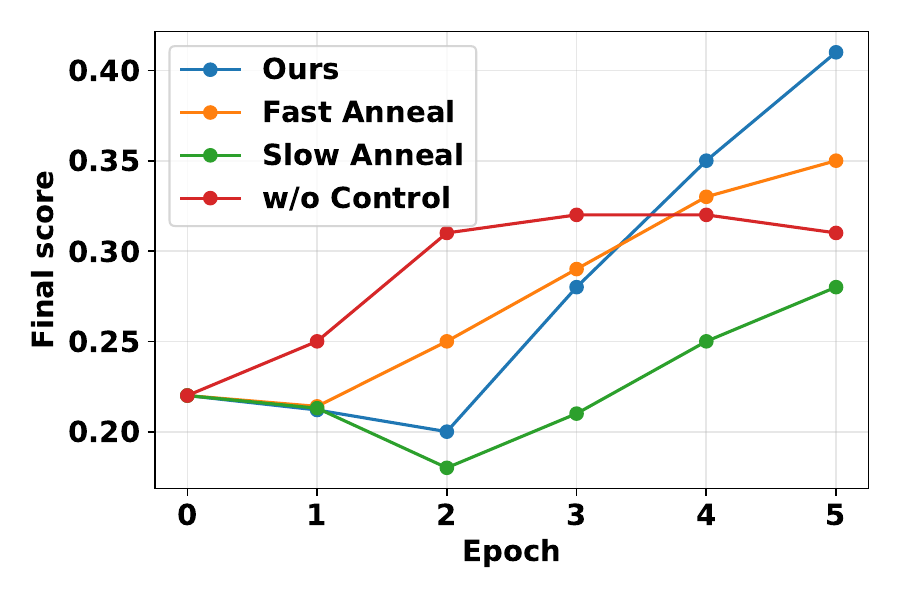}
    }
    \vspace{-7pt}
    \caption{
    Schedule analysis (LLM: Qwen2.5-3B-Instruct; RL algorithm: PPO).
    (a) Per-step utility under no-control evaluation after different annealing stages.
    (b) Final performance under different annealing schedules. See \Cref{para:implementation_schedule_analysis} for implementation details.
    }
    \label{fig:schedule_analysis}

    \end{minipage}
    \hfill
        \begin{minipage}[t]{0.36\textwidth}
    \vspace{0pt}
        \centering
    \includegraphics[width=\linewidth]{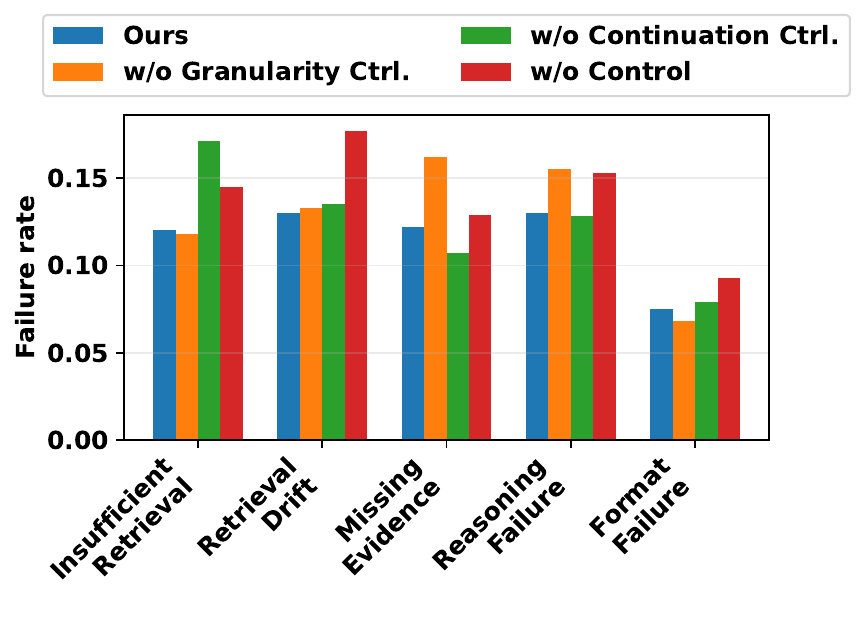}
    \vspace{-25pt}
    \caption{
    Error analysis (LLM: Qwen2.5-3B-Instruct; RL algorithm: PPO; Dataset: HotpotQA).
    See \Cref{para:implementation_error_analysis} for implementation details and result analysis.
    }
    \label{fig:error_analysis}
    \end{minipage}
\end{figure*}

\subsection{Main Results}

\Cref{tab:main} reports results across seven datasets, with qualitative examples in \Cref{appendix:example_op}.
We highlight three observations.
\textbf{(1) \Ours consistently outperforms strong baselines.}
Compared with Search-R1-instruct, \Ours improves average EM by 9.4 points with Qwen2.5-7B and 8.6 points with Qwen2.5-3B, respectively.
\textbf{(2) Information control improves retrieval-based reasoning.}
\Ours outperforms both R1 and Search-R1, showing that effective reasoning requires not only external retrieval but also control over when and how retrieved information is used.
\textbf{(3) The gains are consistent across task types.}
\Ours improves on all benchmarks, suggesting that adaptive information control benefits diverse evidence-seeking settings.

\subsection{Analysis}

\paragraph{Training Dynamics.}
\Cref{fig:joint-study} shows that information control improves optimization and learned search behavior.
Compared with vanilla PPO, \Ours achieves higher training rewards under the same setup.
We use PPO by default because it is more stable under annealed control-forcing, while GRPO exhibits action-format degeneration in our setting (see detailed explanations in \Cref{appendix:ppo_vs_grpo}).
As control messages decrease, the policy performs more search and expansion actions on its own, suggesting progressive internalization of the desired retrieval behavior.

\paragraph{No-Control Behavior and Stopping Quality.}
We further examine whether the acquisition behavior learned under
training-time control transfers to autonomous inference.
As shown in \Cref{tab:no_control_behavior}, \Ours adapts retrieval
depth to task demands: compared with the no-control ablation, it
performs fewer search steps on NQ, but more on the
multi-hop HotpotQA benchmark, suggesting that the
learned policy is neither uniformly conservative nor uniformly
aggressive in retrieval.

We further validate the learned stopping decisions through
counterfactual interventions.
Forcing one additional search step after the policy decides to answer
reduces EM.
Conversely, removing the final search step leads to substantially larger performance drops.
Together, these results indicate that \Ours internalizes meaningful,
task-dependent search continuation behavior and can identify useful
stopping points without external control at test time.

\begin{figure*}[t]
\centering
\includegraphics[width=\linewidth]{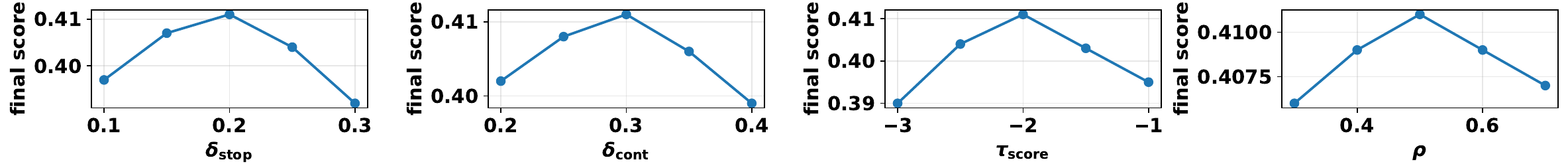}
\begin{minipage}[t]{0.48\textwidth}
\vspace{0pt}
\centering
\includegraphics[width=\linewidth]{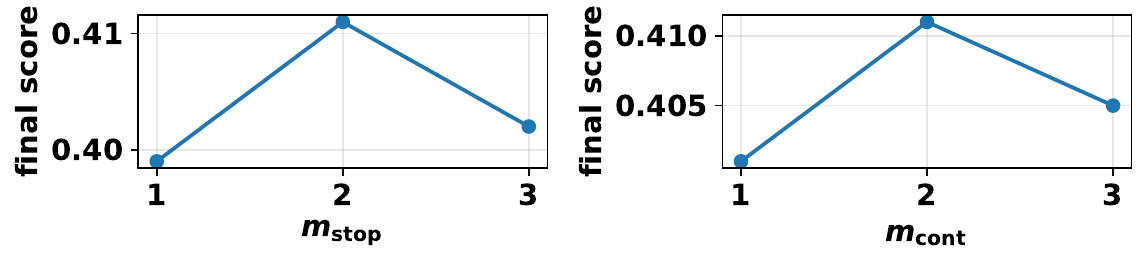}
\vspace{-20pt}
\caption{Control-hyperparameter sensitivity analysis (LLM: Qwen2.5-3B-Instruct; RL Alg: PPO). See \Cref{para:implementation_control_sensitivity} for implementation details and result analysis.}
\label{fig:control_threshold_sensitivity}
\end{minipage}
\hfill
\begin{minipage}[t]{0.48\textwidth}
\vspace{0pt}
\centering
\includegraphics[width=\linewidth]{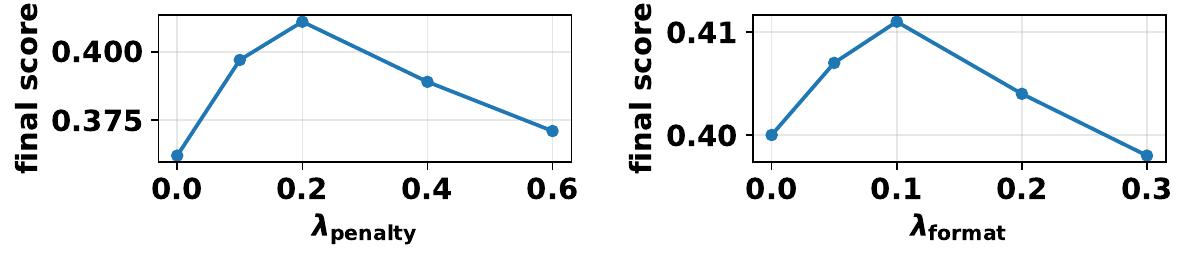}
\vspace{-18pt}
\caption{
Reward-hyperparameter sensitivity analysis (LLM: Qwen2.5-3B-Instruct; RL Alg: PPO).
See \Cref{para:implementation_reward_sensitivity} for implementation details and result analysis.
}
\label{fig:reward_sensitivity}
\end{minipage}
\end{figure*}

\paragraph{Hyperparameter Sensitivity.}
\Cref{fig:control_threshold_sensitivity,fig:reward_sensitivity} show that \Ours is robust to moderate hyperparameter changes.
Intermediate control values work best by balancing premature stopping and unnecessary continuation.
For reward design, performance is most sensitive to the tool-usage penalty.
Alternative utility formulations are further analyzed in
\Cref{tab:utility_ablation}.

\paragraph{Gold-Answer-Free Utility.}
We additionally evaluate a gold-answer-free instantiation based on
belief shift and uncertainty reduction.
It still outperforms Search-R1, indicating that the framework
is not inherently tied to gold-answer-based utility estimation.
Results are reported in \Cref{tab:gold_free_variant}.

\begin{table}[t]
\centering
\small
\caption{
No-control search behavior and stopping quality
(Qwen2.5-3B-Instruct; PPO).
}
\label{tab:no_control_behavior}
\resizebox{\linewidth}{!}{
\begin{tabular}{lccccc}
\toprule
\textbf{Dataset}
& \multicolumn{2}{c}{\textbf{Avg. Search Steps}}
& \multicolumn{3}{c}{\textbf{EM}} \\
\cmidrule(lr){2-3}
\cmidrule(lr){4-6}
& w/o Control & \Ours
& Ori. & Forced Cont. & Trunc. \\
\midrule
NQ
& 2.0 & {1.6}
& \textbf{0.533} & 0.518 & 0.401 \\
HotpotQA
& 1.8 & {2.4}
& \textbf{0.402} & 0.381 & 0.294 \\
\bottomrule
\end{tabular}
}
\end{table}
\begin{table}[t]
\centering
\small
\caption{Disentangling adaptive information control from reward shaping
and hierarchical evidence construction
(Qwen2.5-3B-Instruct; PPO).}
\label{tab:source_ablation}
\begin{tabular}{lc}
\toprule
\textbf{Method} & \textbf{Avg. EM} \\
\midrule
\Ours & \textbf{0.411} \\
Outcome reward + w/ Control & 0.355 \\
Composite reward + w/o Control & 0.303 \\
Hierarchical evidence + outcome reward & 0.311 \\
\quad + w/o Control & \\
Search-R1-instruct & 0.325 \\
\bottomrule
\end{tabular}
\end{table}

\paragraph{Schedule Sensitivity.}
\Cref{fig:schedule_analysis} shows that later checkpoints achieve higher early-step utility under no-control evaluation, suggesting gradual internalization of effective search behavior.
The default annealing schedule achieves the best final no-control performance, while overly fast or slow annealing hurts performance.

\paragraph{Ablation Study.}
\Cref{tab:ablation} shows that search continuation control and granularity control are both important and complementary.
For reward design, the tool-usage penalty has the largest effect among reward-related ablations, highlighting the importance of action-validity regularization.

\paragraph{Disentangling the Sources of Improvement.}
We further examine whether the gains of \Ours arise from adaptive
information control itself, rather than reward shaping or hierarchical
evidence construction.
As shown in \Cref{tab:source_ablation}, retaining information control with an outcome-only reward yields substantially higher average EM than using the composite reward without information control.
Moreover, hierarchical evidence construction without adaptive control underperforms Search-R1.
These results indicate that neither reward shaping nor the hierarchical
interface alone explains the gains; adaptive information control is the
key component that makes hierarchical evidence acquisition effective.

\paragraph{Error Analysis.}
We manually analyze 200 HotpotQA failures and group them into five categories: \texttt{insufficient retrieval}, \texttt{retrieval drift}, \texttt{missing supporting evidence}, \texttt{reasoning failure}, and \texttt{format failure}.
As shown in \Cref{fig:error_analysis}, removing continuation mainly increases \texttt{insufficient retrieval}, while removing granularity mainly increases \texttt{missing supporting evidence}.
Overall, \Ours reduces the main behavioral failure modes.

\paragraph{Robustness Analysis.}
We evaluate \Ours with BM25 to test robustness beyond the main E5 retriever setting.
As reported in \Cref{tab:bm25_results}, \Ours still improves over Search-R1 under BM25 retrieval.
We also discuss the model-family selection and scope of our evaluation in \Cref{para:robustness_analysis}.

\section{Related Work}

\paragraph{Large Language Models with Retrieval.}
Large language models (LLMs) demonstrate strong reasoning and coding abilities~\citep{xiong2024large,yang2024harnessing,yang2025neuro,yang2026stabilizing,he2024give,he2025selfgiveassociativethinkinglimited,li2025schoenfeld,yu2025causaleval,cao2025towards,cao2026pushing,gungordu2026pathwise}, but often suffer from limited factual coverage and hallucinations \citep{zhang2023siren}.
Retrieval-Augmented Generation (RAG) \citep{lewis2020retrieval} addresses this issue by incorporating external documents into the model context, while subsequent work~\citep{cao2025lego,ref:survey,xu2026graphwalker,li2026timem} extends retrieval to both parametric and interactive settings, including REALM \citep{guu2020retrieval}, FiD \citep{izacard2021leveraging}, RETRO \citep{borgeaud2022improving}, and Atlas \citep{izacard2022few}.
Another line of work handles long documents by selecting salient segments before generation~\citep{yang2024compressor, xiong2025long}.
In parallel, tool-based approaches invoke search engines during reasoning~\cite{pei2025scope}, as in IRCoT \citep{trivedi2022interleaving}, ReAct \citep{yao2023react}, Toolformer \citep{schick2023toolformer}, and Search-R1 \citep{jin2025search}.

However, existing retrieval-augmented approaches largely assume that acquiring more information is beneficial and typically append retrieved content to the context using fixed or heuristic strategies.
This often leads to redundant evidence accumulation, context saturation, and noisy reasoning in complex information environments.
In contrast, our work explicitly regulates information acquisition through adaptive control over both retrieval continuation and information granularity.

\paragraph{Reinforcement Learning for LLM Reasoning and Tool Use.}
Reinforcement learning has been widely used to optimize LLMs for complex behaviors such as reasoning and tool use~\citep{li2025verifiable,li2026clawenvkit}.
RLHF~\citep{ouyang2022training} and related methods such as DPO \citep{rafailov2023direct} and other variants~\cite{bao2025exploring} rely on preference-based supervision, while recent studies show that outcome-based RL can induce strong reasoning capabilities using only task-level rewards \citep{shao2024deepseekmath, guo2025deepseek}.
Several works further extend LLM optimization to search- and tool-augmented settings, including WebGPT \citep{nakano2021webgpt}, Toolformer \citep{schick2023toolformer}, TIGER \citep{yang2024can}, and Search-R1 \citep{jin2025search}.
Despite these advances, most approaches rely primarily on sparse outcome-level rewards, which provide limited guidance for intermediate decisions such as whether to continue retrieval or how much information to acquire.
As a result, agents often exhibit brittle behaviors, including premature stopping, over-retrieval, and unreliable tool use.

\paragraph{Information Control in Search-Augmented Reasoning.}
Effective exploration and resource allocation are central challenges in sequential decision-making.
Prior work uses intrinsic rewards, count-based exploration, and curiosity-driven objectives to encourage novelty and state coverage \citep{bellemare2016unifying, pathak2017curiosity}, while rational meta-reasoning and adaptive computation study how limited decision resources should be allocated \citep{russell1991right,zilberstein2011metareasoning}.
In the context of LLMs, researchers have explored exploration strategies and process-level supervision to improve reasoning diversity and stability \citep{xiong2025deliberate, xiongdeliberate,fu2026counterfactual}.
Recent work on adaptive retrieval and selective context construction further highlights the need to control information acquisition under limited budgets \citep{wang2026webclipperefficientevolutionweb,shao2026lifting}.
However, existing approaches typically address this issue only partially, e.g., through improved retrieval, stopping heuristics, or selective context construction, rather than formulating a unified RL control problem over acquisition \emph{extent} and \emph{resolution}.

\section{Conclusion}

We propose an adaptive information control framework for search-augmented reasoning based on information utility.
The framework regulates information acquisition along two complementary axes, search continuation as control over {extent}, and hierarchical expansion as control over {resolution}, and, through annealed control-forcing during online reinforcement learning, enables the model to internalize more effective information acquisition behavior without external intervention at test time.
Experiments across multiple tasks and datasets show consistent gains in reasoning accuracy, training stability, and evidence utilization.

\section*{Limitations}

This work focuses on text-based search-augmented reasoning.
We evaluate on question answering benchmarks across multiple datasets and model settings, but do not study broader settings such as dynamic corpora, multi-tool agents, multi-agent~\cite{zhang2025agentrouter,zhang2026mapro,shi2026ng}, multilingual~\cite{sun2025causalabstain} or multimodal reasoning~\citep{yang2026survey,li2026kg,li2026mrmad}.
Extending adaptive information control to these broader settings is left for future work.

Like other retrieval-augmented language models, our framework may still propagate errors from retrieved evidence or learn suboptimal retrieval behavior from imperfect training signals.
We do not study deployment in high-stakes domains, and such use would require additional safeguards and evaluation.

\section*{Acknowledgments}
This work is supported in part by DARPA SciFy program, Award No.HR001125C0302.

\bibliography{custom}

\newpage

\appendix

\section{Information Utility}
\label{appendix:info_utility}

The value of external information acquisition is inherently \emph{state-dependent}
and must be assessed relative to the agent’s current reasoning state.
We formalize this notion through \emph{information utility},
which measures the marginal value of newly acquired information
for the downstream task.

As described in \Cref{sec:info_utility}, information acquisition is organized at the level of
\emph{search steps}.
We distinguish between two levels of indexing:
let \( t = \{0,1,\ldots,T-1 \}\) index primitive actions
(e.g., \texttt{retrieve}, \texttt{expand}, \texttt{answer}),
and let \( l = \{0,1,\ldots,L-1\} \) index search steps,
each corresponding to a single retrieval event.
Let \( t_l \) denote the primitive step at which the \(l\)-th retrieval is executed.
The \(l\)-th search step starts at \(t_l\) and includes the retrieval action
together with all subsequent expansion actions
until the next retrieval or termination.
Let \(t_{l+1}\) denote the primitive step of the next retrieval (or the termination boundary),
so that all expansions triggered by the \(l\)-th retrieval are completed by step \(t_{l+1}-1\).

Let \( u \) denote the task, and let \( s_{t_l} \) denote the agent’s reasoning state
immediately before executing the \(l\)-th retrieval.
We denote by \( \mathcal{C}_t \) the \emph{injected node set} in the agent context
after primitive step \(t\), which may include both internal nodes and leaf nodes under hierarchical granularity control.

\paragraph{Retrieval Output vs. Injected Evidence.}
Under granularity control, retrieval exposes a \emph{hierarchical evidence structure},
while expansions determine which nodes are actually injected into the context.
We denote by \( e_l \) the \emph{retrieval output} at the \(l\)-th search step:
\begin{equation}
e_l \triangleq \{\mathcal{G}_l^{(i)}\}_{i=1}^{k}, \quad
\mathcal{G}_l^{(i)}=(\mathcal{V}_l^{(i)},\mathcal{E}_l^{(i)}),
\end{equation}
where each retrieved source is a rooted tree with node set \(\mathcal{V}_l^{(i)}\) and directed refinement edges \(\mathcal{E}_l^{(i)}\).

Expansions triggered by the \(l\)-th retrieval inject a subset of nodes from the retrieved hierarchies into the context,
causing the injected set \(\mathcal{C}_t\) to grow during the interval \(t\in[t_l,\,t_{l+1}-1]\).
We quantify the \emph{net injected nodes} contributed by the \(l\)-th search step as the set difference
\begin{equation}
\Delta \mathcal{C}_l \triangleq \mathcal{C}_{t_{l+1}-1} \setminus \mathcal{C}_{t_l-1}.
\label{eq:deltaC}
\end{equation}
By construction, \(\Delta\mathcal{C}_l\) captures the aggregate information injected due to the \(l\)-th retrieval and its subsequent expansions,
abstracting away intermediate refinement states.

We additionally define the \emph{retrieved leaf pool} for novelty computation as
\begin{equation}
\tilde{\mathcal{L}}_l \triangleq \mathrm{Leaves}(e_l),
\quad
\tilde{\mathcal{L}}_{<l} \triangleq \bigcup_{j<l}\tilde{\mathcal{L}}_j,
\end{equation}
i.e., \(\tilde{\mathcal{L}}_l\) contains \emph{all} leaf nodes in the retrieved hierarchies at search step \(l\), regardless of whether they are injected.

Since injected nodes are selected from the retrieved hierarchies, we have
\(
\Delta \mathcal{C}_l \subseteq \bigcup_{i=1}^k \mathcal{V}_l^{(i)}
\),
and the injected leaf nodes are a subset of the retrieved leaf pool:
\(
\mathrm{Leaves}(\Delta \mathcal{C}_l) \subseteq \tilde{\mathcal{L}}_l
\).

\paragraph{Information Utility.}
We define the information utility of the \(l\)-th search step as
\begin{equation}
\begin{aligned}
U(e_l) &= \rho \cdot \mathrm{Novelty}(e_l \mid s_{t_l}) + \\
       & (1-\rho) \cdot \mathrm{Effectiveness}(e_l \mid u,s_{t_l})
\end{aligned}
\label{eq:utility}
\end{equation}
where \( \rho \in [0,1] \) balances the contribution of novelty and effectiveness.
Concretely, we instantiate $\mathrm{Novelty}(e_l \mid s_{t_l})\triangleq\mathrm{Novelty}(\tilde{\mathcal{L}}_l\mid \tilde{\mathcal{L}}_{<l})$
and $\mathrm{Effectiveness}(e_l \mid u,s_{t_l})\triangleq\mathrm{Effectiveness}(\Delta\mathcal{C}_l\mid u,s_{t_l})$.
This design decouples \emph{coverage} (novelty over the full retrieved leaf pool) from \emph{impact}
(effectiveness of what is actually injected), enabling the controller to detect redundant retrieval
even when the agent chooses not to expand those leaves.

\paragraph{Novelty.}
Under hierarchical granularity control, retrieved information is organized as a multi-resolution tree,
where internal nodes correspond to coarse representations (e.g., document or section summaries)
and leaf nodes correspond to fine-grained evidence units that contain concrete factual content (e.g., paragraphs).
We define novelty at the level of leaf nodes, and compute it over the \emph{entire} leaf pool returned by retrieval.

Each leaf node is embedded into a shared semantic space using the E5 encoder~\citep{wang2022text}.
For each newly retrieved leaf node \( v \in \tilde{\mathcal{L}}_l \),
we identify its \(k_{\mathrm{nn}}\) nearest neighbors among leaf nodes retrieved in prior search steps,
denoted by \( \tilde{\mathcal{L}}_{<l} \), and compute the average cosine similarity
\begin{equation}
\text{sim}(v)
=
\frac{1}{k_{\mathrm{nn}}}
\sum_{v^{\prime} \in \mathrm{KNN}\,(v,\, \tilde{\mathcal{L}}_{<l},\, k_{\mathrm{nn}})}
\cos(v, v^{\prime}),
\end{equation}
which estimates the degree to which the content of \(v\) overlaps with previously retrieved evidence.
We define the novelty of leaf node \( v \) as
\begin{equation}
\mathrm{Novelty}(v) = 1 - \text{sim}(v),
\end{equation}
and aggregate novelty across the search step by averaging over the retrieved leaf pool:
\begin{equation}
\mathrm{Novelty}(\tilde{\mathcal{L}}_l \mid \tilde{\mathcal{L}}_{<l})
=
\frac{1}{|\tilde{\mathcal{L}}_l|}
\sum_{v \in \tilde{\mathcal{L}}_l}
\bigl(1 - \text{sim}(v)\bigr).
\end{equation}
By restricting novelty evaluation to leaf nodes, this formulation measures redundancy at the level of concrete evidence,
while avoiding spurious similarity between fine-grained content and coarse summaries.

\paragraph{Effectiveness.}
While novelty captures whether newly retrieved information introduces previously unseen content,
effectiveness measures whether the information injected by expansions is \emph{helpful} for solving the task,
i.e., whether it increases the model's likelihood of a correct answer.
Unlike novelty, effectiveness is computed with respect to the net injected nodes contributed by the search step,
\(\Delta \mathcal{C}_l\), which may include both internal and leaf nodes.

Let \(\mathcal{Y}^{*}(u)\) denote the set of acceptable gold answer strings (aliases) for task \(u\).
To isolate the effect of injected evidence from stochastic variations in reasoning,
we condition the language model on the task \(u\), the injected evidence, and a fixed reasoning trace \(c\),
where \(c\) is generated via deterministic decoding under each evidence condition.
Concretely, let \(\mathcal{C}_{t_{l+1}-1}\) denote the injected evidence accumulated up to the end of the \(l\)-th search step.
For each target string \(y \in \mathcal{Y}^*(u)\), we compute a length-normalized mean log-likelihood:
\begin{equation}
s_l(y)
=
\frac{1}{|y|}
\sum_{i=1}^{|y|}
\log \mathbb{P}\!\left(y_i \mid y_{<i},\, u,\, \mathcal{C}_{t_{l+1}-1},\, c\right),
\end{equation}
where \(|y|\) is the number of tokens in \(y\).
We aggregate across aliases using log-mean-exp:
\begin{equation}
S_l
=
\log\!\left(
\frac{1}{|\mathcal{Y}^*(u)|}
\sum_{y \in \mathcal{Y}^*(u)} \exp\!\big(s_l(y)\big)
\right).
\end{equation}

Effectiveness is defined as the \emph{positive improvement} in this target score induced by the newly injected evidence of the \(l\)-th search step:
\begin{equation}
\Delta_l
=
\max\!\big(0,\; S_l - S_{l-1}\big).
\end{equation}
To obtain a bounded score, we rescale \(\Delta_l\) to \([0,1]\) using two thresholds \(\tau_{\text{low}}<\tau_{\text{high}}\):
\begin{equation}
\begin{aligned}
& \mathrm{Effectiveness}(\Delta \mathcal{C}_l \mid u,s_{t_l}) = \\
& \begin{cases}
0, & \Delta_l \le \tau_{\text{low}},\\
\dfrac{\Delta_l-\tau_{\text{low}}}{\tau_{\text{high}}-\tau_{\text{low}}}, & \tau_{\text{low}} < \Delta_l < \tau_{\text{high}},\\
1, & \Delta_l \ge \tau_{\text{high}}.
\end{cases}
\end{aligned}
\end{equation}
By construction, effectiveness is high only when newly injected evidence increases the model's confidence on the gold answer,
and is zero when the evidence decreases or does not improve it.
Note that this effectiveness signal is used only during \emph{training}, when gold answers are available.

When gold answers are \emph{unavailable}, alternative effectiveness signals can be derived from weaker proxies of answer confidence, such as prediction entropy, KL-based confidence change, self-consistency, or verifier-based scores. 
We investigate one such gold-answer-free instantiation based on distribution shift and uncertainty reduction in \Cref{para:gold_free_utility_estimation}.

We illustrate how novelty, effectiveness, and utility evolve with additional evidence in \Cref{fig:novelty_effectiveness_utility_curves} (see \Cref{appendix:retrieval_continuation_control} for hyperparameters used in our paper).
While novelty rapidly decreases after the first retrieval step, effectiveness remains non-zero in later steps, suggesting that later evidence is often less novel but still useful for improving answer confidence.

\paragraph{Properties.}
The proposed information utility satisfies the following intuitive properties
under our definitions:
\begin{enumerate}[label=\arabic*), left=5pt]
\item \textbf{Monotonicity with novel and beneficial evidence.}
When newly retrieved evidence is both novel with respect to the current reasoning state
and increases the model's confidence on the gold answer (i.e., yields positive effectiveness),
the information utility increases accordingly.
Conversely, evidence that is redundant or does not improve the gold-answer likelihood yields little utility gain.

\item \textbf{Diminishing returns after task completion.}
After sufficient evidence for solving the task has been acquired,
additional retrievals tend to be increasingly redundant and provide only limited improvement
to the gold-answer likelihood, leading to diminishing marginal utility.
\end{enumerate}
Alternative utility formulations, including gold-answer-free variants, are discussed and evaluated in \Cref{para:alternative_utility_formulations}.

\paragraph{Discussion.}
We use information utility as an \textbf{external control signal},
rather than incorporating it directly into the RL reward.
This distinguishes \emph{explicit} regulation of information acquisition (via control messages that can intervene at specific steps)
from \emph{implicit} learning of such behaviors through reward shaping.

This design choice is motivated by:
1) separating utility estimation from policy optimization makes the framework modular, allowing the controller and utility definition to be iterated or replaced without changing the underlying RL objective or training pipeline;
2) optimizing the agent policy primarily for process and outcome correctness empirically leads to simpler and more stable RL training.

\section{Adaptive Information Control}

\subsection{Granularity Control via Hierarchical Selective Expansion}
\label{app:granularity_control}

In real-world settings, retrieved information can be voluminous and lengthy, making full-content injection difficult to manage and often unnecessary in the agent context.
Moreover, fine-grained details are not uniformly useful across reasoning stages.
We therefore introduce \emph{granularity control},
which presents retrieval results at a coarse level first
and allows the agent to selectively expand higher-granularity information
only when needed.

Under granularity control, \textbf{retrieval and information refinement are decoupled}:
the agent first retrieves coarse-grained information via \texttt{retrieve}, and then
selectively refines it through explicit \texttt{expand} actions.
Formally, we model external information as a hierarchical structure (\Cref{fig:information_control}).
At search step $l$, the search engine returns a set of $k$ sources
$e_l=\{\mathcal{G}_l^{(i)}\}_{i=1}^{k}$, where each source is represented as a rooted tree
\(
\mathcal{G}_l^{(i)}=(\mathcal{V}_l^{(i)}, \mathcal{E}_l^{(i)}).
\)
Each node \( v \in \mathcal{V}_l^{(i)} \) corresponds to an evidence unit at a particular resolution,
and each directed edge \( (v, v') \in \mathcal{E}_l^{(i)} \) indicates that $v'$ is a refinement of $v$.

After retrieval, instead of injecting all leaf-level content, we initialize by appending only the retrieved root nodes to the current context; the resulting injected set is denoted by
\(\mathcal{C}_{t_l}\).
The agent may then perform a variable number of \texttt{expand} actions to incrementally grow the observed set
until the next retrieval (or termination).
Let $t_{l+1}$ denote the primitive step of the next retrieval (or termination boundary), so that all expansions
triggered by the $l$-th retrieval are completed by step $t_{l+1}-1$.
The injected nodes satisfy
\(
\mathcal{C}_{t_l} \subseteq \mathcal{C}_{t_l+1} \subseteq \cdots \subseteq \mathcal{C}_{t_{l+1}-1}
\subseteq \bigcup_{i=1}^{k}\mathcal{V}_l^{(i)},
\)
and are expanded adaptively as needed.
The net injected nodes contributed by search step $l$ are
\(
\Delta \mathcal{C}_l = \mathcal{C}_{t_{l+1}-1} \setminus \mathcal{C}_{t_l-1},
\)
where $\mathcal{C}_{t_l-1}$ is the injected set right before the $l$-th retrieval.

For \(t' \in \{t_l+1, \ldots, t_{l+1}-1\}\), an expansion action at primitive step \(t'\) is defined as
\(
a_{t'} = (h_{t'}, \alpha_{t'}, \xi_{t'})
\),
where $h_{t'}$ denotes the agent’s thought, $\alpha_{t'} = \texttt{expand}$, and the action parameters
$\xi_{t'} \subseteq \bigcup^{k}_{i=1}\mathcal{E}_l^{(i)}$ specify a set of hierarchy edges \( (v, v') \)
such that $v\in\mathcal{C}_{t'-1}$ and $v'$ is a child of $v$ in the corresponding tree.
Executing $a_{t'}$ updates
\(
\mathcal{C}_{t'}
=
\mathcal{C}_{t'-1}
\;\cup\;
\{\, v' \mid v \in \mathcal{C}_{t'-1},\, (v, v') \in \xi_{t'} \,\},
\)
i.e., newly expanded nodes are added to the observed evidence set.

During training, given the retrieved hierarchies $e_l$, we derive the expansion targets
$\{\mathcal{C}^\ast_{t_l+1}$, $\ldots$, $\mathcal{C}^\ast_{t_{l+1}-1}\}$ using the information utility signal \(U(\cdot)\),
and use them to guide the agent’s expansion decisions.
Concretely, we score all leaf nodes in the retrieved trees and select the top-$k_{\text{expand}}$ leaves.
We then trace these leaves upward, collecting their ancestors layer by layer until reaching the root,
which yields the target observed evidence sets
$\{\mathcal{C}^\ast_{t_l+1},\ldots,\mathcal{C}^\ast_{t_{l+1}-1}\}$.
Given this target, the controller provides explicit guidance in the form of desired expansion edges
$\xi^\ast_{t'}$ for $t'\in\{t_l+1,\ldots,t_{l+1}-1\}$, so that the induced updates follow
\begin{equation}
\mathcal{C}^\ast_{t'}
=
\mathcal{C}^\ast_{t'-1}
\;\cup\;
\{\, v' \mid v \in  \mathcal{C}^\ast_{t'-1}, \, (v,v')\in\xi^\ast_{t'} \,\}.
\end{equation}
The model is trained to select expansion actions aligned with $\xi^\ast_{t'}$, thereby learning a granularity-control policy
that prioritizes high-utility information while minimizing context growth.

\subsection{Search Continuation Control}
\label{app:retrieval_control}
By default, the agent autonomously decides whether to search based on its internal reasoning state.
However, this decision is often suboptimal:
the agent may terminate search prematurely by underestimating the value
of additional information, or overcommit to continued search when no further
useful evidence is available.
We therefore model \emph{search continuation} as an explicit control decision,
where external intervention is applied \emph{only} when utility signals indicate
systematic misjudgment (\Cref{fig:information_control}).

\paragraph{Termination.}
If the information utility remains below a threshold \( \delta_{\text{stop}} \) for
\( m_{\text{stop}} \) consecutive search steps, we define the stopping index
\begin{equation}
l^\star
=
\min_{l \in [m_{\text{stop}}-1, L-1]} 
\max_{j \in [l-m_{\text{stop}}+1,l]} U(e_j) < \delta_{\text{stop}}.
\end{equation}
Upon reaching \( l^\star \), a control signal
\( \kappa = \texttt{Stop searching} \)
is injected, explicitly terminating further search steps.

\paragraph{Continuation.}
Conversely, the agent may attempt to terminate search and proceed to answer generation even when additional evidence is still beneficial.
We trigger a one-shot continuation intervention when (i) the utility of the most recent \(m_{\text{cont}}\) search steps remains consistently high ($\ge\delta_{\text{cont}}$), but (ii) the model is still insufficiently confident on the gold answer under the current evidence.
Concretely, let \(S_l\) denote the aggregated target score (defined in \Cref{appendix:info_utility}) computed under evidence \(\mathcal{C}_{t_{l+1}-1}\).
If the agent attempts to terminate at search-step index \(l\) and
\begin{equation}
\label{eq:continue_trigger_2}
S_l \le \tau_{\text{score}}
\;\wedge\;
\min_{j \in [l-m_{\text{cont}}+1,l]} U(e_j)\ge \delta_{\text{cont}},
\end{equation}
we inject a one-shot control signal
\(\kappa\) = \texttt{Continue the search for one additional step}.
Here \(\tau_{\text{score}}\) is a confidence threshold on the gold-answer score.
Note that \Cref{eq:continue_trigger_2} is used only during \emph{training} when gold answers are available.

\paragraph{Discussion.}
Under this setting, search continuation is primarily governed
by the agent’s learned policy,
while information utility serves as a monitoring signal
that triggers corrective control when necessary.
Detailed hyperparameter settings and ablations are provided in \Cref{appendix:retrieval_continuation_control}.

\subsection{Reinforcement Learning with Information Control}
\label{app:RL_with_info_control}

Agents can use \emph{external control signals} to improve exploration and stabilize early-stage learning (\Cref{fig:retrieval_with_reasoning_with_control}), but the acquired strategies need be internalized into model parameters to enhance intrinsic capabilities at test time.
To this end, we propose two rollout modes under an annealed \emph{control-forcing} RL scheme, and introduce a composite reward that combines answer correctness and tool-usage regularization.

\paragraph{Rollout Modes.}
During rollouts, the agent samples between two modes, selecting mode (1) with probability $p$ and mode (2) with $1-p$.

\noindent
{(1) With Information Control.}
For each task $u$, a controller monitors the utility of retrieved information throughout the rollout.
Upon detecting an abnormal retrieval pattern, the controller triggers a control signal $\kappa$ at time $t^\star$.
Conditioned on the current reasoning state $s_{t^\star}$ and the triggered control signal
$\kappa$, the policy generates the next action as
${a}_{t^\star} \sim \pi_\theta(\cdot \mid u,s_{t^\star}, \kappa)$.

\noindent
{(2) Without Information Control.}
For each task $u$, at each step $t$, the policy $\pi_\theta$ generates thoughts and actions conditioned only on the current state $s_t$ and task: ${a}_{t} \sim \pi_\theta(\cdot \mid u,s_t)$.

\noindent
The prompts corresponding to the two rollout modes are provided in \Cref{appendix:Prompts}.

\paragraph{Update Modes.}

We adopt an annealed \emph{control-forcing curriculum} that gradually removes control signals so that the final policy performs reliably without external intervention.
Concretely, we schedule $p$ across epochs and optimize under a progressively shifting mixture of the two rollout modes: early training uses frequent control, mid training reduces control, and the final stage removes control entirely.
Within each stage, rollouts are generated by the current policy under the corresponding observation regime (i.e., the control signal, when present, is included in the context), and we perform \emph{on-policy} updates with respect to that regime.
Compared with vanilla RL, this curriculum improves stability in early training when the agent is not yet able to produce effective rollouts without guidance, while ensuring that the learned behavior transfers to the no-control setting at convergence.

\paragraph{Reward Design.}
\label{para:reward_design}
For online RL, reward design is critical, as the learning process is directly driven by reward signals.
Motivated by this property, we design a composite reward that integrates answer correctness and tool-usage regularization, providing informative learning signals for search behavior while preserving an outcome-driven reinforcement learning objective.
Building upon outcome rewards based on F1 score, we incorporate explicit penalties for improper tool usage.
The final reward for a reasoning trajectory \( \tau \) is defined as
\begin{equation}
    {r}_{\phi}(\tau, y_{\text{gold}}) = r_{\text{correct}}(\tau, y_{\text{gold}}) - r_{\text{penalty}}(\tau),
\end{equation}
where $y_{\text{gold}}$ is the gold answer, and \(\phi\) denotes reward hyperparameters.

The base reward of correctness is
\begin{equation}
\begin{aligned}
& r_{\text{correct}}(\tau, y_{\text{gold}}) \;=\; \\
& \begin{cases}
\max\!\big(\mathrm{F1}(y_{\text{pred}}, y_{\text{gold}}),\; \lambda_{\text{format}}\big),
& y_{\text{pred}}\neq\varnothing, \\[4pt]
0, & \text{otherwise},
\end{cases}
\end{aligned}
\end{equation}
where \( \lambda_{\text{format}} \) is a format floor ensuring that valid outputs receive a non-zero reward.

To discourage improper tool interactions, we introduce a tool-usage penalty
\begin{equation}
\begin{aligned}
r_{\text{penalty}}(\tau) = \min\!\big(\lambda_{\text{penalty}} \cdot N_{\text{penalty}}(\tau), \\
\lambda^{\max}_{{\text{penalty}}}\big),
\end{aligned}
\end{equation}
where \( N_{\text{penalty}}(\tau) \) counts the number of tool-usage violations in the trajectory.
We consider two types of violations: (i) incorrect tool usage, such as issuing malformed inputs; and (ii) control non-compliance, where the agent fails to follow explicit control messages.
Each violation incurs a penalty scaled by \( \lambda_{{\text{penalty}}} \), with the total penalty capped at \( \lambda^{\max}_{{\text{penalty}}} \) to avoid over-penalization.
Detailed hyperparameter settings and ablations are provided in \Cref{appendix:reward_design}.

\section{Dataset Overview}
\label{appendix:dataset_overview}

We evaluate \Ours on two categories of tasks: general question answering and multi-hop question answering. For general question answering, we use Natural Questions (NQ)~\citep{kwiatkowski2019natural}, TriviaQA~\citep{joshi2017triviaqa}, and PopQA~\citep{mallen2022not}. For multi-hop question answering, we evaluate on HotpotQA~\citep{yang2018hotpotqa}, 2WikiMultiHopQA~\citep{ho2020constructing}, Musique~\citep{trivedi2022musique}, and Bamboogle~\citep{press2022measuring}. 

Natural Questions consists of real Google search queries paired with Wikipedia answers annotated by humans (79,168 training and 3,610 test samples). TriviaQA is a large-scale reading comprehension benchmark; we use its 11,313-example test set. PopQA contains 14,267 entity-centric triples designed to measure parametric knowledge coverage on long-tail entities. HotpotQA is a crowdsourced Wikipedia-based multi-hop dataset requiring reasoning across multiple paragraphs (90,447 training and 7,405 development samples). 2WikiMultiHopQA combines structured and unstructured Wikipedia information; we evaluate on its 12,576-example development split. Musique composes single-hop questions into 2–4 hop problems; we use its 2,417-example development set. Bamboogle is a manually curated set of 125 two-hop compositional questions selected because search engines originally answered them incorrectly. Examples from each dataset are provided in \Cref{box:dataset_examples}.

\section{Implementation Details}

\paragraph{Prompts.}
\label{appendix:Prompts}
In \Cref{search_aug_reason_prompt}, we present all prompts used in our framework, including the search-augmented reasoning prompt and the control messages.

\paragraph{Hierarchical Corpus Construction.}
\label{appendix:hierarchical_corpus}

In the main experiments, we use the 2018 Wikipedia dump following prior search-augmented RL baselines.
Since this corpus is passage-level rather than full-document-level, we instantiate the general hierarchical interface as a two-level \textit{extractive} structure.
For each retrieved passage, the root node contains the passage title and first sentence as a lightweight extractive summary, while the leaf node contains the full passage text.
Retrieval initially exposes only root-level summaries to the agent, and an \texttt{expand} action reveals the corresponding full passage.

We also investigated an exploratory three-level variant using a local corpus built on the Wikipedia search API.
In this setting, each returned item is a full Wikipedia page with section headings and paragraphs.
We use LLMs to generate document-level and section-level summaries and to clean retrieved text when it contains formatting artifacts.
This creates a richer hierarchy consisting of document summaries, section summaries, and paragraph-level content.
However, we do not include this variant in the main results because it changes the retrieval corpus relative to prior baselines, making direct comparison less controlled.

\paragraph{Control Hyperparameters.}
\label{appendix:retrieval_continuation_control}
Unless otherwise specified, we use a unified set of control hyperparameters across all tasks (see definitions in \Cref{appendix:info_utility}).
We embed retrieved passages using the E5 encoder (\texttt{intfloat/e5-base-v2}) \citep{wang2022text},
truncate each passage to at most 512 tokens for encoding, and compute novelty via a $k$-NN estimator with $k_{\mathrm{nn}}=5$.
Effectiveness is computed from the positive improvement in the aggregated gold-answer score \(S_l\), using \(\Delta_l=\max(0,S_l-S_{l-1})\).
We rescale \(\Delta_l\) to \([0,1]\) with thresholds \(\tau_{\text{low}}=0.3\) and
\(\tau_{\text{high}}=3.0\).
The target score \(S_l\) is computed under a deterministic reasoning trace with a maximum of 128 CoT tokens.
We combine novelty and effectiveness as utility,
with \(\rho=0.5\).
We stop searching when utility stays below a threshold \(\delta_{\text{stop}}=0.2\) for
\(m_{\text{stop}}=2\) consecutive search steps.
We consider a one-shot continuation intervention when the recent utility remains high but the model is still
insufficiently confident on the gold answer.
Concretely, we require the utility to exceed a high-utility threshold \(\delta_{\text{cont}}=0.3\) for
\(m_{\text{cont}}=2\) consecutive search steps, and additionally require the current gold-answer score
\(S_l\) to be below a confidence threshold \(\tau_{\text{score}}=-2.0\) (note that \(S_l\) is a length-normalized log-likelihood score and is typically negative).
We observe that the overall control behavior is insensitive to moderate variations around these values.

\paragraph{Reward Hyperparameters.}
\label{appendix:reward_design}
Unless otherwise specified, we use a fixed set of reward hyperparameters across all tasks.
The format floor is set to $\lambda_{\text{format}} = 0.1$, ensuring that trajectories
producing validly formatted outputs receive a minimal positive signal, which stabilizes early-stage training without overshadowing answer correctness.
The per-violation tool-usage penalty is set to $\lambda_{\text{penalty}} = 0.2$, with the
maximum penalty capped at $\lambda^{\max}_{\text{penalty}} = 0.4$, preventing excessive penalization from dominating the reward signal in trajectories with multiple violations.
We find training to be robust to moderate variations of these values.

\paragraph{Training Setup.}
\label{appendix:model_training}

We conduct experiments with Qwen2.5-3B-Instruct and Qwen2.5-7B-Instruct~\citep{yang2024qwen2}.
For retrieval, we use the 2018 Wikipedia dump~\citep{karpukhin2020dense} as the knowledge source and E5~\citep{wang2022text} as the retriever.
Unlike prior methods that append raw retrieved passages to the context, our approach uses hierarchical selective expansion.
For fair comparison, following~\citep{lin2023ra}, we set the number of retrieved passages to 3 for all existing retrieval-based baselines.
For our method, we retrieve 5 candidate summaries but cap evidence usage by limiting the agent to at most 3 expansion nodes, matching the effective evidence budget.

For training, following~\citep{jin2025search}, we merge the training sets of NQ and HotpotQA to form a unified dataset for \Ours. 
We adopt PPO as the RL algorithm, as we observed that GRPO leads to training collapse after a few dozen of optimization steps.
We train for 5 epochs in total and anneal the control probability $p$ in stages, using $p = 0.9$, $0.5$, $0.2$, and $0$ for 2, 1, 1, and 1 epochs, respectively.
Evaluation is conducted on the test or validation sets of seven datasets to assess both in-domain and out-of-domain performance. Exact Match (EM) is used as the evaluation metric, following \cite{yu2024rankrag}.
For inference-style baselines, we use instruct models, as base models fail to follow instructions. For RL tuning methods, experiments are conducted on both base and instruct models. 

For the PPO variant of \Ours, we follow the implementation provided in Verl \citep{sheng2024hybridflow} and set the learning rate of the policy model to $1\times10^{-6}$ and that of the value model to $1\times10^{-5}$.
Training is performed with warm-up ratios of 0.1 and 0.015 for the policy and value models, respectively.
We employ Proximal Policy Optimization with Generalized Advantage Estimation (GAE), using $\lambda_{\text{GAE}}=1$ and $\gamma_{\text{GAE}}=1$.

All PPO experiments are conducted on a single node equipped with eight A100 GPUs.
We use a training batch size of 64 per update, with a PPO mini-batch size of 64 and a micro-batch size of 4 for both the policy and value networks.
The maximum prompt length is set to 5,120 tokens, with a maximum response length of 512 tokens.
To reduce GPU memory consumption, we enable gradient checkpointing and employ Fully Sharded Data Parallel (FSDP) training with CPU parameter offloading.

For efficient rollout generation, we adopt vLLM \citep{kwon2023efficient} with a tensor parallel size of 1 and a GPU memory utilization ratio of 0.4.
Rollout sampling uses a temperature of 1.0.
We use an adaptive KL controller with an initial coefficient of $\beta = 0.001$, together with standard PPO clipping.

For GRPO training, we set the policy learning rate to $1\times10^{-6}$.
We sample six responses per prompt and train the model with a warm-up ratio of 0.1.
GRPO experiments are conducted using the same hardware setup, a training batch size of 32, sequence length limits, and rollout configurations as in PPO.
We use a larger explicit KL penalty ($\beta = 0.01$) for improved training stability.
Unless otherwise specified, gradient checkpointing, FSDP offloading, and vLLM-based rollouts share identical hyperparameters across methods.

Model checkpoints are saved every 100 training steps.
If training becomes unstable, we select the most recent stable checkpoint based on the reward curve; otherwise, the final checkpoint is used for evaluation.
Unless stated otherwise, we set the maximum action budget to 8.
PPO is used as the default RL algorithm, with a detailed comparison between PPO and GRPO provided in \Cref{appendix:ppo_vs_grpo}.
All experiments are conducted with a fixed random seed.

\paragraph{Training Cost.}
On a single node with 8 A100 GPUs, compared with vanilla PPO, \Ours increases the average rollout time from 5.2 to 6.7 minutes and the total wall-clock training time for 1000 steps from 3.61 to 4.66 days.
The additional cost is incurred only during training, mainly from utility estimation and controller construction.
At inference time, the final policy operates without external control and therefore introduces no additional runtime overhead.

\begin{table*}[t]
\centering
\caption{Comparison of our method implemented with PPO and GRPO against vanilla PPO (LLM: Qwen2.5-3B-Instruct).}
\label{tab:ours-ppo-grpo}
\resizebox{0.85\textwidth}{!}{
\begin{tabular}{lcccccccc}
\toprule
\textbf{Method} & \textbf{NQ} & \textbf{TriviaQA} & \textbf{PopQA} & \textbf{HotpotQA} & \textbf{2Wiki} & \textbf{Musique} & \textbf{Bamboogle} & \textbf{Avg.} \\
\midrule
Vanilla PPO & 0.432 & 0.518 & 0.413 & 0.307 & 0.293 & 0.094 & 0.237 & 0.328 \\
\Ours (PPO) & \textbf{0.533} & \textbf{0.645} & \textbf{0.512} & \textbf{0.402} & \textbf{0.371} & \textbf{0.118} & \textbf{0.298} & \textbf{0.411} \\
\Ours (GRPO) & 0.362 & 0.438 & 0.348 & 0.271 & 0.254 & 0.081 & 0.202 & 0.279 \\
\bottomrule
\end{tabular}
}
\end{table*}

\section{Additional Results}

\paragraph{\Ours vs. Vanilla PPO.}
\label{appendix:ours_vs_ppo}
We compare \Ours against vanilla PPO without control signals. Both methods are trained using the same data, reward design, and hyperparameter configuration.  
The training dynamics are shown in Figure~\ref{fig:joint-study}(a), and the evaluation results are reported in Table~\ref{tab:ours-ppo-grpo}. \Ours consistently achieves higher performance than vanilla PPO.
The control signals provide corrective guidance during early training, helping the agent avoid suboptimal retrieval behaviors when the policy is still immature. As training progresses, these behaviors are gradually internalized by the policy, allowing the agent to perform effectively even after control signals are removed. On average, \Ours improves EM by 8.3 points over vanilla PPO, demonstrating that information control substantially improves training stability and final performance in online RL.

\paragraph{PPO vs. GRPO Under Information Control.}
\label{appendix:ppo_vs_grpo}

We evaluate \Ours using PPO and GRPO as the underlying RL algorithm. 
The training dynamics are shown in \Cref{fig:joint-study}, and the final results are summarized in \Cref{tab:ours-ppo-grpo}. 
We make three observations. 
First, GRPO improves faster than PPO in early training. 
This may be partly because PPO relies on a learned critic, whose value estimates can be less reliable at the beginning of training.
Second, PPO exhibits greater stability under control annealing. 
As shown in \Cref{fig:joint-study}, GRPO exhibits reward degradation after extended training in our setting, whereas PPO maintains more stable optimization throughout the annealing process.
Third, PPO achieves higher final no-control performance than GRPO, suggesting that it better transfers controlled behavior into autonomous test-time behavior under the current training setup.

This instability is not unique to our framework. 
Search-R1~\citep{jin2025search} also observes that GRPO can converge faster in the early stage but may become less stable after extended training, while PPO provides more stable optimization. 
To better understand this behavior, we inspect training curves, including KL, policy-gradient loss, clip fraction, reward, valid-action rate, number of valid search actions, prompt length, and response length, together with rollout traces and final correctness. 
We observe that the degradation of GRPO is accompanied by spikes in KL and policy-gradient loss, followed by a sharp drop in valid-action rate and valid search actions. 
This suggests that the instability is closely related to action-format degeneration.

A possible reason is that GRPO estimates advantages from relative rewards within a group of responses for the same prompt. 
When most trajectories are incorrect or receive similar auxiliary rewards, the within-group reward contrast becomes weak. 
At the same time, small discrete differences from tool-usage penalties can create noisy high-contrast updates that do not necessarily correspond to better search behavior. 
This issue is amplified in search-augmented reasoning, where valid tool actions are brittle: a small policy shift can turn a valid \texttt{search} or \texttt{expand} action into a malformed action, after which retrieval fails and subsequent rewards become uninformative. 
Annealed control-forcing further changes the rollout distribution from controlled behavior to autonomous behavior, which can stress group-relative advantage estimation.

In contrast, PPO appears to handle annealed control-forcing more stably in our setting because its learned critic provides a smoother baseline across changing rollout regimes. 
As the control probability decreases, PPO can more gradually transfer controlled behavior into autonomous no-control behavior.

However, this finding does not imply that the proposed framework is conceptually tied to PPO. 
The control mechanism modifies the rollout interface and training-time control targets, rather than the policy-gradient objective itself. 
Instead, our results suggest that different RL algorithms may require different stabilization strategies when combined with annealed information-control training. 
We leave a systematic study of GRPO stabilization, such as smoother annealing, stronger KL regularization, and explicit action-format stabilization, to future work.

\begin{table*}[t]
\centering
\caption{Utility-formulation ablation (LLM: Qwen2.5-3B-Instruct; RL algorithm: PPO). We compare the default additive utility against two alternatives while keeping all other training, control, and reward settings fixed.}
\label{tab:utility_ablation}
\resizebox{0.8\linewidth}{!}{
\begin{tabular}{lcccccccc}
\toprule
\textbf{Utility} & \textbf{NQ} & \textbf{TriviaQA} & \textbf{PopQA} & \textbf{HotpotQA} & \textbf{2Wiki} & \textbf{Musique} & \textbf{Bamboogle} & \textbf{Avg.} \\
\midrule
$U(e_l)$ & \textbf{0.533} & \textbf{0.645} & \textbf{0.512} & \textbf{0.402} & \textbf{0.371} & \textbf{0.118} & \textbf{0.298} & \textbf{0.411} \\
$U_{\text{bin}}(e_l)$ & 0.524 & 0.634 & 0.501 & 0.389 & 0.360 & 0.113 & 0.286 & 0.401 \\
$U_{\text{prod}}(e_l)$ & 0.512 & 0.620 & 0.486 & 0.376 & 0.348 & 0.108 & 0.274 & 0.389 \\
\bottomrule
\end{tabular}
}
\end{table*}
\begin{table*}[t]
\centering
\caption{Gold-answer-free variant analysis (LLM: Qwen2.5-3B-Instruct; RL algorithm: PPO). We compare the gold-free instantiation of \Ours with Search-R1 baselines and the default gold-supervised setting.}
\label{tab:gold_free_variant}
\resizebox{0.85\linewidth}{!}{
\begin{tabular}{lcccccccc}
\toprule
\textbf{Method} & \textbf{NQ} & \textbf{TriviaQA} & \textbf{PopQA} & \textbf{HotpotQA} & \textbf{2Wiki} & \textbf{Musique} & \textbf{Bamboogle} & \textbf{Avg.} \\
\midrule
Search-R1-base & 0.406 & 0.587 & 0.435 & 0.284 & 0.273 & 0.049 & 0.088 & 0.303 \\
Search-R1-instruct & 0.341 & 0.545 & 0.378 & 0.324 & 0.319 & 0.103 & 0.264 & 0.325 \\
\Ours & \textbf{0.533} & \textbf{0.645} & \textbf{0.512} & \textbf{0.402} & \textbf{0.371} & \textbf{0.118} & \textbf{0.298} & \textbf{0.411} \\
\Ours (Gold-free) & 0.491 & 0.610 & 0.472 & 0.360 & 0.330 & 0.094 & 0.253 & 0.373 \\
\bottomrule
\end{tabular}
}
\end{table*}

\paragraph{Alternative Utility Formulations.}
\label{para:alternative_utility_formulations}
To assess whether our results depend on a specific utility instantiation, we additionally evaluate two simple alternatives to the default formulation in \Cref{appendix:info_utility}. 
The default utility is an additive combination of novelty and effectiveness,
\begin{align*}
    U(e_l)=&\rho \cdot \mathrm{Novelty}(e_l \mid s_{t_l}) + \\
    &(1-\rho)\cdot \mathrm{Effectiveness}(e_l \mid u,s_{t_l}),
\end{align*}
where effectiveness is defined from the positive score improvement \(\Delta_l=\max(0,S_l-S_{l-1})\) and then rescaled to \([0,1]\) using thresholds \(\tau_{\text{low}}\) and \(\tau_{\text{high}}\).

We compare against the following alternatives.
{(1) Binary-effectiveness utility.}
We replace the rescaled effectiveness term with a binary indicator of whether the newly injected evidence yields a positive target-score improvement:
\[
\mathrm{Effectiveness}^{\mathrm{bin}}(e_l)
=
\mathbb{I}[S_l-S_{l-1}>0].
\]
The resulting utility is
\begin{align*}
    U_{\mathrm{bin}}(e_l)
=&
\rho \cdot \mathrm{Novelty}(e_l \mid s_{t_l})
+\\
&(1-\rho)\cdot \mathrm{Effectiveness}^{\mathrm{bin}}(e_l).
\end{align*}
This variant tests whether the exact continuous shaping of effectiveness is important, or whether a coarse improvement signal is already sufficient.

{(2) Multiplicative utility.}
We also replace the additive combination with a multiplicative interaction:
\begin{align*}
U_{\mathrm{prod}}(e_l)
=
&\mathrm{Novelty}(e_l \mid s_{t_l})
\cdot\\
&\mathrm{Effectiveness}(e_l \mid u,s_{t_l}).
\end{align*}
This variant assigns high utility only when the retrieved evidence is both novel and effective, and therefore tests whether the additive formulation is preferable to a stricter interaction rule.

All other training, control, and reward settings are kept identical to the default configuration. The results in \Cref{tab:utility_ablation} show that the default additive utility performs best overall, while both alternatives remain competitive but are less effective. This suggests that the proposed framework is not tied to a single fragile formulation, while also indicating that continuous effectiveness shaping and additive combination provide the most reliable control signal in our setting.

\paragraph{Gold-answer-free Utility Estimation.}
\label{para:gold_free_utility_estimation}
\Ours is not restricted to gold-answer dependent utility estimation.
To examine whether the proposed principle can be instantiated without gold-answer access, we also evaluate a gold-free variant based on decision-impact proxies.
Specifically, let \(u\) denote the task, \(s_{t_l}\) the current context, \(e_l\) the newly injected evidence, and \(\mathcal{Y}(u)\) the candidate answer set. We compute the model's answer distribution before and after evidence injection:
\begin{equation}
\mathbb{P}_{\mathrm{pre}}(y) = \mathbb{P}(y \mid u, s_{t_l}),
\end{equation}
\begin{equation}
\mathbb{P}_{\mathrm{post}}(y) = \mathbb{P}(y \mid u, s_{t_l}, e_l),
\end{equation}
where $y \in \mathcal{Y}(u)$ denotes a candidate final answer.

We then measure the evidence-induced belief shift using KL divergence:
\begin{equation}
D_{\mathrm{KL}}\!\left(\mathbb{P}_{\mathrm{post}} \,\|\, \mathbb{P}_{\mathrm{pre}}\right)
=
\sum_{y \in \mathcal{Y}(u)} \mathbb{P}_{\mathrm{post}}(y)
\log \frac{\mathbb{P}_{\mathrm{post}}(y)}{\mathbb{P}_{\mathrm{pre}}(y)}.
\end{equation}

We also measure whether the injected evidence reduces answer uncertainty using entropy:
\begin{equation}
H(\mathbb{P}) = - \sum_{y \in \mathcal{Y}(u)} \mathbb{P}(y) \log \mathbb{P}(y),
\end{equation}
\begin{equation}
\Delta H = H(\mathbb{P}_{\mathrm{pre}}) - H(\mathbb{P}_{\mathrm{post}}).  
\end{equation}

The gold-free effectiveness score is then defined as
\begin{equation}
\begin{split}
& \mathrm{Effectiveness}_{\mathrm{gold\text{-}free}}(e_l \mid u,s_{t_l}) = \\
& \alpha D_{\mathrm{KL}}\!\left(\mathbb{P}_{\mathrm{post}} \,\|\, \mathbb{P}_{\mathrm{pre}}\right) + \beta \max(0, \Delta H),
\end{split}
\end{equation}
where \(\alpha, \beta \ge 0\) control the relative weights of distributional change and confidence gain. 

Similarly, for continuation control in \Cref{eq:continue_trigger}, we replace the gold-answer target score with a gold-free distribution-shift measure,
\begin{equation}
S_{\mathrm{gold\text{-}free}}=\left\|\mathbb{P}_{\mathrm{post}} - \mathbb{P}_{\mathrm{pre}}\right\|.
\end{equation}
We evaluate this gold-answer-free variant under the same experimental setup, with results reported in \Cref{tab:gold_free_variant}.
The variant still outperforms Search-R1 baselines. 
This suggests that the gains do not solely depend on gold-answer access and that the proposed framework remains effective when instantiated with gold-free decision-impact signals.

\paragraph{Control-hyperparameter Sensitivity Analysis.}
\label{para:implementation_control_sensitivity}
We study the sensitivity of the main control hyperparameters using Qwen2.5-3B-Instruct with PPO, while fixing all other training and inference settings to the default configuration described in \Cref{appendix:model_training,appendix:retrieval_continuation_control}.
In particular, we keep the retrieval setup, reward design, training data, annealing schedule, action budget, and PPO hyperparameters unchanged, and vary one control hyperparameter at a time.

We examine six hyperparameters in search continuation control: the stop threshold $\delta_{\text{stop}}$, stop patience $m_{\text{stop}}$, continuation threshold $\delta_{\text{cont}}$, score threshold $\tau_{\text{score}}$, continuation patience $m_{\text{cont}}$, and utility mixing weight $\rho$.
Unless it is the target of variation, each hyperparameter is fixed to its default value:
$\delta_{\text{stop}}=0.2$,
$m_{\text{stop}}=2$,
$\delta_{\text{cont}}=0.3$,
$\tau_{\text{score}}=-2.0$,
$m_{\text{cont}}=2$,
and $\rho=0.5$.
Each configuration is trained under the same 5-epoch PPO setup on the merged NQ + HotpotQA training data and evaluated with EM under the same validation protocol as in the main experiments.
The results (\Cref{fig:control_threshold_sensitivity}) show that performance is stable under moderate perturbations around the default setting.
In general, intermediate values yield the best trade-off between under-searching and over-searching:
smaller $\delta_{\text{stop}}$ or larger $m_{\text{stop}}$ tends to delay stopping, while larger $\delta_{\text{stop}}$ or smaller $m_{\text{stop}}$ can lead to premature termination;
similarly, overly permissive or overly conservative continuation settings, controlled by $\delta_{\text{cont}}$, $\tau_{\text{score}}$, and $m_{\text{cont}}$, both reduce final performance.
Overall, these results suggest that the proposed controller is robust and does not rely on narrow hyperparameter tuning.

\begin{table*}[t]
\centering
\caption{Effect of information control and reward shaping (LLM: Qwen2.5-3B-Instruct; RL algorithm: PPO). We compare the default setting with a control-only variant using outcome reward and a no-control variant using the composite reward.}
\label{tab:control_vs_reward_shaping}
\resizebox{0.85\linewidth}{!}{
\begin{tabular}{lcccccccc}
\toprule
\textbf{Method} & \textbf{NQ} & \textbf{TriviaQA} & \textbf{PopQA} & \textbf{HotpotQA} & \textbf{2Wiki} & \textbf{Musique} & \textbf{Bamboogle} & \textbf{Avg.} \\
\midrule
\Ours & {0.533} & {0.645} & {0.512} & {0.402} & {0.371} & {0.118} & {0.298} & {0.411} \\
Outcome reward + w/ Control & 0.482 & 0.601 & 0.465 & 0.342 & 0.330 & 0.090 & 0.174 & 0.355 \\
Composite reward + w/o Control & 0.406 & 0.545 & 0.378 & 0.284 & 0.273 & 0.049 & 0.088 & 0.303 \\
\bottomrule
\end{tabular}
}
\end{table*}

\begin{table*}[t]
\centering
\caption{Effect of hierarchical evidence construction (LLM: Qwen2.5-3B-Instruct; RL algorithm: PPO). We compare the default setting with a hierarchical-evidence-only variant and Search-R1.}
\label{tab:hierarchy_effect}
\resizebox{0.85\linewidth}{!}{
\begin{tabular}{lcccccccc}
\toprule
\textbf{Method} & \textbf{NQ} & \textbf{TriviaQA} & \textbf{PopQA} & \textbf{HotpotQA} & \textbf{2Wiki} & \textbf{Musique} & \textbf{Bamboogle} & \textbf{Avg.} \\
\midrule
\Ours & 0.533 & 0.645 & 0.512 & 0.402 & 0.371 & 0.118 & 0.298 & 0.411 \\
{Hierarchical evidence + outcome}
& \multirow{2}{*}{0.402}
& \multirow{2}{*}{0.535}
& \multirow{2}{*}{0.380}
& \multirow{2}{*}{0.291}
& \multirow{2}{*}{0.278}
& \multirow{2}{*}{0.080}
& \multirow{2}{*}{0.210}
& \multirow{2}{*}{0.311} \\
\; reward + w/o Control
& & & & & & & & \\
Search-R1-instruct & 0.341 & 0.545 & 0.378 & 0.324 & 0.319 & 0.103 & 0.264 & 0.325 \\
\bottomrule
\end{tabular}
}
\end{table*}

\paragraph{Reward-hyperparameter Sensitivity Analysis.}
\label{para:implementation_reward_sensitivity}
We study the sensitivity of the main reward hyperparameters using Qwen2.5-3B-Instruct with PPO, while fixing all other training, retrieval, and control settings to the default configuration described in \Cref{appendix:model_training,appendix:reward_design,appendix:retrieval_continuation_control}.
In each experiment, we vary one reward hyperparameter at a time and keep the remaining reward terms at their default values.
We examine two reward hyperparameters: the per-violation penalty coefficient $\lambda_{\text{penalty}}$ and the format floor $\lambda_{\text{format}}$.
Unless it is the target of variation, we set
$\lambda_{\text{penalty}}=0.2$
and
$\lambda_{\text{format}}=0.1$.

Each configuration is trained under the same 5-epoch PPO setup on the merged NQ+HotpotQA training data and evaluated with EM under the same validation protocol as in the main experiments.
The results show that performance is more sensitive to $\lambda_{\text{penalty}}$, confirming the importance of discouraging malformed tool usage and control non-compliance during training.
In contrast, moderate variations of $\lambda_{\text{format}}$ have a milder effect, suggesting that the format floor mainly serves as an auxiliary stabilization signal.
Overall, these results (\Cref{fig:reward_sensitivity}) indicate that the reward design is robust to moderate hyperparameter variations and does not rely on narrow tuning.

\paragraph{Schedule Sensitivity Analysis.}
\label{para:implementation_schedule_analysis}
We analyze the role of the annealed control-forcing schedule from two perspectives.
First, under the default schedule described in \Cref{appendix:model_training}, we examine how the no-control utility profile evolves across training stages.
As shown in \Cref{fig:schedule_analysis}(a), we evaluate the policy after the completion of each annealing stage, corresponding to checkpoints after training with $p=0.9$, $0.5$, $0.2$, and $0$, and compare them with the no-training baseline.
The results show that later checkpoints exhibit consistently higher utility in the early search steps under no-control evaluation, suggesting that the policy gradually internalizes more effective information-seeking behavior as external control is annealed away.

Second, we compare different annealing schedules in terms of final no-control performance, as shown in \Cref{fig:schedule_analysis}(b).
Our default schedule trains for 5 epochs in total, using control probabilities $p=0.9$, $0.5$, $0.2$, and $0$ for 2, 1, 1, and 1 epochs, respectively.
We compare this schedule against a faster variant, which allocates 1, 1, 1, and 2 epochs to the same four stages and therefore enters the no-control phase earlier, and a slower variant, which allocates 3, 1, 1, and 0 epochs and thus never reaches the fully no-control stage during training.
We also include a no-control baseline trained without external control signals throughout.
The default schedule achieves the best overall performance, while overly fast annealing weakens early guidance and overly slow annealing delays policy internalization.
These results support the role of annealed control-forcing in balancing training-time guidance and test-time autonomy.

\begin{table*}[t]
\centering
\caption{Performance comparison with different retrievers (LLM: Qwen2.5-7B; RL algorithm: PPO). We compare \Ours with Search-R1 using E5 and BM25 retrieves.}
\label{tab:bm25_results}
\resizebox{0.85\linewidth}{!}{
\begin{tabular}{lcccccccc}
\toprule
\textbf{Method} & \textbf{NQ} & \textbf{TriviaQA} & \textbf{PopQA} & \textbf{HotpotQA} & \textbf{2Wiki} & \textbf{Musique} & \textbf{Bamboogle} & \textbf{Avg.} \\
\midrule
Search-R1 (E5) & {0.480} & {0.638} & {0.457} & {0.433} & 0.382 & {0.196} & {0.432} & {0.431} \\
\Ours (E5) & \textbf{0.558} & \textbf{0.682} & \textbf{0.521} & \textbf{0.471} & \textbf{0.439} & \textbf{0.221} & \textbf{0.458} & \textbf{0.479} \\
\hdashline
\specialrule{0em}{1pt}{1pt}
Search-R1 (BM25) & 0.341 & 0.607 & 0.322 & 0.404 & 0.370 & 0.137 & 0.280 & 0.352 \\
\Ours (BM25) & \textbf{0.430} & \textbf{0.654} & \textbf{0.391} & \textbf{0.428} & \textbf{0.415} & \textbf{0.162} & \textbf{0.333} & \textbf{0.402} \\
\bottomrule
\end{tabular}
}
\end{table*}

\paragraph{No-Control Behavior and Stopping Quality.}
\label{para:no_control_stopping_quality}

To better characterize agent behavior during no-control inference, we report the average number of search steps.
As shown in \Cref{tab:no_control_behavior}, 
\Ours adapts its search behavior to task difficulty.
For NQ, which often requires less evidence, \Ours performs fewer search steps, suggesting reduced over-search.
For HotpotQA, which often requires multi-hop evidence, \Ours performs more search steps, suggesting better avoidance of premature answering.

We further evaluate whether the learned stopping behavior is appropriate through two counterfactual interventions.
First, in \emph{forced continuation}, after the policy decides to answer, we force it to conduct one additional search step and then re-answer.
Second, in \emph{search truncation}, we remove the final search step and ask the policy to re-answer using the truncated context.
As shown in \Cref{tab:no_control_behavior},
forced continuation reduces accuracy on both datasets, suggesting that the learned policy does not generally stop prematurely.
Conversely, removing the final search step substantially reduces accuracy, indicating that the final retrieval is often useful rather than redundant.
Together, these results provide direct evidence that \Ours internalizes meaningful stopping behavior under no-control inference.

\paragraph{Control vs. Reward Shaping.}
\label{para:control_vs_reward_shaping}

We further examine whether the gains of \Ours primarily come from adaptive information control or from reward shaping.
The no-control ablation, denoted as {Composite reward + w/o Control}, keeps the same reward design but removes the control mechanisms.
In contrast, keeping the control mechanisms while using an outcome-only reward, denoted as {Outcome reward + w/ Control}.
As shown in \Cref{tab:control_vs_reward_shaping}, information control remains beneficial even with an outcome-only objective.
Meanwhile, using the composite reward without control yields substantially lower performance.
These results suggest that the main gains are not merely a consequence of reward shaping; rather, adaptive information control is a key factor in improving search-augmented reasoning behavior.

\paragraph{Effect of Hierarchical Evidence Construction.}
\label{para:hierarchy_effect}
We further examine the effect of hierarchical evidence construction.
To isolate this factor, we evaluate a hierarchical-evidence-only variant that uses the same hierarchical evidence interface but removes adaptive information control.
As shown in \Cref{tab:hierarchy_effect}, this variant underperforms Search-R1, suggesting that hierarchical evidence construction alone does not explain the gains.
When the agent initially observes only summaries and must learn expansion actions from sparse outcome rewards, it cannot reliably learn when to expand evidence without control guidance.
These results indicate that adaptive information control is the key factor that makes hierarchical expansion effective.

\paragraph{Robustness Analysis.}
\label{para:robustness_analysis}

We further examine whether \Ours remains effective under different retrieval backends.
The proposed utility estimation does not require the retriever itself to be dense or differentiable: novelty can be computed by encoding the retrieved evidence with the same evidence encoder, while effectiveness is computed from the LLM's answer likelihood.
Thus, the control framework can in principle be applied on top of different retrieval systems.

To test this, we replace the E5 retriever with BM25 while keeping the utility estimation procedure unchanged.
This setting isolates whether the gains of \Ours rely on a specific retrieval backend.
The results for Qwen2.5-7B-base are reported in \Cref{tab:bm25_results}.
As shown, \Ours still outperforms Search-R1 under BM25 retrieval, suggesting that the proposed control mechanism is not limited to a certain retriever.

For model families, we note that online RL over tool-use trajectories requires a certain level of initial capability.
In particular, the base model should be able to follow structured action formats, generate usable search queries, parse retrieved evidence, maintain reasoning over multi-turn interactions, and tolerate RL updates without severe format collapse.
Our current experiments follow the common setup in recent search-augmented RL work, including Search-R1, which primarily evaluates Qwen models.
This provides a controlled comparison under a widely used setting, while broader evaluation across other model families remains an important direction for future work.

\paragraph{Error Analysis.}
\label{para:implementation_error_analysis}
To better understand the remaining failure modes, we analyze no-control evaluation trajectories by grouping incorrect predictions into five categories:
\emph{insufficient retrieval}, \emph{retrieval drift}, \emph{missing supporting evidence}, \emph{reasoning/evidence-use failure}, and \emph{format/control-flow failure}.
Here, \emph{insufficient retrieval} refers to cases where the agent answers with too little evidence, typically after very few search steps or without sufficient expansion;
\emph{retrieval drift} denotes cases where the retrieved context supports the model prediction but not the gold answer, suggesting that search has been led in an incorrect direction;
\emph{missing supporting evidence} covers cases where the retrieved context does not contain the gold answer or sufficient evidence supporting it;
\emph{reasoning failure} refers to cases where the gold answer or supporting evidence is already present in the retrieved context but the final prediction is still incorrect;
and \emph{format failure} includes malformed answer tags, invalid-action contamination, or trajectories that fail to terminate with a valid final answer.
These categories are defined operationally from the retrieved context of each rollout.
In particular, \emph{missing supporting evidence} indicates that the retrieved context lacks sufficient support for the gold answer, but does not distinguish between retrieval failure and cases where such evidence is absent or difficult to match in the corpus.

For labeling, we first identify format failures from rollout traces using the parser state and final-action validity.
We then identify insufficient-retrieval cases from trajectories that terminate after very limited search and expansion before answering.
For the remaining incorrect cases, we distinguish retrieval-side and reasoning-side failures based on whether the retrieved context contains the gold answer or sufficient supporting evidence.
When the retrieved context supports the model prediction but not the gold answer, we label the case as retrieval drift.
When the retrieved context contains neither the gold answer nor sufficient supporting evidence, we label it as missing supporting evidence.
When the gold answer or supporting evidence is already present in the retrieved context but the final prediction remains incorrect, we label it as reasoning failure.
Borderline cases caused by alias variation or answer granularity are treated separately and are not emphasized in the main taxonomy.

We manually analyze the same 200 sampled questions from HotpotQA across all methods (LLM: Qwen2.5-3B-Instruct; RL algorithm: PPO) to compare their dominant failure modes under a consistent labeling protocol.
As shown in \Cref{fig:error_analysis}, removing continuation mainly increases insufficient-retrieval errors, while removing granularity most strongly increases retrieval misses.
Compared with these ablations, \Ours reduces the major behavioral failure modes overall, although reasoning errors remain the largest residual source of failure.
These observations are consistent with the design of our controller: continuation control primarily mitigates premature stopping, while granularity control improves evidence acquisition and utilization.

\paragraph{Example Outputs.}
\label{appendix:example_op}
In \Cref{examples_our_framework}, we present representative examples of \Ours under settings with and without control signals, evaluated on both single-hop and multi-hop questions.
These examples illustrate how the agent interleaves reasoning with retrieval and selective expansion, while control messages guide the agent to regulate search behavior.
In particular, the examples highlight how continuation and termination controls help avoid unnecessary retrieval steps and support more effective evidence use during reasoning.

\onecolumn

\begin{tcolorbox}[
    colback=gray!15,
    colframe=black,
    title=Example: Search-Augmented Reasoning Tasks,
    sharp corners,
    fonttitle=\bfseries,
    breakable,
    enhanced,
    arc=4mm,
    rounded corners,
    label={box:dataset_examples}
]
\scriptsize
\begin{verbatim}
[NQ]
Question: who has the right of way in international waters
Answer: Neither vessel

Question: how many episodes are in series 7 game of thrones
Answer: seven


[TriviaQA]
Question: Who wrote the novel Evening Class?
Answer: Maeve Binchy

Question: Which item of clothing cost Isadora Duncan her life?
Answer: Scarf- it was caught in a moving car wheel


[PopQA]
Question: What is the capital of Mali?
Answer: Bamako

Question: What is Dominick Bellizzi's occupation?
Answer: jockey


[HotpotQA]
Question: Dixit Dominus is a psalm setting by who, which was a German, later British, baroque composer who spent the bulk of his 
career in London, becoming well known for his operas, oratorios, anthems, and organ concertos?
Answer: George Frideric Handel

Question: Zimbabwe's Guwe Secondary School has a sister school in what New York county?
Answer: Nassau County


[2WikiMultiHopQA]
Question: When did John V, Prince Of Anhalt-Zerbst's father die? 
Answer: 12 June 1516 

Question: Who is older, Aryeh Ben-Eliezer or Jason Pociask?
Answer: Aryeh Ben-Eliezer


[Musique]
Question: Where is Ulrich Walter's employer headquartered?
Answer: Cologne

Question: Who founded the company that distributed the film UHF?
Answer: Mike Medavoy


[Bamboogle]
Question: The main actor of Indiana Jones is a licensed what?
Answer: pilot

Question: How many people died in the second most powerful earthquake ever recorded?
Answer: 131
\end{verbatim}
\end{tcolorbox}

\begin{tcolorbox}[
    colback=gray!15,
    colframe=black,
    title=Example: Suboptimal Search Behavior (Search-R1-qwen2.5-3b-it-grpo on NQ),
    sharp corners,
    fonttitle=\bfseries,
    breakable,
    enhanced,
    arc=4mm,
    rounded corners,
    label={box:error_cases_motivation}
]
\begin{lstlisting}[breaklines=true,basicstyle=\rmfamily\scriptsize,columns=fullflexible,keepspaces=true]
[USER PROMPT]
Answer the given question. You must conduct reasoning inside <think> and </think> first every time you get new information. After reasoning, if you find you lack some knowledge, you can call a search engine by <search> query </search> and it will return the top searched results between <information> and </information>. You can search as many times as you want. If you find no further external knowledge needed, you can directly provide the answer inside <answer> and </answer>, without detailed illustrations. For example, <answer> Beijing </answer>. 
Question: when is the next scandal episode coming out?

[ASSISTANT RESPONSE]
To determine when the next Scandal episode is coming out, I will follow these steps:

1. Identify the current season of Scandal.
2. Find out the date of the last episode of the current season.
3. Determine the date of the next episode of the current season.
4. Provide the date of the next episode if available.

Let's start with the first step.

<search>Scandal current season</search>
<information>
Doc 1(Title: "Scandal (TV series)") ABC renewed ""Scandal"" for a sixth season, which premiered on January 26, 2017. The entire season is about the murder of President-elect Frankie Vargas. It is revealed that his wife, Luna, was behind his murder because she wanted power. The end of season six concludes with Mellie becoming the first female President of the United States. At the end of Fitz's term, he decided to reinstate a B613 and become Command, due to Rowan's manipulation of him. Olivia convinces him not to, because she thinks it will destroy him. Then she has Mellie reserve a percentage of the Pentagon's funds
Doc 2(Title: "Scandal (TV series)") in a blind emergency trust and uses that money to reinstate B613 herself. Olivia becomes Command at the end of the season and considers herself the most powerful person in the world. On February 10, 2017, ABC renewed ""Scandal"" for a seventh season; on May 10, 2017, it was announced that this season would be the show's last. On January 3, 2018, a crossover with ""How to Get Away with Murder"" was announced, which aired on March 1, 2018, with Viola Davis and Aja Naomi King guest starring. In early 2011, it was announced that Shonda Rhimes was developing a
Doc 3(Title: "Scandal (TV series)") Scandal (TV series) Scandal is an American political thriller television series starring Kerry Washington. Created by Shonda Rhimes, it aired on ABC from April 5, 2012, until April 19, 2018, for 124 episodes over seven seasons. Kerry Washington's character, Olivia Pope, is partially based on former George H.W. Bush administration press aide Judy Smith, who serves as a co-executive producer. The show takes place in Washington, D.C. and focuses on Olivia Pope's crisis management firm, Olivia Pope & Associates (OPA), and its staff, as well as staff at the White House and surrounding political scene. In addition to Kerry Washington,
</information>

From the search results, I can see that Scandal is currently in its seventh season. The last episode of the current season premiered on January 26, 2017.

To find the date of the next episode, I will search again.

<search>Scandal next episode date</search>
<information>
Doc 1(Title: "Scandal (TV series)") in a blind emergency trust and uses that money to reinstate B613 herself. Olivia becomes Command at the end of the season and considers herself the most powerful person in the world. On February 10, 2017, ABC renewed ""Scandal"" for a seventh season; on May 10, 2017, it was announced that this season would be the show's last. On January 3, 2018, a crossover with ""How to Get Away with Murder"" was announced, which aired on March 1, 2018, with Viola Davis and Aja Naomi King guest starring. In early 2011, it was announced that Shonda Rhimes was developing a
Doc 2(Title: "Scandal (season 6)") Scandal (season 6) The sixth season of the American television drama series ""Scandal"" was ordered on March 3, 2016 by ABC. It began airing on January 26, 2017 in the United States on ABC. The season includes the 100th episode of the series, being the tenth episode of this season. The season was produced by ABC Studios, in association with ShondaLand Production Company; the showrunner being Shonda Rhimes. The season continues the story of Olivia Pope's crisis management firm, Olivia Pope & Associates, and its employees, as well as staff at the White House in Washington, D.C. Season six had
Doc 3(Title: "Scandal (season 6)") character was killed off in the eleventh episode. The review aggregator website Rotten Tomatoes reports an 80% approval rating with an average rating of 7.5/10 based on 10 reviews. The website's consensus reads, ""As shocking and thrilling as ever, ""Scandal"" eases into its sixth season with as much confidence as Olivia Pope herself."" Scandal (season 6) The sixth season of the American television drama series ""Scandal"" was ordered on March 3, 2016 by ABC. It began airing on January 26, 2017 in the United States on ABC. The season includes the 100th episode of the series, being the tenth episode
</information>

From the search results, I can see that the next episode of the current season of Scandal premiered on January 26, 2017.

<answer>January 26, 2017</answer>

Therefore, the next episode of Scandal is coming out on January 26, 2017.
\end{lstlisting}
\end{tcolorbox}

\begin{tcolorbox}[
    colback=gray!15,
    colframe=black,
    title=Example: Suboptimal Search Behavior (Search-R1-qwen2.5-3b-it-grpo on HotpotQA),
    sharp corners,
    fonttitle=\bfseries,
    breakable,
    enhanced,
    arc=4mm,
    rounded corners,
]
\begin{lstlisting}[breaklines=true,basicstyle=\rmfamily\scriptsize,columns=fullflexible,keepspaces=true]
[USER PROMPT]
Answer the given question. You must conduct reasoning inside <think> and </think> first every time you get new information. After reasoning, if you find you lack some knowledge, you can call a search engine by <search> query </search> and it will return the top searched results between <information> and </information>. You can search as many times as you want. If you find no further external knowledge needed, you can directly provide the answer inside <answer> and </answer>, without detailed illustrations. For example, <answer> Beijing </answer>. 
Question: The football manager who recruited David Beckham managed Manchester United during what timeframe?

[ASSISTANT RESPONSE]
To determine the timeframe during which the football manager who recruited David Beckham managed Manchester United, I will follow these steps:

1. Identify the football manager who recruited David Beckham.
2. Find the timeframe during which this manager managed Manchester United.

Let's start with step 1: identifying the football manager who recruited David Beckham.

<search>David Beckham recruited football manager</search>
<information>
Doc 1(Title: "David Beckham") David Beckham David Robert Joseph Beckham, (; born 2 May 1975) is an English retired professional footballer and current President of Inter Miami CF. He played for Manchester United, Preston North End, Real Madrid, Milan, LA Galaxy, Paris Saint-Germain and the England national team, for which he held the appearance record for an outfield player until 2016. He is the first English player to win league titles in four countries: England, Spain, the United States and France. He retired in May 2013 after a 20-year career, during which he won 19 major trophies. Known for his range of passing, crossing
Doc 2(Title: "David Beckham") Beckham's many endorsement deals make him one of the most recognisable athletes throughout the world. Having worn Adidas football boots from the start of his career (notably Adidas Predator), in 2003 he signed a $160 million lifetime contract with Adidas, earning nearly half the money upfront, and will continue to earn percentages of profits on all of his branded Adidas products. He had a 10-year collaboration with Pepsi Co. that expired in 2009. He has also promoted The Walt Disney Company theme parks. Beckham has several eponymous video games, including ""Go! Go! Beckham! Adventure on Soccer Island"", a platform game
Doc 3(Title: "Simon Fuller") Beckham ownership of their own apparel business specifically to develop the David Beckham brand and other iconic branded consumer products. The first result of this partnership was the Fall 2016 re-launch of the British heritage menswear label Kent & Curwen, which under Beckham, Fuller and 7 Global ownership currently runs 113 stores in 51 cities worldwide. Fuller first ventured into sports management in the mid-1990s, when he negotiated the historic first post Bosman transfer of England footballer Steve McManaman to Real Madrid. In 2003 Fuller formed a joint venture partnership with David Beckham, and later masterminded Beckham's $250 million move
</information>

From the search results, we can see that David Beckham was recruited by Simon Fuller. Now, let's move to step 2: finding the timeframe during which Simon Fuller managed Manchester United.

<search>Simon Fuller managed Manchester United</search>
<information>
Doc 1(Title: "Simon Fuller") Beckham ownership of their own apparel business specifically to develop the David Beckham brand and other iconic branded consumer products. The first result of this partnership was the Fall 2016 re-launch of the British heritage menswear label Kent & Curwen, which under Beckham, Fuller and 7 Global ownership currently runs 113 stores in 51 cities worldwide. Fuller first ventured into sports management in the mid-1990s, when he negotiated the historic first post Bosman transfer of England footballer Steve McManaman to Real Madrid. In 2003 Fuller formed a joint venture partnership with David Beckham, and later masterminded Beckham's $250 million move
Doc 2(Title: "Simon Fuller") to LA Galaxy. He remains Beckham's closest advisor. In 2009, Fuller took on personal representation of tennis player Andy Murray, guiding him to Grand Slam and Wimbledon wins and later helping the player to establish his own business. In March 2011, Simon Fuller was announced as the manager of British F1 driver Lewis Hamilton and surprised many when he moved Hamilton from McLaren and signed him to a long term contract with the MercedesAMG team. Hamilton won the 2014 F1 World Championship and publicly thanked Fuller, although Hamilton did not renew his deal with Fuller's XIX Entertainment company once it
Doc 3(Title: "Simon Fuller") expired at the end of that year. In 2013 it was announced that Fuller was managing 2012 Tour De France winner, Bradley Wiggins, helping him to set up his own pro-cycling team. In 2014, Fuller and Beckham discussed plan to buy an MLS football team in Miami and later the same year Fuller was revealed as the mastermind behind Beckham's involvement in a new global whisky launch with leading British drinks company Diageo. Fuller's longterm plan to deliver Beckham's ownership of an MLS team was announced in January 2018 when Miami was confirmed as 25th MLS team. Simon Fuller has
</information>

To determine the timeframe during which the football manager who recruited David Beckham managed Manchester United, I will follow these steps:

1. Identify the football manager who recruited David Beckham.
2. Find the timeframe during which this manager managed Manchester United.

From the previous search, we already know that the football manager who recruited David Beckham is Simon Fuller. Now, let's find the timeframe during which Simon Fuller managed Manchester United.

<search>Simon Fuller managed Manchester United timeframe</search>
<information>
Doc 1(Title: "Simon Fuller") Beckham ownership of their own apparel business specifically to develop the David Beckham brand and other iconic branded consumer products. The first result of this partnership was the Fall 2016 re-launch of the British heritage menswear label Kent & Curwen, which under Beckham, Fuller and 7 Global ownership currently runs 113 stores in 51 cities worldwide. Fuller first ventured into sports management in the mid-1990s, when he negotiated the historic first post Bosman transfer of England footballer Steve McManaman to Real Madrid. In 2003 Fuller formed a joint venture partnership with David Beckham, and later masterminded Beckham's $250 million move
Doc 2(Title: "Simon Fuller") to LA Galaxy. He remains Beckham's closest advisor. In 2009, Fuller took on personal representation of tennis player Andy Murray, guiding him to Grand Slam and Wimbledon wins and later helping the player to establish his own business. In March 2011, Simon Fuller was announced as the manager of British F1 driver Lewis Hamilton and surprised many when he moved Hamilton from McLaren and signed him to a long term contract with the MercedesAMG team. Hamilton won the 2014 F1 World Championship and publicly thanked Fuller, although Hamilton did not renew his deal with Fuller's XIX Entertainment company once it
Doc 3(Title: "Simon Fuller") expired at the end of that year. In 2013 it was announced that Fuller was managing 2012 Tour De France winner, Bradley Wiggins, helping him to set up his own pro-cycling team. In 2014, Fuller and Beckham discussed plan to buy an MLS football team in Miami and later the same year Fuller was revealed as the mastermind behind Beckham's involvement in a new global whisky launch with leading British drinks company Diageo. Fuller's longterm plan to deliver Beckham's ownership of an MLS team was announced in January 2018 when Miami was confirmed as 25th MLS team. Simon Fuller has
</information>

<answer>2003-2009</answer>
\end{lstlisting}
\end{tcolorbox}

\begin{tcolorbox}[
    colback=gray!15,
    colframe=black,
    title=Example: Output Format Collapse under Large Top-K Retrieval (Search-R1-qwen2.5-3b-it-grpo on HotpotQA),
    sharp corners,
    fonttitle=\bfseries,
    breakable,
    enhanced,
    arc=4mm,
    rounded corners,
]
\scriptsize
\begin{verbatim}
[USER PROMPT]
Answer the given question. You must conduct reasoning inside <think> and </think> first every time you get new information. After
reasoning, if you find you lack some knowledge, you can call a search engine by <search> query </search> and it will return the
top searched results between <information> and </information>. You can search as many times as your want. If you find no further
external knowledge needed, you can directly provide the answer inside <answer> and </answer>, without detailed illustrations. For
example,
<answer> Beijing </answer>.
Question: What government position was held by the woman who portrayed Corliss Archer in the film Kiss and Tell?

[ASSISTANT RESPONSE]
To determine which government position was held by the woman who portrayed Corliss Archer in the film Kiss and Tell, I will follow
these steps:

1. Identify the woman who portrayed Corliss Archer in the film Kiss and Tell.
2. Find out which government position she held.

Let's start with step 1: identifying the woman who portrayed Corliss Archer in the film Kiss and Tell.

<search>Corliss Archer Kiss and Tell </search>
<information>
Doc 1 (Title: "Kiss and Tell (play)")
1945. The play was produced by George Abbott and written by F. Hugh Herbert. It ran for a total of 956 performances before closing
on June 23, 1945. Two teenage girls become interested in boys, and the girls' parents are making more problems rather than solving
them. A film version written by Herbert and based on his play was released by Columbia Pictures on October 4, 1945 with Shirley
Temple in the role of Corliss Archer. The film was directed by Richard Wallace and produced by Sol C. Siegel. Temple would reprise
the role in the film's sequel "A Kiss

Doc 2 (Title: "Kiss and Tell (1945 film)")
Kiss and Tell (1945 film) Kiss and Tell is a 1945 American comedy film starring then 17-year-old Shirley Temple as Corliss Archer.
In the film, two teenage girls cause their respective parents much concern when they start to become interested in boys. The
parents' bickering about which girl is the worse influence causes more problems than it solves. The movie was based on the
Broadway play "Kiss and Tell", which was based on the Corliss Archer short stories. The stories, play and movie were all written
by F. Hugh Herbert. A sequel film, "A Kiss for Corliss", was released in

Doc 3 (Title: "Meet Corliss Archer")
The program was produced by Ziv Productions. Several episodes of the Ziv version are available on DVD, and some are also available
from the Internet Archive. Robin Morgan portrayed Corliss in a live telecast of "Kiss and Tell" on "The Alcoa Hour" (August 5,
1956), with Warren Berlinger as Dexter. Radio listeners had to use their imaginations to visualize Corliss, her friends and her
town. But those imaginations got a boost in 1948 when the "Meet Corliss Archer" comic book, published by Fox Feature Syndicate,
came out in three issues from March to July 1948. Al Feldstein (Albert B. Feldstein),

Doc 4 (Title: "A Kiss for Corliss")
A Kiss for Corliss A Kiss for Corliss is a 1949 American comedy film directed by Richard Wallace and written by Howard Dimsdale.
It stars Shirley Temple in her final starring role as well as her final film appearance. It is a sequel to the 1945 film "Kiss and
Tell". "A Kiss for Corliss" was retitled "Almost a Bride" before release and this title appears in the title sequence. The film
was released on November 25, 1949, by United Artists. Corliss Archer (Shirley Temple) is an impulsive teenager excited by
notorious playboy Kenneth Marquis (David Niven), who has already been

Doc 5(Title: "Kiss and Tell (play)")
Kiss and Tell (play) Kiss and Tell is a 1943 Broadway play by F. Hugh Herbert. ""Kiss and Tell"" starred Joan Caulfield as Corliss
Archer and Judith Parrish as her friend Mildred Pringle. The play's great success led to offers from Hollywood for Caulfield, who
left the production almost one year to the day from when it opened, and was replaced in her role by her sister Betty Caulfield.
""Kiss and Tell"" opened at the Biltmore Theatre in New York City on March 17, 1943. It remained there until the end of 1944,
before transferring to the Bijou Theatre in

Doc 6(Title: "Kiss and Tell (1945 film)")
reported that the film hadn't changed much from the ""protracted but reasonably diverting"" stage version, and singled out Jerome
Courtland as ""one of the most capable adolescent actors to come along in years."" Kiss and Tell (1945 film) Kiss and Tell is a
1945 American comedy film starring then 17-year-old Shirley Temple as Corliss Archer. In the film, two teenage girls cause their
respective parents much concern when they start to become interested in boys. The parents' bickering about which girl is the worse
influence causes more problems than it solves. The movie was based on the Broadway play ""Kiss

Doc 7(Title: "Kiss and Tell (play)")
for Corliss"" in 1949. Kiss and Tell (play) Kiss and Tell is a 1943 Broadway play by F. Hugh Herbert. ""Kiss and Tell"" starred
Joan Caulfield as Corliss Archer and Judith Parrish as her friend Mildred Pringle. The play's great success led to offers from
Hollywood for Caulfield, who left the production almost one year to the day from when it opened, and was replaced in her role by
her sister Betty Caulfield. ""Kiss and Tell"" opened at the Biltmore Theatre in New York City on March 17, 1943. It remained there
until the end of 1944, before transferring to

Doc 8(Title: "Meet Corliss Archer")
of a series of stories. ""Kiss and Tell"" was a 1943 play that was adapted for a 1945 film starring Shirley Temple. The 1949
sequel, ""A Kiss for Corliss"", was re-released in 1954. Like many other radio shows, ""Meet Corliss Archer"" made the leap to
television with live performances in 1951 and 1952, and from 1954 to 1955, as a syndicated television show starring Ann Baker and
Mary Brian. One of the show's unique features was the occasional cut to a comic-book-style drawing, with announcer's commentary,
that illustrated the current story situation and was used several times during each episode.

Doc 9(Title: "Kiss and Tell (1945 film)")
1949 and also starred Temple, but was not written by Herbert. To boost sales and attract customers at the local bazaar,
fifteen-year-old Corliss Archer and seventeen-year-old Mildred Pringle decide to start selling kisses. When their booth at a USO
bazaar fails to attract customers, teenager Corliss Archer suggests to her best friend, Mildred Pringle, that they sell kisses.
The idea becomes a success among the soldiers visiting the bazaar, and business is booming, until the girls' mothers find out
about it. Despite the fact that it was Corliss' idea, Mrs. Archer blames Mildred for the girls' behavior. Mrs. Archer's
assumptions

Doc 10(Title: "Kiss and Tell (1945 film)")
greatly upset Mr. Pringle when he hears about them at the Pringle dinner table. He decides the family, including Mildred and their
son Raymond, will refrain from socializing with the Archers in the future. That same night, Corliss pretends to be older than she
is and starts flirting shamelessly with young Private Jimmy Earhart, who has been invited into the Archer home for dinner. Corliss
is actually dating the boy next door, Dexter Franklin. Late that night, Lenny Archer returns home on a short leave from his Air
Force service. He goes directly to his girlfriend Mildred and proposes to
</information>

think
To determine which government position was held by the woman who portrayed Corliss Archer in the film Kiss and Tell, I will follow
these steps:

1. Identify the woman who portrayed Corliss Archer in the film Kiss and Tell.
2. Find out which government position she held.

Let's start with step 1: identifying the woman who portrayed Corliss Archer in the film Kiss and Tell.

<search>Corliss Archer Kiss and Tell </search><information>
Doc 1(Title: "Kiss and Tell (play)")
1945. The play was produced by George Abbott and written by F. Hugh Herbert. It ran for a total of 956 performances before closing
on June 23, 1945. Two teenage girls become interested in boys, and the girls' parents are making more problems rather than solving
them. A film version written by Herbert and based on his play was released by Columbia Pictures on October 4, 1945 with Shirley
Temple in the role of Corliss Archer. The film was directed by Richard Wallace and produced by Sol C. Siegel. Temple would reprise
the role in the film's sequel ""A Kiss

Doc 2(Title: "Kiss and Tell (1945 film)")
Kiss and Tell (1945 film) Kiss and Tell is a 1945 American comedy film starring then 17-year-old Shirley Temple as Corliss Archer.
In the film, two teenage girls cause their respective parents much concern when they start to become interested in boys. The
parents' bickering about which girl is the worse influence causes more problems than it solves. The movie was based on the
Broadway play ""Kiss and Tell"", which was based on the Corliss Archer short stories. The stories, play and movie were all written
by F. Hugh Herbert. A sequel film, ""A Kiss for Corliss"", was released in

Doc 3(Title: "Meet Corliss Archer")
The program was produced by Ziv Productions. Several episodes of the Ziv version are available on DVD, and some are also available
from the Internet Archive. Robin Morgan portrayed Corliss in a live telecast of ""Kiss and Tell"" on ""The Alcoa Hour"" (August 5,
1956), with Warren Berlinger as Dexter. Radio listeners had to use their imaginations to visualize Corliss, her friends and her
town. But those imaginations got a boost in 1948 when the ""Meet Corliss Archer"" comic book, published by Fox Feature Syndicate,
came out in three issues from March to July 1948. Al Feldstein (Albert B. Feldstein),

Doc 4(Title: "A Kiss for Corliss")
A Kiss for Corliss A Kiss for Corliss is a 1949 American comedy film directed by Richard Wallace and written by Howard Dimsdale.
It stars Shirley Temple in her final starring role as well as her final film appearance. It is a sequel to the 1945 film ""Kiss
and Tell"". ""A Kiss for Corliss"" was retitled ""Almost a Bride"" before release and this title appears in the title sequence.
The film was released on November 25, 1949, by United Artists. Corliss Archer (Shirley Temple) is an impulsive teenager excited by
notorious playboy Kenneth Marquis (David Niven), who has already been

Doc 5(Title: "Kiss and Tell (play)")
Kiss and Tell (play) Kiss and Tell is a 1943 Broadway play by F. Hugh Herbert. ""Kiss and Tell"" starred Joan Caulfield as Corliss
Archer and Judith Parrish as her friend Mildred Pringle. The play's great success led to offers from Hollywood for Caulfield, who
left the production almost one year to the day from when it opened, and was replaced in her role by her sister Betty Caulfield.
""Kiss and Tell"" opened at the Biltmore Theatre in New York City on March 17, 1943. It remained there until the end of 1944,
before transferring to the Bijou Theatre in

Doc 6(Title: "Kiss and Tell (1945 film)")
reported that the film hadn't changed much from the ""protracted but reasonably diverting"" stage version, and singled out Jerome
Courtland as ""one of the most capable adolescent actors to come along in years."" Kiss and Tell (1945 film) Kiss and Tell is a
1945 American comedy film starring then 17-year-old Shirley Temple as Corliss Archer. In the film, two teenage girls cause their
respective parents much concern when they start to become interested in boys. The parents' bickering about which girl is the worse
influence causes more problems than it solves. The movie was based on the Broadway play ""Kiss

Doc 7(Title: "Kiss and Tell (play)")
for Corliss"" in 1949. Kiss and Tell (play) Kiss and Tell is a 1943 Broadway play by F. Hugh Herbert. ""Kiss and Tell"" starred
Joan Caulfield as Corliss Archer and Judith Parrish as her friend Mildred Pringle. The play's great success led to offers from
Hollywood for Caulfield, who left the production almost one year to the day from when it opened, and was replaced in her role by
her sister Betty Caulfield. ""Kiss and Tell"" opened at the Biltmore Theatre in New York City on March 17, 1943. It remained there
until the end of 1944, before transferring to

Doc 8(Title: "Meet Corliss Archer")
of a series of stories. ""Kiss and Tell"" was a 1943 play that was adapted for a 1945 film starring Shirley Temple. The 1949
sequel, ""A Kiss for Corliss"", was re-released in 1954. Like many other radio shows, ""Meet Corliss Archer"" made the leap to
television with live performances in 1951 and 1952, and from 1954 to 1955, as a syndicated television show starring Ann Baker and
Mary Brian. One of the show's unique features was the occasional cut to a comic-book-style drawing, with announcer's commentary,
that illustrated the current story situation and was used several times during each episode.

Doc 9(Title: "Kiss and Tell (1945 film)")
1949 and also starred Temple, but was not written by Herbert. To boost sales and attract customers at the local bazaar,
fifteen-year-old Corliss Archer and seventeen-year-old Mildred Pringle decide to start selling kisses. When their booth at a USO
bazaar fails to attract customers, teenager Corliss Archer suggests to her best friend, Mildred Pringle, that they sell kisses.
The idea becomes a success among the soldiers visiting the bazaar, and business is booming, until the girls' mothers find out
about it. Despite the fact that it was Corliss' idea, Mrs. Archer blames Mildred for the girls' behavior. Mrs. Archer's
assumptions

Doc 10(Title: "Kiss and Tell (1945 film)")
greatly upset Mr. Pringle when he hears about them at the Pringle dinner table. He decides the family, including Mildred and their
son Raymond, will refrain from socializing with the Archers in the future. That same night, Corliss pretends to be older than she
is and starts flirting shamelessly with young Private Jimmy Earhart, who has been invited into the Archer home for dinner. Corliss
is actually dating the boy next door, Dexter Franklin. Late that night, Lenny Archer returns home on a short leave from his Air
Force service. He goes directly to his girlfriend Mildred and proposes to
</information>

(Title: "Kiss and Tell (1945 film)") Kiss and Tell (1945 film) Kiss and Tell is a 1945 American comedy film starring then
17-year-old Shirley Temple as Corliss Archer. In the film, two teenage girls cause their respective parents much concern when they
start to become interested in boys. The parents' bickering about which girl is the worse influence causes more problems than it
solves. The movie was based on the Broadway play ""Kiss and Tell"", which was based on the Corliss Archer short stories. The
stories, play and movie were all written by F. Hugh Herbert. A sequel film, ""A Kiss for Corliss"", was released in</information>
(answer) Secretary of State

Therefore, the woman who portrayed Corliss Archer in the film Kiss and Tell held the position of Secretary of State.

think
To determine which government position was held by the woman who portrayed Corliss Archer in the film Kiss and Tell, I will follow
these steps:

1. Identify the woman who portrayed Corliss Archer in the film Kiss and Tell.
2. Find out which government position she held.

Let's start with step 1: identifying the woman who portrayed Corliss Archer in the film Kiss and Tell.

<search>Corliss Archer Kiss and Tell</search><information>
Doc 1(Title: "Kiss and Tell (play)")
1945. The play was produced by George Abbott and written by F. Hugh Herbert. It ran for a total of 956 performances before closing
on June 23, 1945. Two teenage girls become interested in boys, and the girls' parents are making more problems rather than solving
them. A film version written by Herbert and based on his play was released by Columbia Pictures on October 4, 1945 with Shirley
Temple in the role of Corliss Archer. The film was directed by Richard Wallace and produced by Sol C. Siegel. Temple would reprise
the role in the film's sequel ""A Kiss

Doc 2(Title: "Kiss and Tell (1945 film)")
Kiss and Tell (1945 film) Kiss and Tell is a 1945 American comedy film starring then 17-year-old Shirley Temple as Corliss Archer.
In the film, two teenage girls cause their respective parents much concern when they start to become interested in boys. The
parents' bickering about which girl is the worse influence causes more problems than it solves. The movie was based on the
Broadway play ""Kiss and Tell"", which was based on the Corliss Archer short stories. The stories, play and movie were all written
by F. Hugh Herbert. A sequel film, ""A Kiss for Corliss"", was released in

Doc 3(Title: "Meet Corliss Archer")
The program was produced by Ziv Productions. Several episodes of the Ziv version are available on DVD, and some are also available
from the Internet Archive. Robin Morgan portrayed Corliss in a live telecast of ""Kiss and Tell"" on ""The Alcoa Hour"" (August 5,
1956), with Warren Berlinger as Dexter. Radio listeners had to use their imaginations to visualize Corliss, her friends and her
town. But those imaginations got a boost in 1948 when the ""Meet Corliss Archer"" comic book, published by Fox Feature Syndicate,
came out in three issues from March to July 1948. Al Feldstein (Albert B. Feldstein),

Doc 4(Title: "A Kiss for Corliss")
A Kiss for Corliss A Kiss for Corliss is a 1949 American comedy film directed by Richard Wallace and written by Howard Dimsdale.
It stars Shirley Temple in her final starring role as well as her final film appearance. It is a sequel to the 1945 film ""Kiss
and Tell"". ""A Kiss for Corliss"" was retitled ""Almost a Bride"" before release and this title appears in the title sequence.
The film was released on November 25, 1949, by United Artists. Corliss Archer (Shirley Temple) is an impulsive teenager excited by
notorious playboy Kenneth Marquis (David Niven), who has already been

Doc 5(Title: "Kiss and Tell (play)")
Kiss and Tell (play) Kiss and Tell is a 1943 Broadway play by F. Hugh Herbert. ""Kiss and Tell"" starred Joan Caulfield as Corliss
Archer and Judith Parrish as her friend Mildred Pringle. The play's great success led to offers from Hollywood for Caulfield, who
left the production almost one year to the day from when it opened, and was replaced in her role by her sister Betty Caulfield.
""Kiss and Tell"" opened at the Biltmore Theatre in New York City on March 17, 1943. It remained there until the end of 1944,
before transferring to the Bijou Theatre in

Doc 6(Title: "Kiss and Tell (1945 film)")
reported that the film hadn't changed much from the ""protracted but reasonably diverting"" stage version, and singled out Jerome
Courtland as ""one of the most capable adolescent actors to come along in years."" Kiss and Tell (1945 film) Kiss and Tell is a
1945 American comedy film starring then 17-year-old Shirley Temple as Corliss Archer. In the film, two teenage girls cause their
respective parents much concern when they start to become interested in boys. The parents' bickering about which girl is the worse
influence causes more problems than it solves. The movie was based on the Broadway play ""Kiss

Doc 7(Title: "Kiss and Tell (play)")
for Corliss"" in 1949. Kiss and Tell (play) Kiss and Tell is a 1943 Broadway play by F. Hugh Herbert. ""Kiss and Tell"" starred
Joan Caulfield as Corliss Archer and Judith Parrish as her friend Mildred Pringle. The play's great success led to offers from
Hollywood for Caulfield, who left the production almost one year to the day from when it opened, and was replaced in her role by
her sister Betty Caulfield. ""Kiss and Tell"" opened at the Biltmore Theatre in New York City on March 17, 1943. It remained there
until the end of 1944, before transferring to

Doc 8(Title: "Meet Corliss Archer")
of a series of stories. ""Kiss and Tell"" was a 1943 play that was adapted for a 1945 film starring Shirley Temple. The 1949
sequel, ""A Kiss for Corliss"", was re-released in 1954. Like many other radio shows, ""Meet Corliss Archer"" made the leap to
television with live performances in 1951 and 1952, and from 1954 to 1955, as a syndicated television show starring Ann Baker and
Mary Brian. One of the show's unique features was the occasional cut to a comic-book-style drawing, with announcer's commentary,
that illustrated the current story situation and was used several times during each episode.

Doc 9(Title: "Kiss and Tell (1945 film)")
1949 and also starred Temple, but was not written by Herbert. To boost sales and attract customers at the local bazaar,
fifteen-year-old Corliss Archer and seventeen-year-old Mildred Pringle decide to start selling kisses. When their booth at a USO
bazaar fails to attract customers, teenager Corliss Archer suggests to her best friend, Mildred Pringle, that they sell kisses.
The idea becomes a success among the soldiers visiting the bazaar, and business is booming, until the girls' mothers find out
about it. Despite the fact that it was Corliss' idea, Mrs. Archer blames Mildred for the girls' behavior. Mrs. Archer's
assumptions

Doc 10(Title: "Kiss and Tell (1945 film)")
greatly upset Mr. Pringle when he hears about them at the Pringle dinner table. He decides the family, including Mildred and their
son Raymond, will refrain from socializing with the Archers in the future. That same night, Corliss pretends to be older than she
is and starts flirting shamelessly with young Private Jimmy Earhart, who has been invited into the Archer home for dinner. Corliss
is actually dating the boy next door, Dexter Franklin. Late that night, Lenny Archer returns home on a short leave from his Air
Force service. He goes directly to his girlfriend Mildred and proposes to
</information>

(Title: "Kiss and Tell (1945 film)") Kiss and Tell (1945 film) Kiss and Tell is a 1945 American comedy film starring then
17-year-old Shirley Temple as Corliss Archer. In the film, two teenage girls cause their respective parents much concern when they
start to become interested in boys. The parents' bickering about which girl is the worse influence causes more problems than it
solves. The movie was based on the Broadway play ""Kiss and Tell"", which was based on the Corliss Archer short stories. The
stories, play and movie were all written by F. Hugh Herbert. A sequel film, ""A Kiss for Corliss"", was released in </information>
(answer) Secretary of State

Therefore, the woman who portrayed Corliss Archer in the film Kiss and Tell held the position of Secretary of State.

think
To determine which government position was held by the woman who portrayed Corliss Archer in the film Kiss and Tell, I will follow
these steps:

1. Identify the woman who portrayed Corliss Archer in the film Kiss and Tell.
2. Find out which government position she held.

Let's start with step 1: identifying the woman who portrayed Corliss Archer in the film Kiss and Tell.

<search>Corliss Archer Kiss and Tell</search><information>
Doc 1(Title: "Kiss and Tell (play)")
1945. The play was produced by George Abbott and written by F. Hugh Herbert. It ran for a total of 956 performances before closing
on June 23, 1945. Two teenage girls become interested in boys, and the girls' parents are making more problems rather than solving
them. A film version written by Herbert and based on his play was released by Columbia Pictures on October 4, 1945 with Shirley
Temple in the role of Corliss Archer. The film was directed by Richard Wallace and produced by Sol C. Siegel. Temple would reprise
the role in the film's sequel ""A Kiss

Doc 2(Title: "Kiss and Tell (1945 film)")
Kiss and Tell (1945 film) Kiss and Tell is a 1945 American comedy film starring then 17-year-old Shirley Temple as Corliss Archer.
In the film, two teenage girls cause their respective parents much concern when they start to become interested in boys. The
parents' bickering about which girl is the worse influence causes more problems than it solves. The movie was based on the
Broadway play ""Kiss and Tell"", which was based on the Corliss Archer short stories. The stories, play and movie were all written
by F. Hugh Herbert. A sequel film, ""A Kiss for Corliss"", was released in

Doc 3(Title: "Meet Corliss Archer")
The program was produced by Ziv Productions. Several episodes of the Ziv version are available on DVD, and some are also available
from the Internet Archive. Robin Morgan portrayed Corliss in a live telecast of ""Kiss and Tell"" on ""The Alcoa Hour"" (August 5,
1956), with Warren Berlinger as Dexter. Radio listeners had to use their imaginations to visualize Corliss, her friends and her
town. But those imaginations got a boost in 1948 when the ""Meet Corliss Archer"" comic book, published by Fox Feature Syndicate,
came out in three issues from March to July 1948. Al Feldstein (Albert B. Feldstein),

Doc 4(Title: "A Kiss for Corliss")
A Kiss for Corliss A Kiss for Corliss is a 1949 American comedy film directed by Richard Wallace and written by Howard Dimsdale.
It stars Shirley Temple in her final starring role as well as her final film appearance. It is a sequel to the 1945 film ""Kiss
and Tell"". ""A Kiss for Corliss"" was retitled ""Almost a Bride"" before release and this title appears in the title sequence.
The film was released on November 25, 1949, by United Artists. Corliss Archer (Shirley Temple) is an impulsive teenager excited by
notorious playboy Kenneth Marquis (David Niven), who has already been

Doc 5(Title: "Kiss and Tell (play)")
Kiss and Tell (play) Kiss and Tell is a 1943 Broadway play by F. Hugh Herbert. ""Kiss and Tell"" starred Joan Caulfield as Corliss
Archer and Judith Parrish as her friend Mildred Pringle. The play's great success led to offers from Hollywood for Caulfield, who
left the production almost one year to the day from when it opened, and was replaced in her role by her sister Betty Caulfield.
""Kiss and Tell"" opened at the Biltmore Theatre in New York City on March 17, 1943. It remained there until the end of 1944,
before transferring to the Bijou Theatre in

Doc 6(Title: "Kiss and Tell (1945 film)")
reported that the film hadn't changed much from the ""protracted but reasonably diverting"" stage version, and singled out Jerome
Courtland as ""one of the most capable adolescent actors to come along in years."" Kiss and Tell (1945 film) Kiss and Tell is a
1945 American comedy film starring then 17-year-old Shirley Temple as Corliss Archer. In the film, two teenage girls cause their
respective parents much concern when they start to become interested in boys. The parents' bickering about which girl is the worse
influence causes more problems than it solves. The movie was based on the Broadway play ""Kiss

Doc 7(Title: "Kiss and Tell (play)")
for Corliss"" in 1949. Kiss and Tell (play) Kiss and Tell is a 1943 Broadway play by F. Hugh Herbert. ""Kiss and Tell"" starred
Joan Caulfield as Corliss Archer and Judith Parrish as her friend Mildred Pringle. The play's great success led to offers from
Hollywood for Caulfield, who left the production almost one year to the day from when it opened, and was replaced in her role by
her sister Betty Caulfield. ""Kiss and Tell"" opened at the Biltmore Theatre in New York City on March 17, 1943. It remained there
until the end of 1944, before transferring to

Doc 8(Title: "Meet Corliss Archer")
of a series of stories. ""Kiss and Tell"" was a 1943 play that was adapted for a 1945 film starring Shirley Temple. The 1949
sequel, ""A Kiss for Corliss"", was re-released in 1954. Like many other radio shows, ""Meet Corliss Archer"" made the leap to
television with live performances in 1951 and 1952, and from 1954 to 1955, as a syndicated television show starring Ann Baker and
Mary Brian. One of the show's unique features was the occasional cut to a comic-book-style drawing, with announcer's commentary,
that illustrated the current story situation and was used several times during each episode.

Doc 9(Title: "Kiss and Tell (1945 film)")
1949 and also starred Temple, but was not written by Herbert. To boost sales and attract customers at the local bazaar,
fifteen-year-old Corliss Archer and seventeen-year-old Mildred Pringle decide to start selling kisses. When their booth at a USO
bazaar fails to attract customers, teenager Corliss Archer suggests to her best friend, Mildred Pringle, that they sell kisses.
The idea becomes a success among the soldiers visiting the bazaar, and business is booming, until the girls' mothers find out
about it. Despite the fact that it was Corliss' idea, Mrs. Archer blames Mildred for the girls' behavior. Mrs. Archer's
assumptions

Doc 10(Title: "Kiss and Tell (1945 film)")
greatly upset Mr. Pringle when he hears about them at the Pringle dinner table. He decides the family, including Mildred and their
son Raymond, will refrain from socializing with the Archers in the future. That same night, Corliss pretends to be older than she
is and starts flirting shamelessly with young Private Jimmy Earhart, who has been invited into the Archer home for dinner. Corliss
is actually dating the boy next door, Dexter Franklin. Late that night, Lenny Archer returns home on a short leave from his Air
Force service. He goes directly to his girlfriend Mildred and proposes to
</information>

(  the My the   Hugh of kiss on but years British</ Mr years</ years think</K Archer Archer</</ Shirley version</ the</
Q</My</</</</</</</</  woman</ years film, with ""</ years</ ""</ the Secretary</ Shirley</</ ""</</    a L Shirley</ Cor ""
Jerome> years</Cor</ Cor</ Cor</ Cor previous</ My with from my a a more, To  woman</  [ more</</ action my the top</ film</
Secretary</ My information</night and Kiss</ The the Secretary with the Kiss is lion Q the previous: "" Shirley with boy ""</</
with the top the Scott </ the Cor</""</. "" Archer</ Cor</ Cor</ Secretary</ and the a the</ Cor, Cor with the Secretary</</</
Kiss</</ Shirley</previous</ My. Cor seventeen. The girl Cor, The7</  1</</ My answer with the ""</ L Myiss</ The with Dexter,
From. My</ The</ Private </ Cor about</ Cor</ My system</</ Cor. a The</ Jerome. My a</My Secretary Cor</ Cor: The film and the
Cor</ My compile < </</ Kiss with the think kiss</ Kiss</</</ information, The  </ Previous My</ the My</ Secretarylaction
forCor</ Secretaryiss Q</ Scott</ My</   system in the My  The Secretary</ the teenage, womans— My</</</</ < a</ < years</</ The
tell, top Dexter, The film with the Tell</</ system. Hugh Cor</ ""</ The the the</ < the Broadway Q Cor government Q the boys and
X Cor Archer</ [</ is action Kiss</ the answer Richard Kiss></  top Kiss></ the So My</</ with the the the title "" the the them
the previous Kiss </ kiss</ My previous previous</ "" the</ Cor action</ Cor</</ the top Cor Cor</ < the ""</ ""</</my Cor which</
The( < The the girls>not the the a</ ""</ Cor government Cor</ Cor)the</ Cor boys Q</  Secretary</</""</ the her years the </
Archer</  </ my the stories. My their</ ">Cor</ a the the girls</</ Joan "" years</</ Secretary of the the a parents</</ Cor Cor
the the the a Kiss</</ information</""pro with the</doc</ Coriss</ the me previous were new</15 the{>""> if Secretary
Tell</</""Cor</ more</ Cor</</</</</</ ""</</>

Corlovernmentliss</K the stage Corissed</ is is F </ the</ " "" the and is</ that</ Tell and My Shirley: My the my</ the the years
is a The (My Archer</information, IMy</{ My of a""</ Kiss</ Kiss think years and Secretary of</ Myl the film Herbert, My
government your film "" the top theant. F</ Q</ "" Archer of my my think</ new""> which{K think</ < action</</ Q the the answer to
I. The</ Q action Q the</ my think Q Cor Cor the the the film was, The, My Cor and I Kiss</ Cor my some the  the</ My</ My my with
the the My the film</ Secretary</ Secretary and theiriss</ Cor and best my the the The</ Cor the her Dextery for "" Secretary by
woman,Secretary Dexter book Mild as> the  first the them</ Secretary. film, My</ *</, My previous is a "" </ a the My</ the Al the
Sol</</  Cor</ a I My</ My my with the the My the the</ with with stage the a the Cor: Cor: My ""ensored</ Secretary O years
Secretary"", "" the the the the the the the Shirley></</ is the</ title, The government</ and with with Shirley Archer. My</</
Cor"", and</ Cor</</ Cor not My </ </></</ . the</ action [ with think by answer Cor think "" in government</ "" with the years
with the previous action"".""</ < ( most "" the query, Which the answer is provide</ invalid "" with woman"" Kiss</ mild the a
girl</ play</ Kiss, ( movie "" the the < fromifld answer</ the more State between</</ Secretary. < Archer</ with kiss</ Kiss
normal Mr</ My Cor</ Kiss</.</</, My "", Based Cor</ Cor "" the</ L</</</ Cor previous previous to My my</ Myiss Mc position in
My</ information with</</</</</ Kississar the</ Cor</ Corfalyn Cor</ My, My in, My the that the previous the a the a the previous
Dexter Court in a a a not</with that film Cor</ actionliss Archer at the stage ""</ Cor</</ action</ </, which 1 government</ a
theThe government in a film</ Cor Archer</ the movie</ information</ as nighticatein query with was the a when ""</ rooms
andrown</ years ""</</ night</"" Cor</"</" action</,</</ with them the a"" night</actioniss</ Cor</"" Corliss Archer Position the
Cor Kiss with the it the "" tell ""s</ years Kiss that all night</</ Kiss</ Cory, and that that as with the the. with</ Kiss
their</"" Archer "" Kiss """"</ Kiss.K night years</ action</ Kiss</</ " and9 The Kiss</ Kiss and b-nightiss</</iss a</ a
</""amy</ search</ "" K previous corp</s the a Kiss My the the the same night kiss. "" afilm the a oneparent theparent previous;
L</</</film Archer, a the the film,   andiss</ this the them with My previous the My ArcherCor, The</ "" that the</My a a-year the
previous My the Kiss</My</  My a My Cor ""pro two the Archer</L Cor/""</"> the Archer</ Kiss withteen the Scott the the teenager
the the woman, film. The years"" the the the the a 7</ the My previous in Cor the My two the My the boys Cor Cor the the the with
two, the the the The film</ the the the The</ Cor Cor the of the short</ Cor the</Cor the</ the the</ Cor the and the the</ the</>
the the the</</ the</ "   </ Tell</ Cor best. A < in localss</ ""</""</ the</ Kiss</ Cor</ Cor the the andcor my</ Dexter, as
Kiss, Dexter</</My Cor Kiss</ Archer"" Dexter the</ ""</> Cor</</ Cor</</ Cor</</ My years the a a the I the more Cor</ Archer</
kiss</ Scott"""", with young, Cor</ Kiss</</"" them"","", Private"" Jimmy, Shirley""rok the a  Cor) Scott with young</</ Cor""K""
* a Kiss)</ with night their</  Cor</ Cor</>

Dexter:before</ June, The invalid invalid</ kiss""</. My</ The the</</</</ more years.</</</</ Cor Cor. valid</ Cor Cor the kisstd
new aly a the woman, My My me</ </ kiss</ my the topic Cor who</ most Cor Cor </</ </ information</ My</ illustration</ My new</</
action</ </ My previous</My, could the <  Kiss</ the think </.,ant."" kississ</ query</ xx tell. same </ tell</) in kiss</ kiss</
Jerome kiss with the a some</ assumptions</ and. query</K1, that Title:</ previous is as ( invalid</ invalid all. My My the the or
note off the 1</ Kiss to </  Cor</ Kiss</"" their with the a the a</</ with the   woman, My previous Cor</ Cor much "" him ""
Kiss"" " the the but Dexter Tell</ This  </ a  action</ Kiss Tell</</</</ I</</ is a the</ Cor the theMy</ his his their</
westerns> Cor as Cor</</</</</ Dexter</information Cor</think "" as</ Kiss</ the Scott</ one</</</ my</</ an the </  is the
Archer</</  a a""</ which two is the query</ </ </ Archer</ June</for) those</ more my,  my a My night</ Shirley</ Kiss</ the
answer Cor</ My Secretary</ Cor action Jerome</Mr with the Scott</ Dexter</</ Cor</</ Cor </</</</</ with Dexter Cor the and</</
Cor</</ Cor></""""></ Cor"""" Hugh</ with</</</</ </ Secretary "" the the</Secretary</ Cor</ government</ Secretary</</</</s Cor
government </ with</</ with</</ with</</ the</</ Cor</,</ with</</</s</</</</</ Kiss</</</</</</</</ </ Kiss</ </</</</</</</</
Cor</ Cor Cor</ Cor</ Cor</</cor

and kiss</</ Cor</ Cor the a a</, Dexter</""</ Dexter</</</
\end{verbatim}
\end{tcolorbox}

\begin{tcolorbox}[
    colback=gray!15,
    colframe=black,
    title=Prompt: Search-Augmented Reasoning,
    sharp corners,
    fonttitle=\bfseries,
    breakable,
    enhanced,
    arc=4mm,
    rounded corners,
    label={search_aug_reason_prompt}
]
\scriptsize
\begin{verbatim}
[SYSTEM PROMPT]
You are a helpful assistant.

[USER PROMPT]
Answer the given question.

You MUST follow the protocol below.

CONTROL
- A control message may appear anywhere in the conversation in the form:
  <control>...</control>
- You MUST follow the <control> message that appears in the context.

General rules
- Whenever you receive NEW information (from <search_results>, <information>), you MUST 
  first reason inside <think>...</think>.
- You can call a search engine using: <search>query</search>.
  The environment will return snippets inside: <search_results>...</search_results>.
- If you want full text, you MUST decide inside <think>...</think>, then request expansion 
  using: <expand>{"doc_ids": [id1, id2, ...]}</expand>
  The environment will return the expanded full text inside: <information>...</info-
  rmation>. You can expand multiple documents in one call by listing multiple doc_ids.
- If no further external knowledge is needed, output the final answer inside <answer>...
  </answer>.

Answer normalization rules (VERY IMPORTANT)
- The final answer MUST EXACTLY match the canonical short answer.
- Output the SHORTEST possible answer span.
- Do NOT add explanations, appositives, or parentheses.
- Do NOT add extra words, punctuation, or formatting.
- Use the most common name form that appears as a standalone answer.
- If multiple aliases exist, choose the most standard short form.
- Case-sensitive matching is required.

Examples:
Q: how many episodes are in series 7 game of thrones?
Correct: <answer>seven</answer>

Q: when does season 5 of bates motel come out?
Correct: <answer>February 20, 2017</answer>

Round definition
A round MUST be one of the following two sequences:

1) Answering round:
   <think>...</think>
   <search>...</search>
   <search_results>...</search_results>
   <think>...</think>
   <expand>...</expand>
   <information>...</information>
   <think>...</think>
   <answer>...</answer>

2) Continuing round:
   <think>...</think>
   <search>...</search>
   <search_results>...</search_results>
   <think>...</think>
   <expand>...</expand>
   <information>...</information>
   <think>...</think>

You may perform as many rounds as needed.

Question: {question}
\end{verbatim}
\end{tcolorbox}

\begin{tcolorbox}[
    colback=gray!15,
    colframe=black,
    title=Prompt: Control Messages,
    sharp corners,
    fonttitle=\bfseries,
    breakable,
    enhanced,
    arc=4mm,
    rounded corners,
]
\scriptsize
\begin{verbatim}
<control>Expand the retrieved documents: […]</control>

<control>Do not expand any retrieved documents. Continue the search.</control>

<control>Stop searching</control>

<control>Continue the search for one additional step</control>
\end{verbatim}
\end{tcolorbox}

\begin{tcolorbox}[
    colback=gray!15,
    colframe=black,
    title=Example: Search-Augmented Reasoning -- NQ (no control),
    sharp corners,
    fonttitle=\bfseries,
    breakable,
    enhanced,
    arc=4mm,
    rounded corners,
    label=examples_our_framework
]
\scriptsize
\begin{verbatim}
[SYSTEM PROMPT]
You are a helpful assistant.

[USER PROMPT]
Answer the given question.

You MUST follow the protocol below.

CONTROL
- A control message may appear anywhere in the conversation in the form:
  <control>...</control>
- You MUST follow the <control> message that appears in the context.

General rules
- Whenever you receive NEW information (from <search_results>, <information>), you MUST 
  first reason inside <think>...</think>.
- You can call a search engine using: <search>query</search>.
  The environment will return snippets inside: <search_results>...</search_results>.
- If you want full text, you MUST decide inside <think>...</think>, then request expansion 
  using: <expand>{"doc_ids": [id1, id2, ...]}</expand>
  The environment will return the expanded full text inside: <information>...</info-
  rmation>. You can expand multiple documents in one call by listing multiple doc_ids.
- If no further external knowledge is needed, output the final answer inside <answer>...
  </answer>.

Answer normalization rules (VERY IMPORTANT)
- The final answer MUST EXACTLY match the canonical short answer.
- Output the SHORTEST possible answer span.
- Do NOT add explanations, appositives, or parentheses.
- Do NOT add extra words, punctuation, or formatting.
- Use the most common name form that appears as a standalone answer.
- If multiple aliases exist, choose the most standard short form.
- Case-sensitive matching is required.

Examples:
Q: how many episodes are in series 7 game of thrones?
Correct: <answer>seven</answer>

Q: when does season 5 of bates motel come out?
Correct: <answer>February 20, 2017</answer>

Round definition
A round MUST be one of the following two sequences:

1) Answering round:
   <think>...</think>
   <search>...</search>
   <search_results>...</search_results>
   <think>...</think>
   <expand>...</expand>
   <information>...</information>
   <think>...</think>
   <answer>...</answer>

2) Continuing round:
   <think>...</think>
   <search>...</search>
   <search_results>...</search_results>
   <think>...</think>
   <expand>...</expand>
   <information>...</information>
   <think>...</think>

You may perform as many rounds as needed.

Question: when was the first election held in india?

[ASSISTANT RESPONSE]
<think>To answer the question, I will first perform a search about The First General 
Election of India.</think> <search>The First General Election of India</search>

[USER PROMPT]
<search_results>
Rank=1 doc_id=17180261 score=0.8472
Title: 1951–52 elections in India
Snippet: 1951–52 elections in India The Election Commission of India held the first 
presidential elections of India on May 2, 1952. ...

Rank=2 doc_id=1748760 score=0.8442
Title: Politics of India
Snippet: the fundamental rights of citizens, issuing writs for their enforcement in cases 
of violation. Governments are formed through elections held every five years (unless otherwise 
specified), by parties that secure a majority of members in their respective lower houses (Lok 
Sabha in the central government and Vidhan Sabha in states). ...

Rank=3 doc_id=303938 score=0.8441
Title: History of the Republic of India
Snippet: a few weeks after independence by launching tribal ""lashkar"" (militia) from 
Waziristan, in an effort to secure Kashmir, the future of which hung in the balance. ...

Rank=4 doc_id=2481910 score=0.8429
Title: 1920 Indian general election
Snippet: 1920 Indian general election General elections were held in British India in 1920 
to elect members to the Imperial Legislative Council and the Provincial Councils. ...

Rank=5 doc_id=3063492 score=0.8415
Title: 1951–52 elections in India
Snippet: enlarged slightly by the transfer of minor territories from Bihar in 1956 under 
States Reorganisation Act, 1956. 1951–52 elections in India The Election Commission of India 
held the first presidential elections of India on May 2, 1952. ...
</search_results>

[ASSISTANT RESPONSE]
<think>The search results show documents about the first general election of India, but 
the snippets don't give enough details. I need to expand the most relevant documents 
to find when was the first election held in india. Document 17180261 seems most promising.
</think>
<expand>{"doc_ids": [17180261]}</expand>

[USER PROMPT]
<information>
[DOC 17180261]
"1951–52 elections in India"
1951–52 elections in India The Election Commission of India held the first presidential 
elections of India on May 2, 1952. Dr. Rajendra Prasad won his first election with 507,400 
votes over his nearest rival K.T. Shah who got 92,827 votes. General elections to the first 
Lok Sabha since independence were held in India between 25 October 1951 and 21 February 1952. 
The Indian National Congress (INC) stormed into power, winning 364 of the 489 seats. Jawaharlal 
Nehru became the first democratically elected Prime Minister of the country. : On 1 November 
1956, Ajmer State was merged into Rajasthan under States
</information>

[ASSISTANT RESPONSE]
<think>The paragraph states the electoral event took place during 1951–52.</think> 
<answer>1951–52</answer>
\end{verbatim}
\end{tcolorbox}

\begin{tcolorbox}[
    colback=gray!15,
    colframe=black,
    title=Example: Search-Augmented Reasoning -- HotpotQA (no control),
    sharp corners,
    fonttitle=\bfseries,
    breakable,
    enhanced,
    arc=4mm,
    rounded corners,
]
\scriptsize
\begin{verbatim}
[SYSTEM PROMPT]
You are a helpful assistant.

[USER PROMPT]
Answer the given question.

You MUST follow the protocol below.

CONTROL
- A control message may appear anywhere in the conversation in the form:
  <control>...</control>
- You MUST follow the <control> message that appears in the context.


General rules
- Whenever you receive NEW information (from <search_results>, <information>), you MUST 
  first reason inside <think>...</think>.
- You can call a search engine using: <search>query</search>.
  The environment will return snippets inside: <search_results>...</search_results>.
- If you want full text, you MUST decide inside <think>...</think>, then request expansion 
  using: <expand>{"doc_ids": [id1, id2, ...]}</expand>
  The environment will return the expanded full text inside: <information>...</info-
  rmation>. You can expand multiple documents in one call by listing multiple doc_ids.
- If no further external knowledge is needed, output the final answer inside <answer>...
  </answer>.


Answer normalization rules (VERY IMPORTANT)
- The final answer MUST EXACTLY match the canonical short answer.
- Output the SHORTEST possible answer span.
- Do NOT add explanations, appositives, or parentheses.
- Do NOT add extra words, punctuation, or formatting.
- Use the most common name form that appears as a standalone answer.
- If multiple aliases exist, choose the most standard short form.
- Case-sensitive matching is required.

Examples:
Q: how many episodes are in series 7 game of thrones?
Correct: <answer>seven</answer>

Q: when does season 5 of bates motel come out?
Correct: <answer>February 20, 2017</answer>


Round definition
A round MUST be one of the following two sequences:

1) Answering round:
   <think>...</think>
   <search>...</search>
   <search_results>...</search_results>
   <think>...</think>
   <expand>...</expand>
   <information>...</information>
   <think>...</think>
   <answer>...</answer>

2) Continuing round:
   <think>...</think>
   <search>...</search>
   <search_results>...</search_results>
   <think>...</think>
   <expand>...</expand>
   <information>...</information>
   <think>...</think>

You may perform as many rounds as needed.

Question: What UNESCO World Heritage Site is located in Tamar Ziskind's hometown?


[ASSISTANT RESPONSE]
<think>To answer the question, I will first perform a search about Heritage Sites and the 
Mediterranean Coast.</think>
<search>Heritage Sites and the Mediterranean Coast</search>


[USER PROMPT]
<search_results>
Rank=1 doc_id=20132687 score=0.8224
Title: Heritage coast
Snippet: Heritage coast A heritage coast is a strip of coastline in England and Wales, the 
extent of which is defined by agreement between the relevant statutory national agency and 
the relevant local authority. ...

Rank=2 doc_id=4178593 score=0.8125
Title: World Heritage Site
Snippet: World Heritage Site A World Heritage Site is a landmark or area which is selected 
by the United Nations Educational, Scientific and Cultural Organization (UNESCO) as having 
cultural, historical, scientific or other form of significance, and is legally protected by 
international treaties. ...

Rank=3 doc_id=18702191 score=0.8108
Title: Heritage coast
Snippet: has only very sporadic stretches. The first heritage coast was Beachy Head with 
its famous white cliffs. Heritage coasts listed clockwise around the English coast from 
Northumberland: Heritage coasts listed clockwise around Welsh coast from southeast: Heritage 
coast A heritage coast is a strip of coastline in England and Wales, the extent of which is 
defined by agreement between the relevant statutory national agency and the relevant local 
authority. ...

Rank=4 doc_id=10719879 score=0.8094
Title: Causses and Cévennes
Snippet: the impact thereof on national history. In 2009, ICOMOS also requested that 
France justify the boundary of the property and provide a basis for managing and 
sustaining the site's agro-pastoral traditions. ...

Rank=5 doc_id=7802883 score=0.8091
Title: Specially Protected Areas of Mediterranean Importance
Snippet: Specially Protected Areas of 
Mediterranean Importance Specially Protected Areas of Mediterranean Importance (SPAMI) 
are sites ""of importance for conserving the components of biological diversity in the 
Mediterranean; contain ecosystems specific to the Mediterranean area or the habitats of 
endangered species; are of special interest at the scientific, aesthetic, cultural or 
educational levels"" ...
</search_results>


[ASSISTANT RESPONSE]
<think>The first result is about Heritage coast but the snippet doesn't mention Tamar 
Ziskind or Haifa. I need to expand it to see the full content and find the relevant 
information.</think>
<expand>{"doc_ids": [20132687]}</expand>


[USER PROMPT]
<information>
[DOC 20132687]
"Heritage coast"
Heritage coast A heritage coast is a strip of coastline in England and Wales, the extent 
of which is defined by agreement between the relevant statutory national agency and the 
relevant local authority. Such areas are recognised for their natural beauty, wildlife and 
heritage and amongst the purposes of definition is support for these qualities and 
enabling enjoyment of them by the public. For England this national agency is Natural 
England (having formerly been the Countryside Agency) and for Wales it is Natural 
Resources Wales (which took over the role from its predecessor body, Countryside Council 
for Wales). 1,057 km
</information>


[ASSISTANT RESPONSE]
<think>The document is about Heritage coast but still doesn't mention Tamar Ziskind or 
Haifa. I need to search more specifically.</think>
<search>Tamar Ziskind Haifa UNESCO World Heritage Site</search>


[USER PROMPT]
<search_results>
Rank=1 doc_id=13745445 score=0.8266
Title: Haifa
Snippet: Haifa Haifa ( ' ; ') is the third-largest city in Israel – after Jerusalem and 
Tel Aviv– with a population of in . ...

Rank=2 doc_id=9256800 score=0.8197
Title: Tourism in Israel
Snippet: BC. Anati instead places the Exodus, based on other archaeological evidence at 
around 2300 BC Beit Guvrin-Maresha National Park is a national park in central Israel, 13 
kilometers from Kiryat Gat, encompassing the ruins ...

Rank=3 doc_id=20479906 score=0.8188
Title: Ein Tamar
Snippet: Ein Tamar earn their livelihood from agriculture. Peppers and melons the most 
popular crops. Ein Tamar Ein Tamar (, ""lit. ...

Rank=4 doc_id=20479905 score=0.8174
Title: Ein Tamar
Snippet: Ein Tamar Ein Tamar (, ""lit."" Date Palm Spring) is a moshav in the northern 
Arava valley in Israel. Located south of the Dead Sea, it falls under the jurisdiction of
Tamar Regional Council. ...

Rank=5 doc_id=19846497 score=0.8150
Title: Al-Maghtas
Snippet: place after an initiative of Jordanian royalty, namely Prince Ghazi. The site has then 
seen several archaeological digs, four papal visits and state visits, and attracts tourists and 
pilgrimage activity. ...
</search_results>


[ASSISTANT RESPONSE]
<think>The first result is about Haifa. I need to expand it to see the full text 
and find the relevant information.</think>
<expand>{"doc_ids": [13745445]}</expand>


[USER PROMPT]
<information>
[DOC 13745445]
"Haifa"
Haifa Haifa ( ' ; ') is the third-largest city in Israel – after Jerusalem and Tel Aviv– with 
a population of in . The city of Haifa forms part of the Haifa metropolitan area, the second- 
or third-most populous metropolitan area in Israel. It is home to the Bahá'í World Centre, a 
UNESCO World Heritage Site and a destination for Bahá'í pilgrims. Built on the slopes of Mount 
Carmel, the settlement has a history spanning more than 3,000 years. The earliest known settlement 
in the vicinity was Tell Abu Hawam, a small port city established in the Late Bronze
</information>


[ASSISTANT RESPONSE]
<think>The expanded document still doesn't mention Tamar Ziskind. I need to 
search more broadly for information about UNESCO World Heritage Site in Haifa.</think>
<search>UNESCO World Heritage Site in Haifa beach hill country Israel</search>


[USER PROMPT]
<search_results>
Rank=1 doc_id=13745445 score=0.8647
Title: Haifa
Snippet: Haifa Haifa ( ' ; ') is the third-largest city in Israel – after Jerusalem and 
Tel Aviv– with a population of in . ...

Rank=2 doc_id=13745498 score=0.8460
Title: Haifa
Snippet: five buildings with 85,000 square meters of space on a 31-duman (7.75 acre) site 
is being built adjacent to the Matam industrial park. ...

Rank=3 doc_id=14707018 score=0.8452
Title: Haifa Bay
Snippet: Haifa Bay The Bay of Haifa or Haifa Bay (, ""Mifratz Haifa""), formerly Gulf of 
Acre or Bay of Acre, is a bay along the Mediterranean coast of Northern Israel. ...

Rank=4 doc_id=5070278 score=0.8446
Title: Haifa
Snippet: including the Port of Haifa. The middle level is on the slopes of Mount Carmel 
and consists of older residential neighborhoods, while the upper level consists of modern 
neighborhoods looking over the lower tiers. ...

Rank=5 doc_id=5070277 score=0.8412
Title: Haifa
Snippet: of liberal Arabic 
speaking culture, as it was under British colonial rule. The Arabic speaking 
neighborhoods, which are mixed Muslim and Christian, are in the lowlands near the sea
while Jewish neighborhoods are at highe...
</search_results>


[ASSISTANT RESPONSE]
<think>The search results are not giving me the specific information. I need to think 
more about geography: Haifa is in Israel and has a site. The famous ruins at Tel Aviv 
are UNESCO sites. Let me search for known UNESCO World Heritage sites in Haifa.</think>
<search>Tel Aviv UNESCO World Heritage Site Haifa</search>


[USER PROMPT]
<search_results>
Rank=1 doc_id=13745445 score=0.8806
Title: Haifa
Snippet: Haifa Haifa ( ' ; ') is the third-largest city in Israel – after Jerusalem and 
Tel Aviv– with a population of in . ...

Rank=2 doc_id=3555477 score=0.8433
Title: Tel Aviv
Snippet: outlook of the Bauhaus and similar schools to the local conditions there, 
creating what is recognized as the largest concentration of buildings in the International 
Style in the world. ...

Rank=3 doc_id=13745498 score=0.8428
Title: Haifa
Snippet: five buildings with 85,000 square meters of space on a 31-duman (7.75 acre) site 
is being built adjacent to the Matam industrial park. ...

Rank=4 doc_id=1346942 score=0.8374
Title: Haifa
Snippet: Haifa is considered a relative haven for coexistence between Jews and Arabs. Two 
respected academic institutions, the University of Haifa and the Technion, are located in 
Haifa, in addition to the largest k-12 school in ...

Rank=5 doc_id=1346941 score=0.8371
Title: Haifa
Snippet: Age (14th century BCE). In the 3rd century CE, Haifa was known as a dye-making 
center. Over the millennia, the city has changed hands: being conquered and ruled by the 
Canaanites, Israelites, Phoenicians, Persians, Hasmoneans, Romans, Byzantines, Arabs, 
Crusaders, Ottomans, and the British. ...
</search_results>


[ASSISTANT RESPONSE]
<think>The snippet doesn't show UNESCO information. I need to expand the top document to 
find if it mentions UNESCO sites.</think>
<expand>{"doc_ids": [13745445]}</expand>


[USER PROMPT]
<information>
[DOC 13745445]
Haifa
Haifa Haifa ( ' ; ') is the third-largest city in Israel – after Jerusalem and Tel Aviv– 
with a population of in . The city of Haifa forms part of the Haifa metropolitan area, the 
second- or third-most populous metropolitan area in Israel. It is home to the Bahá'í World 
Centre, a UNESCO World Heritage Site and a destination for Bahá'í pilgrims. Built on the 
slopes of Mount Carmel, the settlement has a history spanning more than 3,000 years. The 
earliest known settlement in the vicinity was Tell Abu Hawam, a small port city 
established in the Late Bronze
</information>


[ASSISTANT RESPONSE]
<think>The document mentions the Bahá'í World Centre as a UNESCO World Heritage Site. The 
question asks: "What UNESCO World Heritage Site is located in Tamar Ziskind's hometown?" 
The hometown is Haifa, Israel. The answer should be the name of the site. The canonical 
short answer is "Bahá'í World Centre".</think>
<answer>Bahá'í World Centre</answer>
\end{verbatim}
\end{tcolorbox}

\begin{tcolorbox}[
    colback=gray!15,
    colframe=black,
    title=Example: Search-Augmented Reasoning -- NQ (with control),
    sharp corners,
    fonttitle=\bfseries,
    breakable,
    enhanced,
    arc=4mm,
    rounded corners,
]
\scriptsize
\begin{verbatim}
[SYSTEM PROMPT]
You are a helpful assistant.

[USER PROMPT]
Answer the given question.

You MUST follow the protocol below.

CONTROL
- A control message may appear anywhere in the conversation in the form:
  <control>...</control>
- You MUST follow the <control> message that appears in the context.


General rules
- Whenever you receive NEW information (from <search_results>, <information>), you MUST 
  first reason inside <think>...</think>.
- You can call a search engine using: <search>query</search>.
  The environment will return snippets inside: <search_results>...</search_results>.
- If you want full text, you MUST decide inside <think>...</think>, then request expansion 
  using: <expand>{"doc_ids": [id1, id2, ...]}</expand>
  The environment will return the expanded full text inside: <information>...</info-
  rmation>. You can expand multiple documents in one call by listing multiple doc_ids.
- If no further external knowledge is needed, output the final answer inside <answer>...
  </answer>.


Answer normalization rules (VERY IMPORTANT)
- The final answer MUST EXACTLY match the canonical short answer.
- Output the SHORTEST possible answer span.
- Do NOT add explanations, appositives, or parentheses.
- Do NOT add extra words, punctuation, or formatting.
- Use the most common name form that appears as a standalone answer.
- If multiple aliases exist, choose the most standard short form.
- Case-sensitive matching is required.

Examples:
Q: how many episodes are in series 7 game of thrones?
Correct: <answer>seven</answer>

Q: when does season 5 of bates motel come out?
Correct: <answer>February 20, 2017</answer>


Round definition
A round MUST be one of the following two sequences:

1) Answering round:
   <think>...</think>
   <search>...</search>
   <search_results>...</search_results>
   <think>...</think>
   <expand>...</expand>
   <information>...</information>
   <think>...</think>
   <answer>...</answer>

2) Continuing round:
   <think>...</think>
   <search>...</search>
   <search_results>...</search_results>
   <think>...</think>
   <expand>...</expand>
   <information>...</information>
   <think>...</think>

You may perform as many rounds as needed.

Question: who is the owner of reading football club?


[ASSISTANT RESPONSE]
<think>To answer the question, I will first perform a search about Ownership of Reading 
Football Club.</think>
<search>Ownership of Reading Football Club</search>


[USER PROMPT]
<search_results>
Rank=1 doc_id=16333587 score=0.8716
Title: Anton Zingarevich
Snippet: The couple married in late 2009 and had a child. Anton Zingarevich Anton 
Zingarevich (Russian: Anton Zingarevich) is a Russian businessman best known for being 
the former owner of Reading Football Club, who play in the En...

Rank=2 doc_id=16333583 score=0.8715
Title: Anton Zingarevich
Snippet: Anton Zingarevich Anton Zingarevich (Russian: Anton Zingarevich) is a Russian 
businessman best known for being the former owner of Reading Football Club, who play in 
the English Championship. ...

Rank=3 doc_id=13506636 score=0.8543
Title: Reading F.C.
Snippet: 2016–17 season. Cherwell Software took sponsorship of the back of the home shirt 
for the 2017–18 season while Euro Cake sponsored the back of the away shirt again for the 
third consecutive 
season. ...

Rank=4 doc_id=3552730 score=0.8498
Title: Reading F.C.
Snippet: Reading F.C. Reading Football Club ( ) is a professional association football club 
based in Reading, Berkshire, England. The team play in the Championship, the second tier of 
English football. ...

Rank=5 doc_id=4273629 score=0.8477
Title: Ownership of Arsenal F.C.
Snippet: Ownership of Arsenal F.C. The ownership of Arsenal Football Club is considerably 
different from that of other clubs in English football. ...
</search_results>


[USER PROMPT]
<control>Expand the retrieved documents: [16333583]</control>


[ASSISTANT RESPONSE]
<think>The search results show that Reading Football Club was owned by Anton Zingarevich
until 2009 and has undergone ownership changes since then. The question is "who is the 
owner of reading football club?" It's asking for the current owner, but the search results
don't directly give the current owner. I need to expand the most relevant documents to get 
the current ownership status. I will expand the second document.</think>
<expand>{"doc_ids": [16333583]}</expand>


[USER PROMPT]
<information>
[DOC 16333583]
""Anton Zingarevich"
Anton Zingarevich (Russian: Anton Zingarevich) is a Russian businessman best known for being 
the former owner of Reading Football Club, who play in the English Championship. Zingarevich's 
bid to take over the club was announced in January 2012, and he formally purchased the club 
on 29 May 2012. Since then, Zingarevich has had minimum interest in the running of the club, 
however did make the decision to sack manager Brian McDermott, and the subsequent employment 
of Nigel Adkins. Zingarevich is the son of Boris Zingarevich, a Russian multi-billionaire. At the 
age of 16, Zingarevich began education at Bearwood
</information>


[ASSISTANT RESPONSE]
<think>The expanded document says "The former owner of Reading Football Club was Anton Zingarevich." 
This is an outdated piece of information. The question 
is "who is the owner of reading football club?" It's asking for the current owner. I need 
to search for more recent information about the current owner.</think>
<search>current owner Reading Football Club</search>


[USER PROMPT]
<search_results>
Rank=1 doc_id=16333587 score=0.8866
Title: Anton Zingarevich
Snippet: The couple married in late 2009 and had a child. Anton Zingarevich Anton 
Zingarevich (Russian: Anton Zingarevich) is a Russian businessman best known for being 
the former owner of Reading Football Club, who play in the En...

Rank=2 doc_id=16333583 score=0.8836
Title: Anton Zingarevich
Snippet: Anton Zingarevich Anton Zingarevich (Russian: Anton Zingarevich) is a Russian 
businessman best known for being the former owner of Reading Football Club, who play in 
the English Championship. ...

Rank=3 doc_id=1911155 score=0.8592
Title: Nigel Howe
Snippet: Nigel Howe Nigel Howe (born 7 April 1958) is a British property developer, and a 
director at Reading Football Club, who currently play in the SkyBet Championship in 
England. ...

Rank=4 doc_id=13506629 score=0.8562
Title: Reading F.C.
Snippet: Wembley where they lost on penalties following 0–0 draw after extra time. On 21 
March 2018, Stam left Reading with immediate effect. ...

Rank=5 doc_id=13506636 score=0.8546
Title: Reading F.C.
Snippet: 2016–17 season. Cherwell Software took sponsorship of the back of the home shirt for 
the 2017–18 season while Euro Cake sponsored the back of the away shirt again for the third 
consecutive season. ...
</search_results>


[USER PROMPT]
<control>Expand the retrieved documents: [13506636]</control>


[ASSISTANT RESPONSE]
<think>The search results still point to Anton Zingarevich as a former owner. The question 
is "who is the owner of reading football club?" It's asking for the current owner. I need 
to search for more recent ownership details. I will expand doc_id=13506636 which is about 
the club.</think>
<expand>{"doc_ids": [13506636]}</expand>


[USER PROMPT]
<information>
[DOC 13506636]
"Reading F.C."
2016–17 season. Cherwell Software took sponsorship of the back of the home shirt for the 
2017–18 season while Euro Cake sponsored the back of the away shirt again for the third 
consecutive season. Reading Football Club ownership structure: 75% Owned by Renhe Sports 
Management Ltd, 100% owned by Xiu Li Dai and Yongge Dai. 25% Owned by Narin Niruttinanon 
Reading F.C. Directors: Xiu Li Dai, Yongge Dai, Narin Niruttinanon, Nigel Howe As of 1 
July 2018, the following players are those in the senior squad who are under 18. <br> 
Reading hold the record for the number of successive league ...
</information>


[ASSISTANT RESPONSE]
<think>The expanded information from doc_id=13506636 clearly states the ownership 
structure: 75% Owned by Renhe Sports Management Ltd, 100% owned by Xiu Li Dai and Yongge 
Dai. However, the question "who is the owner of reading football club?" likely refers to 
the single most relevant owner. The information shows two owners: Xiu Li Dai and Yongge 
Dai, with Yongge Dai being the 100% owner of the club (as 100% owned). So the owner is 
either Xiu Li Dai or Yongge Dai. But typically, the question might expect the primary 
owner. I need to check if there is any other owner besides Xiu Li Dai and Yongge Dai. The 
document mentions Renhe Sports Management Ltd, which is 75% owned. So there is a third 
party. But the question asks "who is the owner." In the common answer format, when 
multiple entities are owners, you might answer with the most significant one. Here, Yongge 
Dai appears to be 100% owner. I should check if there are other documents or recent news 
updates. However, the search results so far are not giving a clearly current owner. I need 
to search more information.</think>
<search>Yongge Dai</search>


[USER PROMPT]
<control>Stop searching</control>


[ASSISTANT RESPONSE]
<think>The control message says 'Stop searching', so I cannot use search or expand. I must 
answer based on the available information. The current answer is not provided by the 
search result or the expanded information. The only relevant information I have from 
doc_id=13506636 is: "75% Owned by Renhe Sports Management Ltd, 100% owned by Xiu Li Dai 
and Yongge Dai." The question asks "who is the owner of reading football club?" The owner 
is Xiu Li Dai and Yongge Dai. I need to give the shortest possible answer span. The most 
standard short form for the owners is likely "Xiu Li Dai" (singular or plural? Since it's 
100% owned by them, both are owners but in competitive terms, one might be called primary 
owner. However, the document mentions "Xiu Li Dai, Yongge Dai" in a list of owners. So the 
answer is Xiu Li Dai. I will answer with Xiu Li Dai.</think>
<answer>Xiu Li Dai</answer>
\end{verbatim}
\end{tcolorbox}

\begin{tcolorbox}[
    colback=gray!15,
    colframe=black,
    title=Example: Search-Augmented Reasoning -- HotpotQA (with control),
    sharp corners,
    fonttitle=\bfseries,
    breakable,
    enhanced,
    arc=4mm,
    rounded corners,
]
\scriptsize
\begin{verbatim}
[SYSTEM PROMPT]
You are a helpful assistant.

[USER PROMPT]
Answer the given question.

You MUST follow the protocol below.

CONTROL
- A control message may appear anywhere in the conversation in the form:
  <control>...</control>
- You MUST follow the <control> message that appears in the context.


General rules
- Whenever you receive NEW information (from <search_results>, <information>), you MUST 
  first reason inside <think>...</think>.
- You can call a search engine using: <search>query</search>.
  The environment will return snippets inside: <search_results>...</search_results>.
- If you want full text, you MUST decide inside <think>...</think>, then request expansion 
  using: <expand>{"doc_ids": [id1, id2, ...]}</expand>
  The environment will return the expanded full text inside: <information>...</info-
  rmation>. You can expand multiple documents in one call by listing multiple doc_ids.
- If no further external knowledge is needed, output the final answer inside <answer>...
  </answer>.


Answer normalization rules (VERY IMPORTANT)
- The final answer MUST EXACTLY match the canonical short answer.
- Output the SHORTEST possible answer span.
- Do NOT add explanations, appositives, or parentheses.
- Do NOT add extra words, punctuation, or formatting.
- Use the most common name form that appears as a standalone answer.
- If multiple aliases exist, choose the most standard short form.
- Case-sensitive matching is required.

Examples:
Q: how many episodes are in series 7 game of thrones?
Correct: <answer>seven</answer>

Q: when does season 5 of bates motel come out?
Correct: <answer>February 20, 2017</answer>


Round definition
A round MUST be one of the following two sequences:

1) Answering round:
   <think>...</think>
   <search>...</search>
   <search_results>...</search_results>
   <think>...</think>
   <expand>...</expand>
   <information>...</information>
   <think>...</think>
   <answer>...</answer>

2) Continuing round:
   <think>...</think>
   <search>...</search>
   <search_results>...</search_results>
   <think>...</think>
   <expand>...</expand>
   <information>...</information>
   <think>...</think>

You may perform as many rounds as needed.

Question: What county in Kentucky has a population over 43,000 and has Kentucky Route 46 
run through it?


[ASSISTANT RESPONSE]
<think>
To find the county in Kentucky that has a population over 43,000 and has Kentucky Route 
46 run through it, I should first search for Kentucky counties and their populations.
</think>
<search>Kentucky counties with population over 43,000 and Kentucky Route 46</search>


[USER PROMPT]
<search_results>
Rank=1 doc_id=17486196 score=0.8561
Title: Kentucky Route 47
Snippet: through more rural areas before heading to the north into forests and intersect-
ing the eastern terminus of KY 1112. KY 47 passes through more farm fields and woodland 
before crossing a CSX railroad line and heading into ...

Rank=2 doc_id=17486194 score=0.8549
Title: Kentucky Route 47
Snippet: Kentucky Route 47 Kentucky Route 47 is a 9.851-mile state highway in Kentucky 
that runs from KY 36 and KY 467 in Sanders to US 42 in Ghent KY 47 mostly stays in Carroll 
County but makes a short entrance into Gallatin Cou...

Rank=3 doc_id=3091540 score=0.8527
Title: Kentucky Route 43
Snippet: Kentucky Route 43 Kentucky Route 43 is a state highway in Kentucky that runs from 
BUS KY 55 and Boone Station Road northeast of Shelbyville to US 421 southeast of 
Pleasureville. ...

Rank=4 doc_id=3091542 score=0.8524
Title: Kentucky Route 43
Snippet: through more rural areas before it ends at an intersection with US 421. Kentucky 
Route 43 Kentucky Route 43 is a state highway in Kentucky that runs from BUS KY 55 and 
Boone Station Road northeast of Shelbyville to US 421 southeast of Pleasureville. ...

Rank=5 doc_id=319230 score=0.8509
Title: U.S. Route 45 in Kentucky
Snippet: turns to the northwest and ends at a junction with US 45 south of the Brookport 
Bridge over the Ohio River from its origin. ...
</search_results>


[USER PROMPT]
<control>Do not expand any retrieved documents. Continue the search.</control>


[ASSISTANT RESPONSE]
<think>
I have found some results related to Kentucky counties with population over 43,000 and the
routes Kentucky Route 47 and Kentucky Route 43. However, none of the snippets clearly
mention a county with a population over 43,000 that has Kentucky Route 46 running through
it. I need to search for information specifically related to Kentucky Route 46 and
counties.
</think>
<search>Kentucky Route 46 and counties with population over 43,000</search>


[USER PROMPT]
<search_results>
Rank=1 doc_id=17486196 score=0.8631
Title: Kentucky Route 47
Snippet: through more rural areas before heading to the north into forests and intersect-
ing the eastern terminus of KY 1112. KY 47 passes through more farm fields and woodland 
before crossing a CSX railroad line and heading into ...

Rank=2 doc_id=17486194 score=0.8613
Title: Kentucky Route 47
Snippet: Kentucky Route 47 Kentucky Route 47 is a 9.851-mile state highway in Kentucky 
that runs from KY 36 and KY 467 in Sanders to US 42 in Ghent KY 47 mostly stays in Carroll 
County but makes a short entrance into Gallatin Cou...

Rank=3 doc_id=319230 score=0.8572
Title: U.S. Route 45 in Kentucky
Snippet: turns to the northwest and ends at a junction with US 45 south of the Brookport 
Bridge over the Ohio River from its origin. ...

Rank=4 doc_id=7320080 score=0.8562
Title: Kentucky Route 46
Snippet: Kentucky Route 46 Kentucky Route 46 (KY 46) is a state highway in Nelson County,
Kentucky. KY 46 runs from KY 52 south of Nelsonville to KY 49 south of Greenbrier. ...

Rank=5 doc_id=3091540 score=0.8549
Title: Kentucky Route 43
Snippet: Kentucky Route 43 Kentucky Route 43 is a state highway in Kentucky that runs from
BUS KY 55 and Boone Station Road northeast of Shelbyville to US 421 southeast of 
Pleasureville. ...
</search_results>


[USER PROMPT]
<control>Expand the retrieved documents: [7320080]</control>


[ASSISTANT RESPONSE]
<think>
The search results found show that Kentucky Route 46 runs through Nelson County, and the
snippet mentions that it ends at KY 49 south of Greenbrier. However, the snippet does not
explicitly state the population of Nelson County or whether it has a population over
43,000. I should expand the information about Nelson County to find the required details.
</think>
<expand>{"doc_ids": ["7320080"]} </expand>


[USER PROMPT]
<information>
[DOC 7320080]
"Kentucky Route 46"
Kentucky Route 46 Kentucky Route 46 (KY 46) is a state highway in Nelson County,
Kentucky. KY 46 runs from KY 52 south of Nelsonville to KY 49 south of Greenbrier. KY 46
heads east from KY 52 as Nat Rogers Road. About northwest of Culvertown, KY 46 begins
heading in a southeast direction. At Culvertown, KY 46 meets U.S. Route 31E (US 31E) and
joins the route heading northbound. KY 46 then branches off US 31E at Balltown. KY 46
heads due south for , before heading southeast. KY 46 continues for about before
terminating at KY 49 about
</information>


[ASSISTANT RESPONSE]
<think>
The information found mentions that Kentucky Route 46 runs through Nelson County,
Kentucky. However, there is no explicit mention of the population of Nelson County. To
proceed, I will need to search for the population of Nelson County, Kentucky.
</think>
<search>Nelson County, Kentucky population</search>


[USER PROMPT]
<search_results>
Rank=1 doc_id=21001401 score=0.8752
Title: Nelson County, Kentucky
Snippet: created Virginia county was named in his honor. According to the U.S. Census 
Bureau, the county has a total area of , of which is land and (1. ...

Rank=2 doc_id=21001402 score=0.8654
Title: Nelson County, Kentucky
Snippet: the population were Hispanics or Latinos of any race. There were 16,826 
households out of which 36.60% had children under the age of 18 living with them, 52. ...

Rank=3 doc_id=21001400 score=0.8611
Title: Nelson County, Kentucky
Snippet: Nelson County, Kentucky Nelson County is a county located in the U.S. state of 
Kentucky. As of the 2010 census, the population was 43,437. ...

Rank=4 doc_id=21001403 score=0.8550
Title: Nelson County, Kentucky
Snippet: 18 to 24, 26.47% from 25 to 44, 27.84% from 45 to 64, and 11.73% who were 65 or 
older. ...

Rank=5 doc_id=21001404 score=0.8429
Title: Nelson County, Kentucky
Snippet: for the county was $18,120. About 10.00% of families and 12.20% of the population 
were below the poverty line, including 15.
</search_results>


[USER PROMPT]
<control>Stop searching</control>


[ASSISTANT RESPONSE]
<think>
The control message says 'Stop searching', so I cannot use search or expand. I must 
answer based on the available information. 

The search results provide population data for Nelson County, Kentucky. According to the
U.S. Census Bureau, as of the 2010 census, Nelson County had a population of 43,437. This
population exceeds 43,000, indicating that Nelson County meets the criteria of having a
population over 43,000.

The snippet also does not specify whether Kentucky Route 46 runs through Nelson County.
However, based on previous search results, we know that Kentucky Route 46 runs through
Nelson County, Kentucky.

Therefore, I can conclude that the county in Kentucky with a population over 43,000 and
which Kentucky Route 46 runs through is Nelson County.
</think>
<answer>Nelson County</answer>
\end{verbatim}
\end{tcolorbox}

\end{document}